%% file: ada_norms.tex
\documentclass{article}

\usepackage{microtype}
\usepackage{graphicx}
\usepackage{subfigure}
\usepackage{booktabs} 

\usepackage{amssymb}
\usepackage{amsmath}
\usepackage{amsthm}
\usepackage{bbm}
\usepackage{enumerate}
\usepackage{enumitem}
\usepackage{microtype}
\usepackage{graphicx}
\usepackage[export]{adjustbox}
\usepackage{float}
\usepackage{algorithm}
\usepackage{algorithmic}
\usepackage{framed}
\usepackage{mdframed}
\usepackage{thmtools} 
\usepackage{thm-restate}
\usepackage{subfigure}
\usepackage{rotating}
\usepackage{hyperref}

\usepackage[numbers]{natbib}
\usepackage[preprint,nonatbib]{neurips_2020}

\input{defs}
\input{math_commands.tex}

\title{Adaptive norms for deep learning with regularized Newton methods} 
\author{%
  Jonas Kohler*, Leonard Adolphs*, Aurelien lucchi \\
  Department of Computer Science\\
  ETH Zurich\\\thanks{Shared first authorship. Correspondence to \textit{jonas.kohler@inf.ethz.ch}.}
  }

\begin{document}

\maketitle

\begin{abstract}
We investigate the use of regularized Newton methods with adaptive norms for optimizing neural networks. This approach can be seen as a second-order counterpart of adaptive gradient methods, which we here show to be interpretable as first-order trust region methods with ellipsoidal constraints. In particular, we prove that the preconditioning matrix used in RMSProp and Adam satisfies the necessary conditions for provable convergence of second-order trust region methods with standard worst-case complexities on general non-convex objectives. Furthermore, we run experiments across different neural architectures and datasets to find that the ellipsoidal constraints constantly outperform their spherical counterpart both in terms of number of backpropagations and asymptotic loss value. Finally, we find comparable performance to state-of-the-art first-order methods in terms of backpropagations, but further advances in hardware are needed to render Newton methods competitive in terms of computational time. 
\end{abstract}

\input{01_introduction}
\input{02_related_work}
\input{03_adagrad.tex}
\input{04_2nd_order.tex}
\input{06_experiments}
\input{07_conclusion.tex}

\bibliography{ada_norms}
\bibliographystyle{plain}

\appendix

\input{08_appendix.tex}
\end{document}

%% file: defs.tex
\newcommand{\norm}[1]{{}\left\| #1 \right\|}

\newcommand{\g}{{\bf g}}
\newcommand{\zero}{{\bf 0}}

\newcommand{\s}{{\bf s}}

\newcommand{\w}{{\bf w}}
\renewcommand{\u}{{\bf u}}

\newcommand{\x}{{\bf x}}
\newcommand{\y}{{\bf y}}

\DeclareMathOperator{\diag}{diag}  
\usepackage{mathtools}

\def\Am{{\bf A}}
\def\Bm{{\bf B}}
\def\Hm{{\bf H}}

\def\Xm{{\bf X}}
\def\Gm{{\bf G}}

\def\Dm{{\bf D}}
\def\Wm{{\bf W}}

\newcommand{\B}{{\mathcal B}}

\renewcommand{\H}{{\cal H}}

\newtheorem{theorem}{Theorem}
\newtheorem{lemma}{Lemma}
\newtheorem{proposition}{Proposition}
\newtheorem{corollary}{Corollary}
\newtheorem{assumption}{Assumption}
\theoremstyle{definition}
\newtheorem{definition}{Definition}

\usepackage{eqparbox}
\renewcommand{\algorithmiccomment}[1]{\hfill\eqparbox{COMMENT}{\# #1}}

%% file: math_commands.tex
\usepackage{amsmath,amsfonts,bm}

\def\eqref#1{equation~\ref{#1}}

\def\1{\bm{1}}

\DeclareMathAlphabet{\mathsfit}{\encodingdefault}{\sfdefault}{m}{sl}
\SetMathAlphabet{\mathsfit}{bold}{\encodingdefault}{\sfdefault}{bx}{n}

\newcommand{\E}{\mathbb{E}}

\newcommand{\R}{\mathbb{R}}

\DeclareMathOperator*{\argmin}{arg\,min}

\DeclareMathOperator{\Tr}{Tr}

%% file: 01_introduction.tex

\section{Introduction}

We consider finite-sum optimization problems of the form
\begin{equation} 
\min_{\w \in \R^d}  \left [ \mathcal{L}(\w) := \sum_{i=1}^n \ell(f(\w,\x_i,\y_i)) \right],
\label{eq:f_x}
\end{equation}
which typically arise in neural network training, e.g. for empirical risk minimization over a set of data points $(\x_i,\y_i) \in \mathbb{R}^{in}\times \mathbb{R}^{out}, i=1,\ldots,n$. Here, $\ell:\mathbb{R}^{out}\times \mathbb{R}^{out}\rightarrow\mathbb{R}^+$ is a convex loss function
and $f:\mathbb{R}^{in}\times \mathbb{R}^d\rightarrow\mathbb{R}^{out}$ represents the neural network mapping parameterized by the concatenation of the weight layers $\w \in \mathbb{R}^d$, which is non-convex due to its multiplicative nature and potentially non-linear activation functions. We assume that $\mathcal{L}$ is lower bounded and twice differentiable, i.e. $\mathcal{L} \in C^2(\R^d,\R)$ and consider finding a first- and second-order stationary point $\bar{\w}$ for which  $\|\nabla \mathcal{L}(\bar{\w})\| \leq \epsilon_g$ and $\lambda_{\min}\left(\nabla^2\mathcal{L}(\bar{\w})\right)\geq -\epsilon_H$.  

In the era of deep neural networks, stochastic gradient descent (SGD) is one of the most widely used training algorithms~\citep{bottou2010large}. What makes SGD so attractive is its simplicity and per-iteration cost that is independent of the size of the training set ($n$) and scale linearly in the dimensionality ($d$). However, gradient descent is known to be inadequate to optimize functions that are ill-conditioned \citep{nesterov2013introductory,shalev2017failures} and thus adaptive gradient methods that employ dynamic, coordinate-wise learning rates based on past gradients---including Adagrad \citep{duchi2011adaptive}, RMSprop \citep{tieleman2012lecture} and Adam \citep{kingma2014adam}---have become a popular alternative, often providing significant speed-ups over SGD.

From a theoretical perspective, Newton methods provide stronger convergence guarantees by appropriately transforming the gradient in ill-conditioned regions according to second-order derivatives. It is precisely this Hessian information that allows \emph{regularized} Newton methods to enjoy superlinear local convergence as well as to provably escape saddle points~\citep{conn2000trust}. 
While second-order algorithms have a long-standing history even in the realm of neural network training \citep{hagan1994training,becker1988improving}, they were mostly considered as too computationally and memory expensive for practical applications. Yet, the seminal work of \cite{martens2010deep} renewed interest for their use in deep learning by proposing efficient \textit{Hessian-free} methods that only access second-order information via matrix-vector products which can be computed at the cost of an additional backpropagation \citep{pearlmutter1994fast,schraudolph2002fast}. Among the class of regularized Newton methods, trust region \citep{conn2000trust} and cubic regularization algorithms \citep{cartis2011adaptive} are the most principled approaches in the sense that they yield the strongest convergence guarantees. Recently, stochastic extensions have emerged ~\citep{xu2017newton,yao2018inexact,kohler2017sub,gratton2017complexity}, which suggest their applicability for deep learning.


We here propose a simple modification to make TR methods even more suitable for neural
network training. Particularly, we build upon the following alternative view on adaptive
gradient methods:

\textit{While gradient descent can be interpreted as a spherically constrained first-order TR method, preconditioned gradient methods---such as Adagrad---can be seen as first-order TR methods with ellipsoidal trust region constraint.}

This observation is particularly interesting since spherical constraints are blind to the underlying geometry of the problem, but ellipsoids can adapt to local landscape characteristics, thereby allowing for more suitable steps in regions that are ill-conditioned. We will leverage this analogy and investigate the use of the Adagrad and RMSProp preconditioning matrices as \textit{ellipsoidal} trust region shapes within a stochastic second-order TR algorithm \citep{xu2017second,yao2018inexact}. Since no ellipsoid fits all objective functions, our main contribution lies in the identification of adequate matrix-induced constraints that lead to provable convergence and significant practical speed-ups for the specific case of deep learning. On the whole, our contribution is threefold: 
\begin{itemize}
\setlength\itemsep{0.02em}
\item We provide a new perspective on adaptive gradient methods that contributes to a better understanding of their inner-workings. 
\item We investigate the first application of ellipsoidal TR methods for deep learning. We show that the RMSProp matrix can directly be applied as constraint inducing norm in second-order TR algorithms while \textit{preserving all convergence guarantees} (Theorem~\ref{th:rate}). 
\item Finally, we provide an experimental benchmark across different real-world datasets and architectures (Section 5). We compare second-order methods also to adaptive gradient methods and show results in terms of backpropagations, epochs, and wall-clock time; a comparison we were not able to find in the literature. 
\end{itemize}

Our main empirical results demonstrate that ellipsoidal constraints prove to be a very effective modification of the trust region method in the sense that they constantly outperform the spherical TR method, both in terms of number of backprogations and asymptotic loss value on a variety of tasks.  

%% file: 02_related_work.tex

\section{Related work}
\label{sec:related_work}
\label{sec:rel_work_2nd_order}

\paragraph{First-order methods} The prototypical method for optimizing Eq.~(\ref{eq:f_x}) is SGD~\citep{robbins1985stochastic}. The practical success of SGD in non-convex optimization is unquestioned and theoretical explanations of this phenomenon are starting to appear. Recent findings suggest the ability of this method to escape saddle points and reach local minima in polynomial time, but they either need to artificially add noise to the iterates \citep{ge2015escaping,lee2016gradient} or make an assumption on the inherent noise of SGD \citep{daneshmand2018escaping}. For neural networks, a recent line of research proclaims the effectiveness of SGD, but the results come at the cost of strong assumptions such as heavy over-parametrization and Gaussian inputs \citep{du2017gradient,brutzkus2017globally,li2017convergence,du2018power,allen2018convergence}. Adaptive gradient methods \citep{duchi2011adaptive,tieleman2012lecture,kingma2014adam} build on the intuition that larger (smaller) learning rates for smaller (larger) gradient components balance their respective influences and thereby the methods behave as if optimizing a more isotropic surface. 
Such approaches have first been suggested for neural nets by \cite{lecun2012efficient} and convergence guarantees are starting to appear \citep{ward2018adagrad,li2018convergence}. However, these are not superior to the $\mathcal{O}(\epsilon_g^{-2})$ worst-case complexity of standard gradient descent~\citep{cartis2012much}.

\vspace{-0.1cm}
\paragraph{Regularized Newton methods}
The most principled class of regularized Newton methods are trust region (TR) and adaptive cubic regularization algorithms (ARC) \citep{conn2000trust,cartis2011adaptive}, which repeatedly optimize a local Taylor model of the objective while making sure that the step does not travel too far such that the model stays accurate. While the former finds first-order stationary points within $\mathcal{O}(\epsilon_g^{-2})$, ARC only takes at most $\mathcal{O}(\epsilon_g^{-3/2})$. However, simple modifications to the TR framework allow these methods to obtain the same accelerated rate~\citep{curtis2017trust}. Both methods take at most $\mathcal{O}(\epsilon_H^{-3})$ iterations to find an $\epsilon_H$ approximate second-order stationary point \citep{cartis2012complexity}. These rates are optimal for second-order Lipschitz continuous functions~\citep{carmon2017lower,cartis2012complexity} and they can be retained even when only sub-sampled gradient and Hessian information is used \citep{kohler2017sub,yao2018inexact,xu2017newton,blanchet2016convergence,liu2018stochastic,cartis2017global}. Furthermore, the involved Hessian information can be computed solely based on Hessian-vector products, which are implementable efficiently for neural networks \citep{pearlmutter1994fast}. This makes these methods particularly attractive for deep learning, but the empirical evidence of their applicability is rather limited. We are only aware of the works of \cite{liu2018stochastic} and \cite{xu2017second}, which report promising first results but are by no means fully encompassing. 

\vspace{-0.1cm}
\paragraph{Gauss-Newton methods}
An interesting line of research proposes to replace the Hessian by (approximations of) the generalized-Gauss-Newton matrix (GGN) within a Levenberg-Marquardt framework\footnote{This algorithm is a simplified TR method, initially tailored for non-linear least squares problems \citep{nocedal2006nonlinear}} \citep{lecun2012efficient,martens2010deep,martens2015optimizing}. 
As the GGN matrix is always positive semidefinite, these methods cannot leverage negative curvature to escape saddles and hence, there exist no second-order convergence guarantees. Furthermore, there are cases in neural networks where the Hessian is better conditioned than the GGN matrix \citep{mizutani2008second}. Nevertheless, the above works report promising preliminary results, most notably \cite{grosse2016kronecker} find that K-FAC can be faster than SGD on a small convnet. On the other hand, recent findings report performance at best comparable to SGD on the much larger ResNet architecture \citep{ma2019inefficiency}. Moreover, \cite{xu2017second} reports many cases where TR and GGN algorithms perform similarly. This line of work can be seen as complementary to our approach since it is straightforward to replace the Hessian in the TR framework with the GGN matrix. Furthermore, the preconditioners used in \cite{martens2010deep} and \cite{chapelle2011improved}, namely diagonal estimates of the empirical Fisher and Fisher matrix, respectively, can directly be used as matrix norms in our ellipsoidal TR framework.

\vspace{-0.1cm}

%% file: 03_adagrad.tex
\section{An alternative view on adaptive gradient methods}\label{sec:adaptive_grad_methods}
Adaptively preconditioned gradient methods update iterates as $\w_{t+1}= \w_t-\eta_t \Am_t^{-1/2} \g_t,$ where $\g_t$ is a stochastic estimate of $\nabla \mathcal{L}(\w_t)$ and  $\Am_t$ is a positive definite symmetric pre-conditioning matrix. In Adagrad, $\Am_{ada,t}$ is the un-centered second moment matrix of the past gradients computed as
\begin{equation}\label{eq:adagrad_ellips}
  \Am_{ada,t}:=\Gm_t \Gm_t^\intercal+ \epsilon \mathbf{I},
\end{equation}
where $\epsilon>0$, $\mathbf{I}$ is the $d\times d$ identity matrix and 
$\Gm_t=[\g_1,\g_2,\ldots,\g_t]$. Building up on the intuition that past gradients might become obsolete in quickly changing non-convex landscapes, RMSprop (and Adam) introduce an exponential weight decay leading to the preconditioning matrix
\begin{equation}\label{eq:rms_ellips}
    \Am_{rms,t}:=\left((1-\beta) \Gm_t\diag(\beta^t,\ldots, \beta^0)\Gm_t^\intercal\right)+\epsilon \mathbf{I},
\end{equation}
where $\beta \in (0,1)$. In order to save computational efforts, the diagonal versions $\diag(\Am_{ada})$ and $\diag(\Am_{rms})$ are more commonly applied in practice, which in turn gives rise to coordinate-wise adaptive stepsizes that are enlarged (reduced) in coordinates that have seen past gradient components with a smaller (larger) magnitude.

\subsection{Adaptive preconditioning as ellipsoidal Trust Region}
\begin{figure}
\centering          
          \begin{tabular}{c@{}c@{}c@{}}
            \adjincludegraphics[width=0.32\linewidth, trim={22pt 22pt 30pt 30pt},clip]{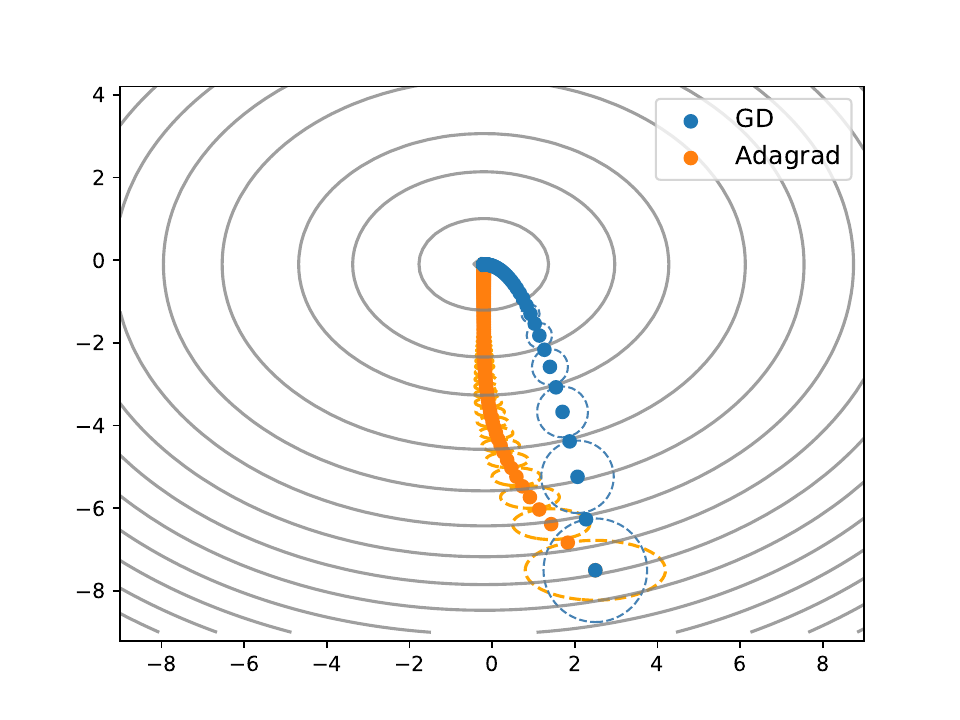} &
             \adjincludegraphics[width=0.32\linewidth, trim={22pt 22pt 30pt 30pt},clip]{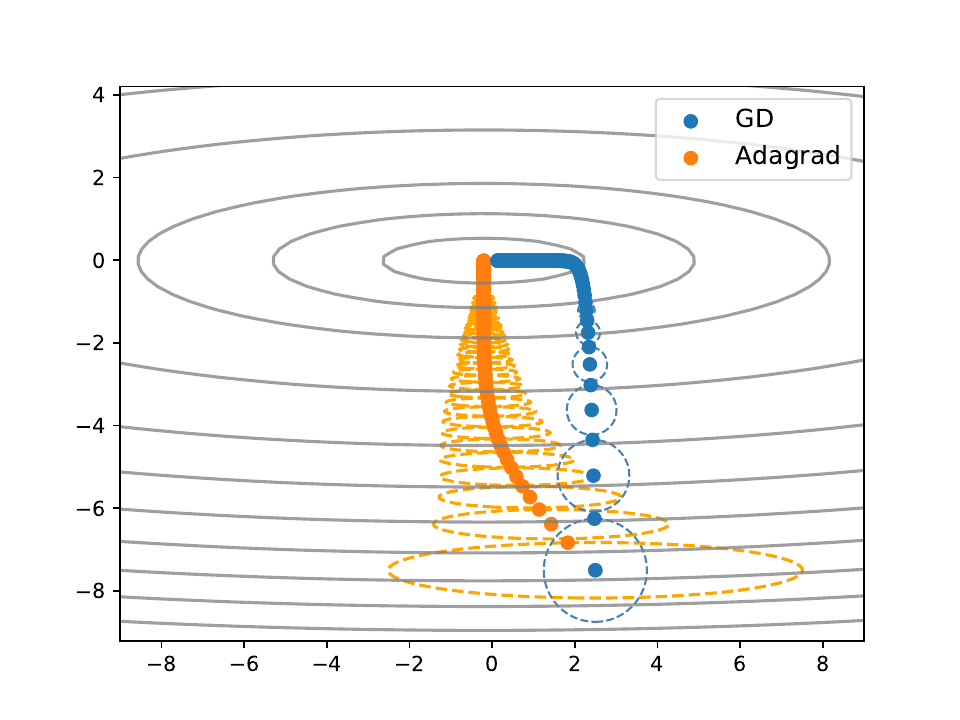} 
             &\adjincludegraphics[width=0.32\linewidth, trim={22pt 22pt 30pt 30pt},clip]{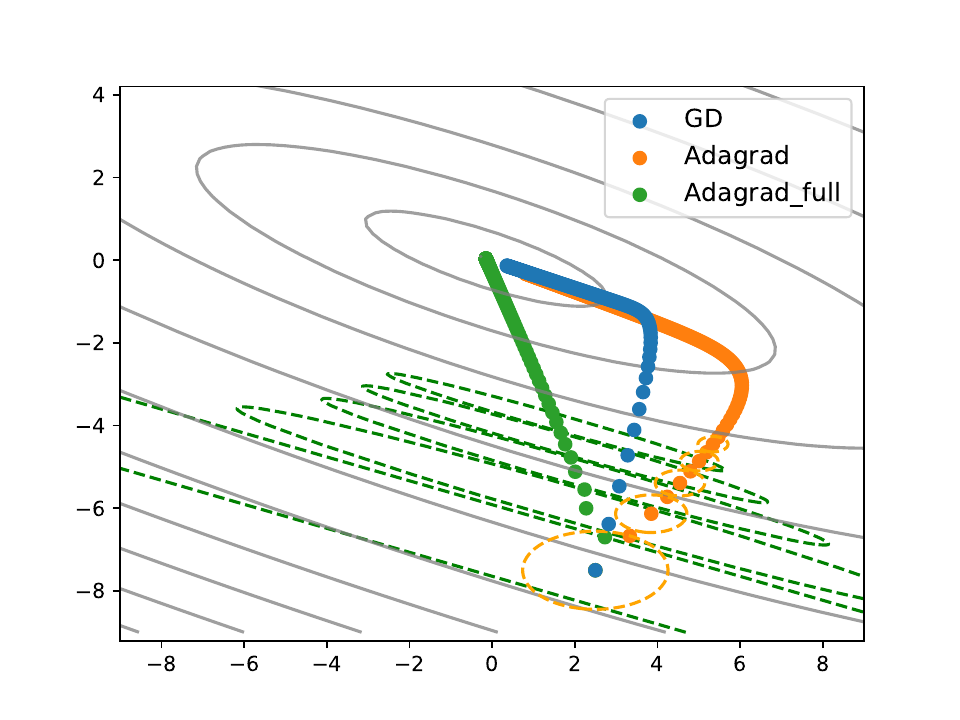} 
             \\  
             \adjincludegraphics[width=0.32\linewidth, trim={22pt 22pt 30pt 30pt},clip]{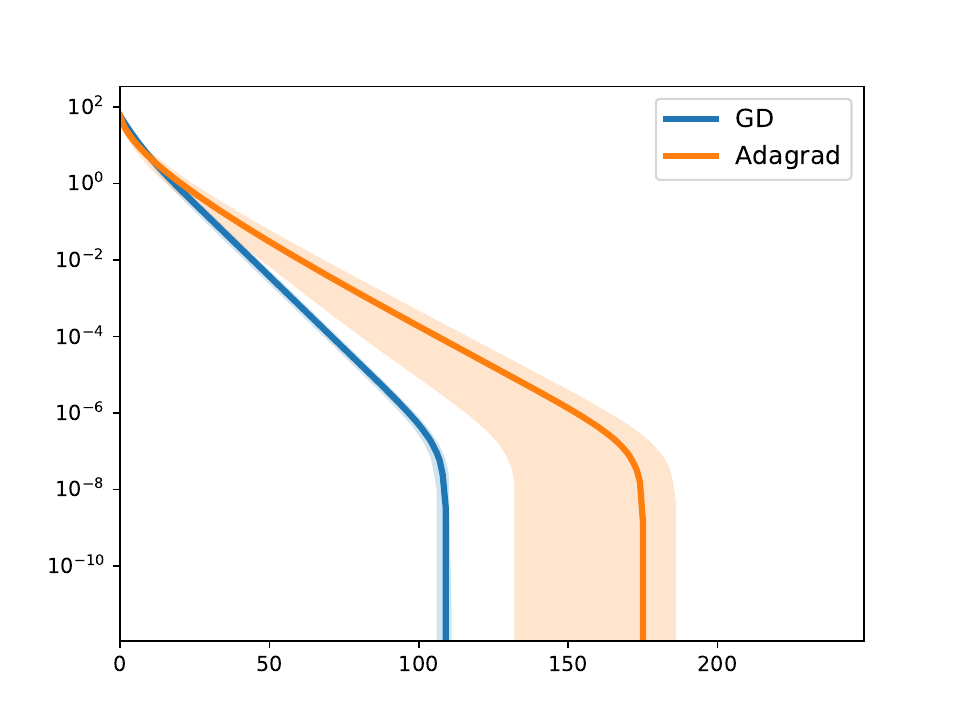} &
             \adjincludegraphics[width=0.32\linewidth, trim={22pt 22pt 30pt 30pt},clip]{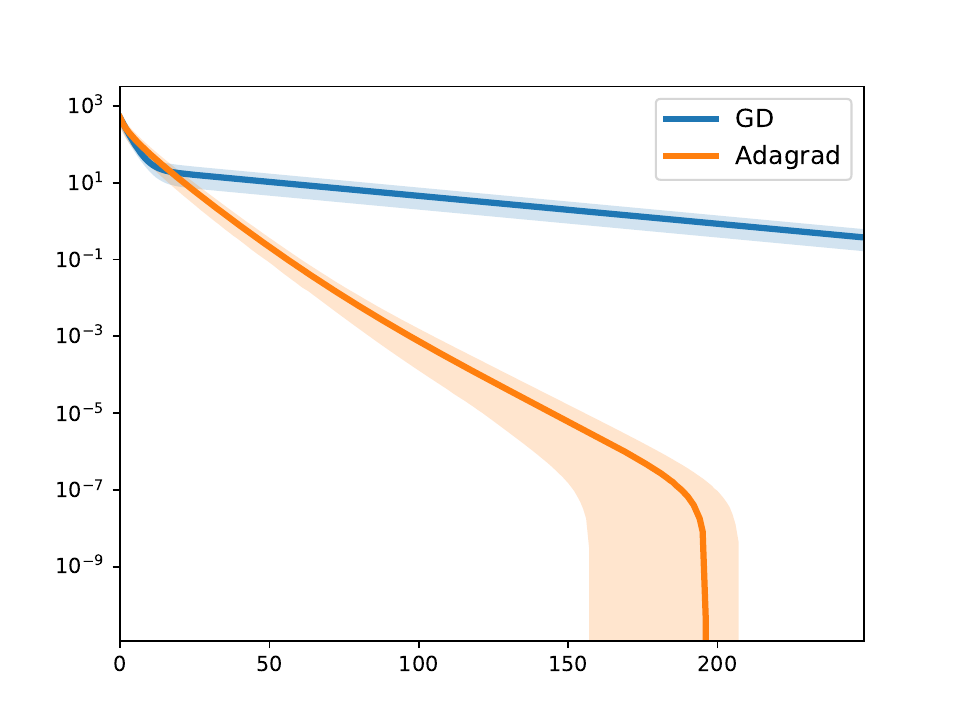}
             &

             \adjincludegraphics[width=0.32\linewidth, trim={22pt 22pt 30pt 30pt},clip]{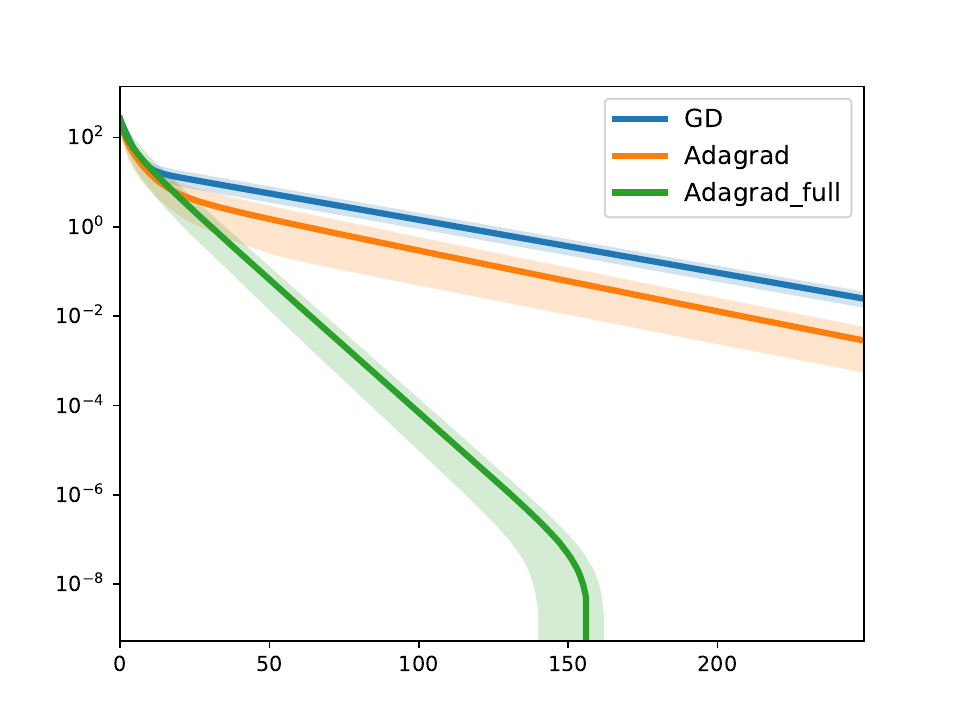}             \\
         	\footnotesize{{$\kappa=2$}}&
            \footnotesize{{$\kappa=20$}}&
            \footnotesize{{$\kappa=20$}}
	  \end{tabular}
          \caption{ \footnotesize{Top: Iterates and implicit trust regions of GD and Adagrad on quadratic objectives with different condition number $\kappa$. Bottom: Average log suboptimality over iterations as well as 90\% confidence intervals of 30 runs with random initialization}}
          \label{fig:quadratics}

\end{figure}

Starting from the fact that adaptive methods employ coordinate-wise stepsizes, one can take a principled view on these methods. Namely, their update steps arise from minimizing a first-order Taylor model of the function $\mathcal{L}$ within an \textit{ellipsoidal} search space around the current iterate $\w_t$, where the diameter of the ellipsoid along a particular coordinate is \textit{implicitly} given by $\eta_t$ and $\|\g_t\|_{\Am_t^{-1}}$. Correspondingly, vanilla (S)GD optimizes the same first-order model within a \textit{spherical} constraint. Figure~\ref{fig:quadratics} (top) illustrates this effect by showing not only the iterates of GD and Adagrad but also the implicit trust regions within which the local models were optimized at each step.\footnote{We only plot every other trust region. Since the models are linear, the minimizer is always on the boundary.} 

It is well known that GD struggles to progress towards the minimizer of quadratics along low-curvature directions (see e.g., \cite{goh2017why}). 
While this effect is negligible for well-conditioned objectives (Fig.~\ref{fig:quadratics}, left), it leads to a drastic slow-down when the problem is ill-conditioned (Fig.~\ref{fig:quadratics}, center). Particularly, once the method has reached the bottom of the valley, it struggles to make progress along the horizontal axis. Here is precisely where the advantage of adaptive stepsize methods comes into play. As illustrated by the dashed lines, Adagrad's search space is damped along the direction of high curvature (vertical axis) and elongated along the low curvature direction (horizontal axis). This allows the method to move further horizontally early on to enter the valley with a smaller distance to the optimizer $\w^*$ along the low curvature direction which accelerates convergence.

Let us now formally establish the result that allows us to re-interpret adaptive gradient methods from the trust region perspective introduced above.

\begin{mdframed}
\begin{lemma}[Preconditioned gradient methods as TR] \label{th:main}
A preconditioned gradient step
\begin{equation}\label{eq:precond_gd_step}
\w_{t+1}-\w_t=\s_{t}:=-\eta_t \Am_t^{-1} \g_t 
\end{equation}
with stepsize $\eta_t>0$, symmetric positive definite preconditioner $\Am_t\in \mathbb{R}^{d\times d}$ and $\g_t\not=0$ minimizes a first-order model around $\w_t\in\mathbb{R}^d $ in an ellipsoid given by $\Am_t$ in the sense that

\begin{equation}\label{eq:1st_TR}
\begin{aligned}
&\s_{t}:= \arg\min_{\s\in \mathbb{R}^d} \left[ m_t^1(\s)= \mathcal{L}(\w_t) + \s^\intercal \g_t\right],\:\text{s.t.}\quad \|\s\|_{\Am_t}  \leq \eta_t\|\g_t\|_{\Am_t^{-1}}.
\end{aligned}
\end{equation}
\end{lemma}
\end{mdframed}

\begin{corollary}[Rmsprop]
The step $\s_{rms,t}:=-\eta_t \Am_{rms,t}^{-1/2} \g_t $ minimizes a first-order Taylor model around $\w_t$ in an ellipsoid given by $\Am_{rms,t}^{1/2}$ (Eq.~\ref{eq:rms_ellips}) in the sense that

\begin{equation}
\begin{aligned}\label{eq:rms_step_tr}
&\s_{rms,t}:= \arg\min_{\s\in \mathbb{R}^d} \left[ m_t^1(\s)= \mathcal{L}(\w_t) + \s^\intercal \g_t\right],\: \text{s.t.}\quad\|\s\|_{\Am_{rms,t}^{1/2}} \leq \eta_t \|\g_t\|_{\Am_{rms,t}^{-1/2}}.
\end{aligned}
\end{equation}
\end{corollary}
Equivalent results can be established for Adam using $\g_{adam,t}:=(1-\beta)\sum_{k=0}^t \beta^{t-k}\g_t$ as well as for Adagrad by replacing the matrix $\Am_{ada}$ into the constraint in Eq.~(\ref{eq:rms_step_tr}). Of course, the update procedure in Eq.~(\ref{eq:1st_TR}) is merely a reinterpretation of the original preconditioned update, and thus the employed trust region radii are defined \textit{implicitly} by the current gradient and stepsize.
\subsection{Diagonal versus full preconditioning}
A closer look at Figure~\ref{fig:quadratics} reveals that the first two problems are perfectly \textit{axis-aligned}, which makes these objectives particularly attractive for diagonal preconditioning. For comparison, we report another quadratic instance, where the Hessian is no longer zero on the off-diagonals (Fig.~\ref{fig:quadratics}, right). As can be seen, this introduces a tilt in the level sets and reduces the superiority of diagonal Adagrad over plain GD. However, using the full preconditioner $\Am_{ada}$ re-establishes the original speed up. Yet, non-diagonal preconditioning comes at the cost of taking the inverse square root of a large matrix, which is why this approach has been relatively unexplored (see \cite{agarwal2018case} for an exception). Interestingly, early results by \cite{becker1988improving} on the curvature of neural nets report a strong diagonal dominance of the Hessian matrix $\nabla^2\mathcal{L}(\w)$. However, the reported numbers are only for tiny networks of at most 256 parameters. We here take a first step towards generalizing these findings to modern day networks. Furthermore, we contrast the diagonal dominance of real Hessians to the expected behavior of random Wigner matrices.\footnote{Of course, Hessians do not have i.i.d. entries but the symmetry of Wigner matrices suggests that this baseline is not completely off.} For further evidence, we also compare Hessians of Ordinary Least Squares (OLS) problems with random inputs. For this purpose, let $\delta_\Am$ define the ratio of diagonal to overall mass of a matrix $\Am$, i.e. $\delta_\Am:= \frac{ \sum_i |\Am_{i,i}| }{\sum_i\sum_j |\Am_{i,j}|}$ as in \citep{becker1988improving}.

\begin{proposition}[Diagonal share of Wigner matrix] \label{prop:wigner}
For a random Gaussian\footnote{The argument naturally extends to any distribution with positive expected absolute values.} Wigner matrix $\Wm$ (see Eq. (\ref{eq:wiggly_wigner})) the diagonal mass of the expected absolute matrix amounts to:
$    
\delta_{\E\left[|\Wm|\right]}=\frac{1}{1+(d-1)\frac{\sigma_2}{\sigma_1}}.
$
\end{proposition}

Thus, if we suppose the Hessian at any given point $\w$ were a random Wigner matrix we would expect the share of diagonal mass to fall with $\mathcal{O}(1/d)$ as the network grows in size. In the following, we derive a similar result for the large $n$ limit in the case of OLS Hessians.

\begin{proposition}[Diagonal share of OLS Hessian]\label{prop:ols}
Let $\Xm\in \mathbb{R}^{d\times n}$ and assume each $\x_{i,j}$ is generated i.i.d. with zero-mean and finite second moment $\sigma^2>0$. Then the share of diagonal mass of the expected matrix  $\E\left[|\Hm_{\text{ols}}|\right]$ amounts to:
$
   \delta_{\E\left[|\Hm_{\text{ols}}|\right]} \overset{n \rightarrow \infty}{\rightarrow} \frac{\sqrt{n}}{\sqrt{n}+(d-1)\sqrt{\frac{2}{\pi}}}.
$
\end{proposition}
Empirical simulations suggest that this result holds already in small $n$ settings (see Figure~\ref{fig:OLS}) and finite $n$ results can be likely derived under assumptions such as Gaussian data.
As can be seen in Figure~\ref{fig:diag_dominance} below, even for a practical batch size of $n=32$ the diagonal mass $\delta_\Hm$ of neural networks stays above both benchmarks for random inputs as well as with real-world data.

\begin{figure}[H]
    \centering
\begin{tabular}{cc}
  \includegraphics[width=0.45\linewidth,valign=c]{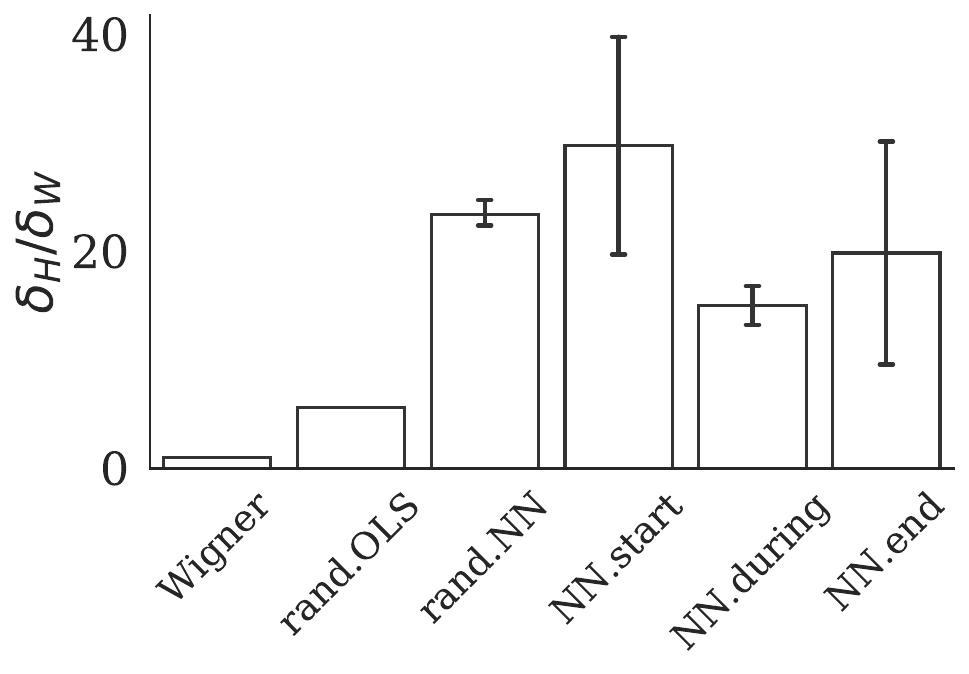} &   \includegraphics[width=0.45\linewidth,valign=c]{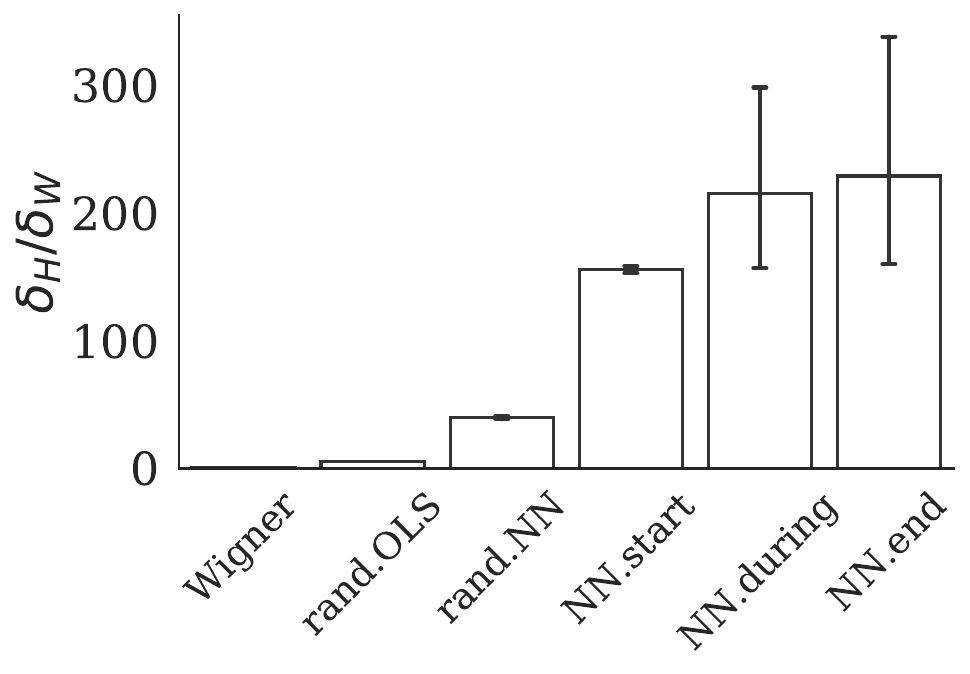}\\
  {\quad \small Simple CNN (62k weights)} & { \quad \small MLP (411k weights)}
\end{tabular}
\caption{ \footnotesize{Diagonal mass of neural network Hessian $\delta_\Hm$ relative to $\delta_{\E\left[|\Wm|\right]}$ and $\delta_{\E\left[|\Hm_{\text{ols}}|\right]}$ of corresponding dimensionality for random inputs as well as at random initialization, middle and after reaching $90\%$ training accuracy with RMSProp on CIFAR-10. Mean and 95\% confidence interval over 10 independent runs.}}\label{fig:diag_dominance}

\end{figure}

 These results are in line with \cite{becker1988improving} and suggest that full matrix preconditioning might indeed not be worth the additional computational cost. Consequently, we use diagonal preconditioning for both first- and second-order methods in all of our experiments in Section~\ref{sec:EXP}. Further theoretical elaborations of these findings present an interesting direction of future research.

%% file: 04_2nd_order.tex
\section{Second-order Trust Region Methods}\label{sec:ETR}

Cubic regularization~\citep{nesterov2006cubic,cartis2011adaptive} and trust region methods belong to the family of globalized Newton methods. Both frameworks compute parameter updates by optimizing regularized (former) or constrained (latter) second-order Taylor models of the objective $\mathcal{L}$ around the current iterate $\w_t$.\footnote{In the following we only treat TR methods, but we emphasize that the use of matrix induced norms can directly be transferred to the cubic regularization framework.} In particular, in iteration $t$ the update step of the trust region algorithm is computed as
\begin{equation}\label{eq:tr_step}
\begin{aligned}
&\min_{\s \in \mathbb{R}^d}\: \left[ m_t(\s) := \mathcal{L}(\w_t)+\g_t^\intercal \s + \frac{1}{2} \s^\intercal \Bm_t \s \right], \: \text{s.t.}\:\:\|\s\|_{\Am_t}\leq \Delta_t,
\end{aligned}
\end{equation}
where $\Delta_t>0$ and $\g_t$ and $\Bm_t$ are either $\nabla \mathcal{L}(\w_t)$ and $\nabla^2\mathcal{L}(\w_t)$ or suitable approximations. The matrix $\Am_t$ induces the shape of the constraint set. So far, the common choice for neural networks is $\Am_t:=\mathbf{I},\;\forall t$ which gives rise to spherical trust regions \citep{xu2017second,liu2018stochastic}. By solving the \textit{constrained} problem (\ref{eq:tr_step}), TR methods overcome the problem that pure Newton steps may be ascending, attracted by saddles or not even computable. 
Please see Appendix B for more details.

\paragraph{Why ellipsoids?} 
There are many sources for ill-conditioning in neural networks such as un-centered and correlated inputs \citep{lecun2012efficient}, saturated hidden units, and different weight scales in different layers~\citep{van1998solving}. While the quadratic term of model (\ref{eq:tr_step}) accounts for such ill-conditioning to some extent, the spherical constraint is completely blind towards the loss surface. Thus, it is advisable to instead measure distances in norms that reflect the underlying geometry (see Chap. 7.7 in \cite{conn2000trust}). The ellipsoids we propose are such that they allow for longer steps along coordinates that have seen small gradient components in the past and vice versa. Thereby the TR shape is adaptively adjusted to fit the current region of the loss landscape. This is not only effective when the iterates are in an ill-conditioned neighborhood of a minimizer (Fig.~\ref{fig:quadratics}), but it also helps to escape elongated plateaus (see autoencoder in Sec.~\ref{sec:EXP}). Contrary to adaptive first-order methods, the diameter ($\Delta_t$) is updated directly depending on whether or not the local Taylor model is an adequate approximation at the current point. 
\subsection{Convergence of ellipsoidal Trust Region methods}
Inspired by the success of adaptive gradient methods, we investigate the use of their preconditioning matrices as norm inducing matrices for second-order TR methods. The crucial condition for convergence is that the applied norms are not degenerate during the entire minimization process in the sense that the ellipsoids do not flatten out (or blow up) completely along any given direction. The following definition formalizes this intuition.
\begin{definition}[Uniformly equivalent norms]\label{a:uniform_equi}
The norms $\|\w\|_{\Am_t}:= \left(\w^\intercal \Am_t \w\right)^{1/2}$ induced by symmetric positive definite matrices $\Am_t$ are called uniformly equivalent, if $\exists \mu\geq 1$ such that $ \forall \w \in \mathbb{R}^d,\forall t=1,2,\ldots$
\vspace{-2mm}
\begin{equation}  \label{eq:uniform_equi}
\begin{aligned}
&\frac{1}{\mu} \|\w\|_{\Am_t} \leq \|\w\|_2 \leq \mu\|\w\|_{\Am_t}.
\end{aligned}\end{equation}
\end{definition}

We now establish a result which shows that the RMSProp ellipsoid is indeed uniformly equivalent. 
\begin{mdframed}
\begin{lemma}[Uniform equivalence] \label{T:uniform_equivalence}
Suppose $\|\g_t\|^2 \leq L_{H}^2$ for all $\w_t \in \mathbb{R}^d,$ $t=1,2,\ldots$ Then there always exists $\epsilon>0$ such that the proposed preconditioning matrices $\Am_{rms,t}$ (Eq.~\ref{eq:rms_ellips}) are uniformly equivalent, i.e. Def.~\ref{a:uniform_equi} holds. The same holds for the diagonal variant.
\end{lemma}
\end{mdframed}

Consequently, the ellipsoids $\Am_{rms,t}$ can directly be applied to any convergent TR framework without losing the guarantee of convergence (\cite{conn2000trust}, Theorem 6.6.8).\footnote{Note that the assumption of bounded batch gradients, i.e. smooth objectives, is common in the analysis of stochastic algorithms \citep{allen2017katyusha,defazio2014saga,schmidt2017minimizing,duchi2011adaptive}.} In Theorem \ref{th:rate} we extend this result by showing the (to the best of our knowledge) first convergence \textit{rate} for ellipsoidal TR methods. Interestingly, similar results cannot be established for $\Am_{ada,t}$, which reflects the widely known vanishing stepsize problem that arises since squared gradients are continuously added to the preconditioning matrix. At least partially, this effect inspired the development of RMSprop~\citep{tieleman2012lecture} and Adadelta~\citep{zeiler2012adadelta}.

\subsection{A stochastic ellipsoidal TR framework for neural network training}\label{sec:TR_applications_NN}

Since neural network training often constitutes a large-scale learning problem in which the number of datapoints $n$ is high, we here opt for a stochastic TR framework in order to circumvent memory issues and reduce the computational complexity. To obtain convergence without computing full derivative information, we first need to assume sufficiently accurate gradient and Hessian estimates.


\begin{assumption}[Sufficiently accurate derivatives]\label{a:sampling}
The approximations of the gradient and Hessian at step $t$ satisfy
\begin{equation*}
\begin{aligned}
   \|\g_t-\nabla \mathcal{L}(\w_t)\| \leq \delta_g \: \text{   and   } \:
   \|\Bm_t-\nabla^2  \mathcal{L}(\w_t)\| \leq \delta_H,
\end{aligned}
\end{equation*}
where $\delta_g\leq \frac{(1-\eta)\epsilon_g}{4}$ and $\delta_H\leq \min\left\{\frac{(1-\eta)v\epsilon_H}{2},1\right\}$, for some $0<v<1$.
\end{assumption}
For finite-sum objectives such as Eq.~(\ref{eq:f_x}), the above condition can be met by random sub-sampling due to classical concentration results for sums of random variables \citep{xu2017newton,kohler2017sub,tripuraneni2017stochastic}. Following these references, we assume access to the full function value in each iteration for our theoretical analysis but we note that convergence can be retained even for fully stochastic trust region methods \citep{gratton2017complexity,chen2018stochastic,blanchet2016convergence} and indeed our experiments in Section \ref{sec:EXP} use sub-sampled function values due to memory constraints. Secondly, we adapt the framework of \cite{yao2018inexact,xu2017newton}, which allows for cheap inexact subproblem minimization, to the case of iteration-dependent constraint norms (Alg. \ref{alg:src}). 

 \begin{algorithm}[]
   \caption{Stochastic Ellipsoidal Trust Region Method}
   \label{alg:src}
\begin{algorithmic}[1]
   \STATE {\bfseries Input:}  
 $\w_0 \in \R^d$,
   $ \gamma >1, 1>\eta>0$,  $\Delta_0>0$
   \FOR{$t=0,1,\dots,\text{until convergence}$}
   \STATE Compute approximations $\g_t$ and $\Bm_t$. 
   \STATE \textbf{If} $\|\g_t\|\leq \epsilon_g$, set $\g_t:=0$.
   \STATE Set $\Am_t:=\Am_{rms,t}$ or $\Am_t:=\text{diag}\left(\Am_{rms,t}\right)$  (see Eq.~(\ref{eq:rms_ellips})).
   \STATE Obtain $\s_t$ by solving $m_t(\s_t)$ approximately.
   \STATE Compute ratio of function over model decrease:
    $
   \: \rho_{t}=\dfrac{\mathcal{L}(\w_t)-\mathcal{L}(\w_t + \s_t)}{m_t(\zero)-m_t(\s_t)}
    $
    \STATE Set
    \begin{equation*} \label{eq:sigma_update}
    \begin{aligned}
        \Delta_{t+1} &= \begin{cases}
    \gamma\Delta_t & \text{ if } \rho_{\mathcal{S},t}\geq\eta \\
    \Delta_t/\gamma & \text{ if } \rho_{\mathcal{S},t}<\eta
    \end{cases}, \:|: \text{ and } \: 
    \w_{t+1} &= \begin{cases}
    \w_{t}+\s_t & \text{ if } \rho_{t} \geq \eta \quad \;\text{ (successful)}\\
    \w_{t} & \text{ otherwise}\quad \text{ (unsuccessful).}
    \end{cases}    
    \end{aligned}
    \end{equation*}
 \ENDFOR
\end{algorithmic}
\end{algorithm}

\begin{assumption}[Approximate model minimization]\label{as:model_decrease}
Each update step $\s_t$ yields at least as much model decrease as the Cauchy- and Eigenpoint simultaneously, i.e.$
    m_t(\s_t)\leq m_t(\s_t^C)$ and $m_t(\s_t)\leq m_t(\s_t^E),
$
where $\s_t^C$ and $\s_t^E$ are defined in Eq.(\ref{eq:cauchy_and_eigenpoint}). 
\end{assumption}

 Finally, given that the adaptive norms induced by $\Am_{rms,t}$ satisfy uniform equivalence as shown in Lemma \ref{th:main}, the following Theorem establishes an ${\cal O}\left(\text{max}\left\{\epsilon_g^{-2}\epsilon_H^{-1},\epsilon_H^{-3}\right\}\right)$ worst-case iteration complexity which effectively matches the one of \cite{yao2018inexact}.
\begin{framed}
\begin{theorem}[Convergence rate of Algorithm \ref{alg:src}]\label{th:rate}
Assume that $\mathcal{L(\w)}$ is second-order smooth with Lipschitz constants $L_g$ and $L_H$.  Furthermore, let Assumption \ref{a:sampling} and \ref{as:model_decrease} hold. Then Algorithm 1 finds an $\mathcal{O}(\epsilon_g,\epsilon_H)$ first- and second-order stationary point in at most $\mathcal{O}\left(\max\left\{\epsilon_g^{-2}\epsilon_H^{-1},\epsilon_H^{-3}\right\}\right)$ iterations.
\end{theorem}
\end{framed}

\begin{figure*}[h!]
 \centering 
          \begin{tabular}{c@{}c@{}c@{}c@{}}
        &   \footnotesize{ResNet18} & \footnotesize{MLP} & \footnotesize{Autoencoder} \\
        \rotatebox[origin=c]{90}{\footnotesize{Fashion-MNIST}} &     \includegraphics[width=0.29\linewidth,valign=c]{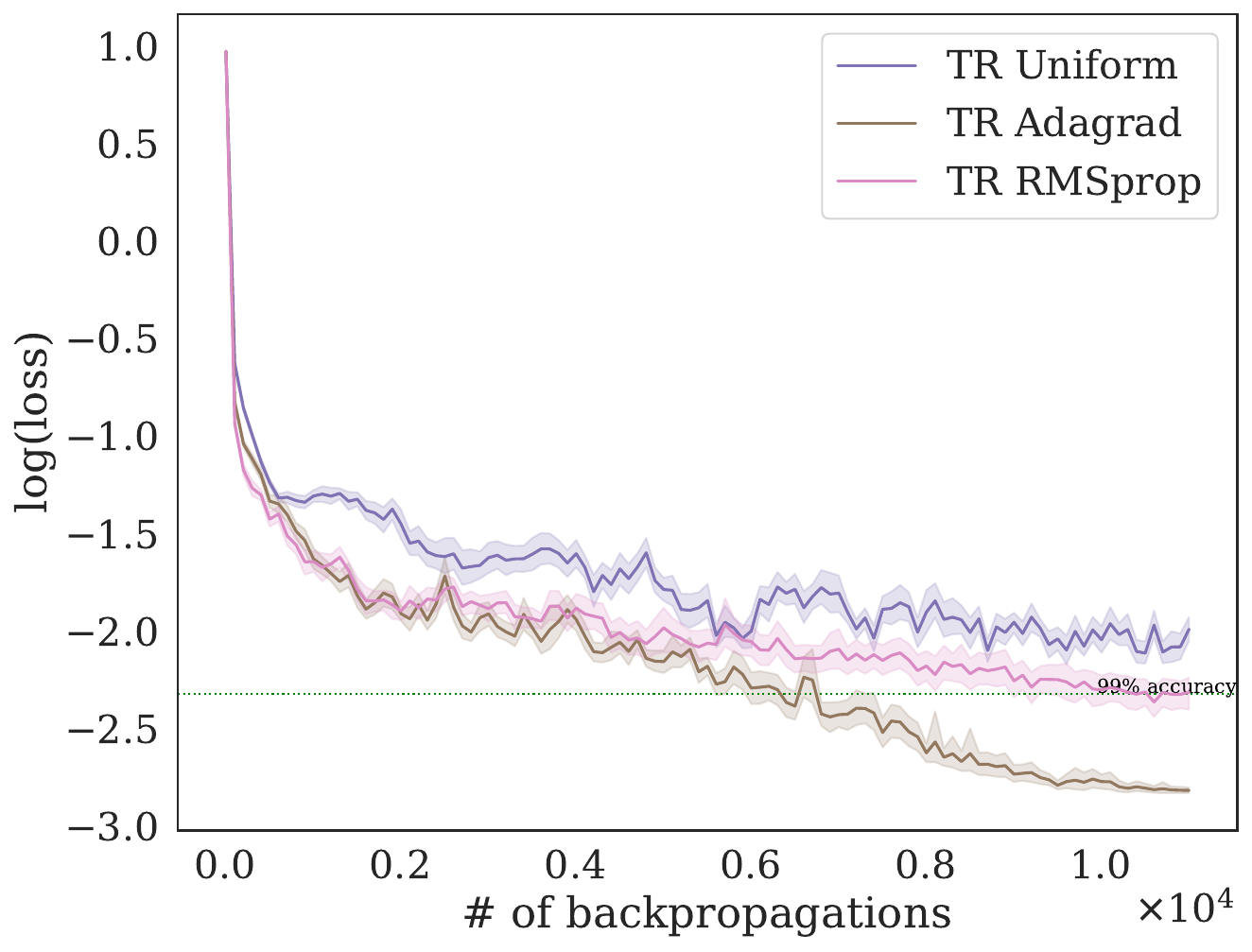}& 
             \includegraphics[width=0.29\linewidth,valign=c]{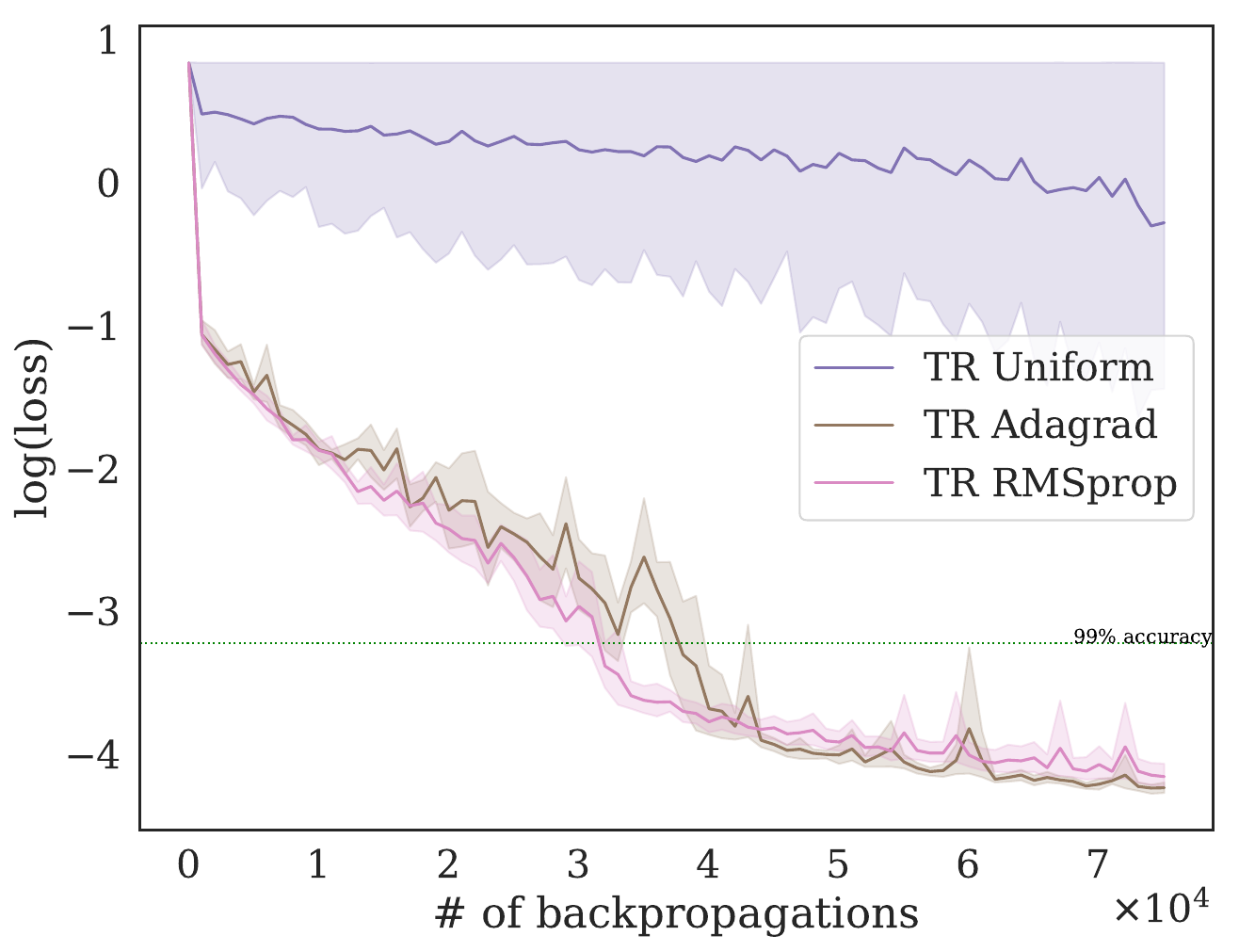} & \includegraphics[width=0.29\linewidth,valign=c]{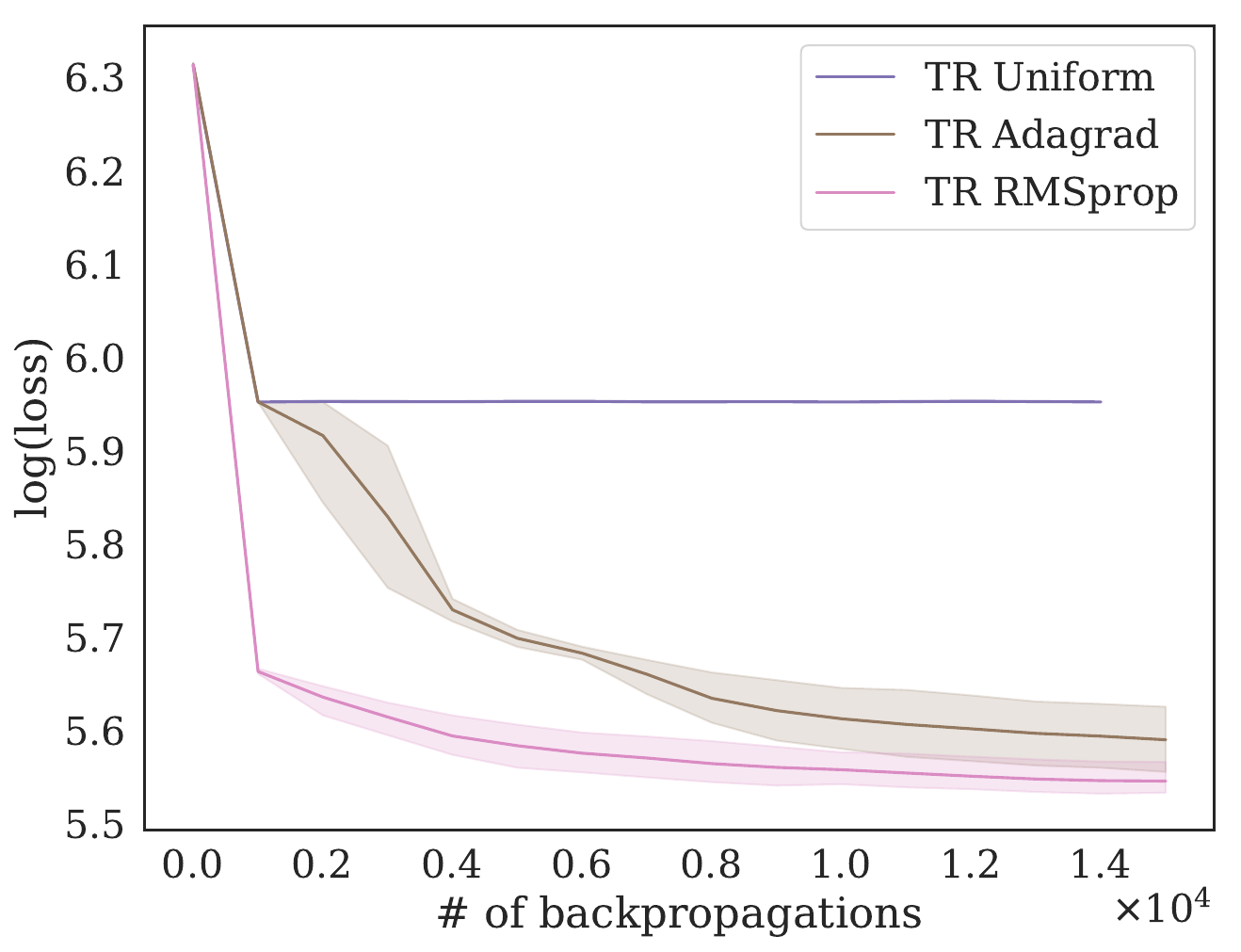}\\
             
             \rotatebox[origin=c]{90}{\footnotesize{CIFAR-10}}&
             \includegraphics[width=0.29\linewidth,valign=c]{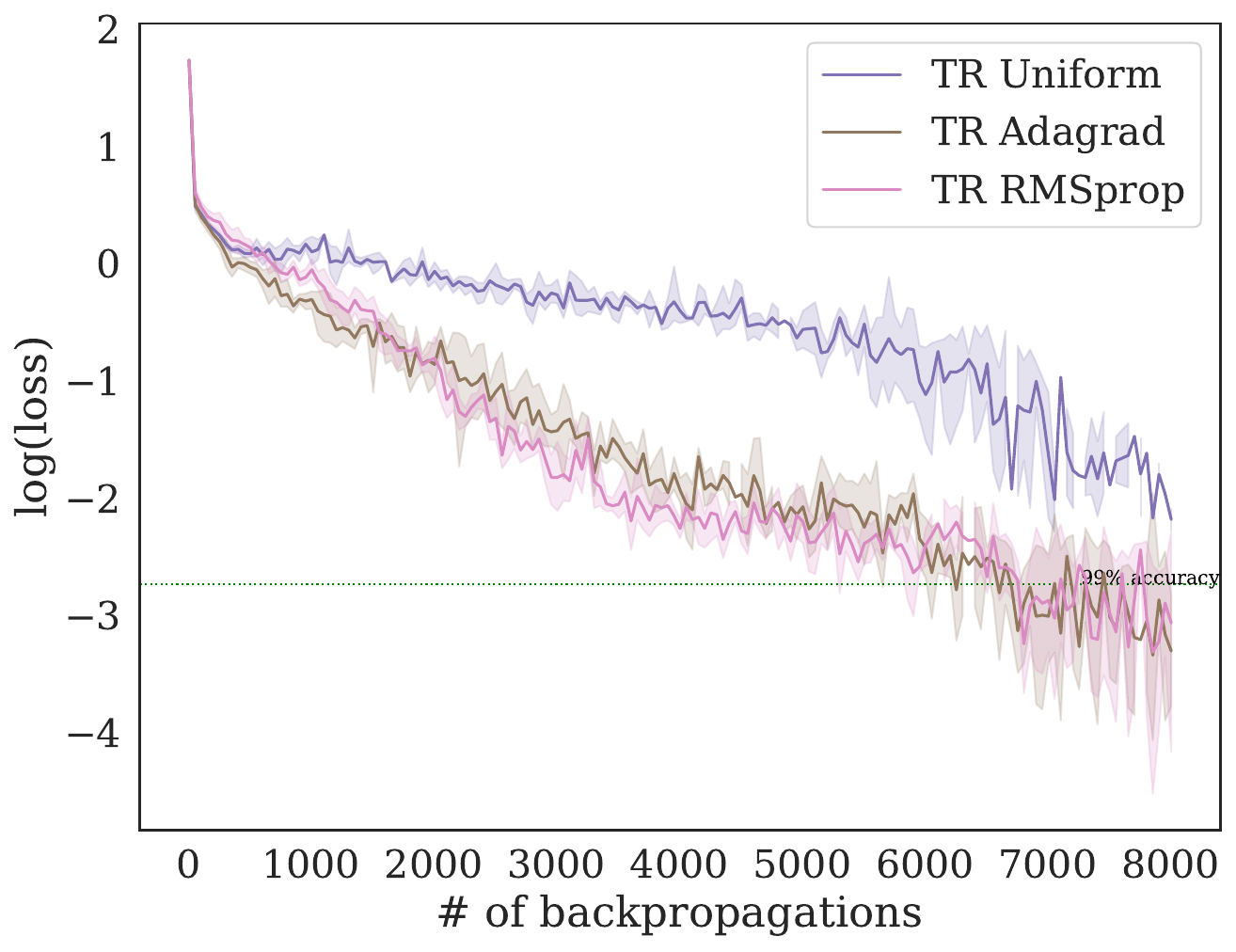}&
             \includegraphics[width=0.29\linewidth,valign=c]{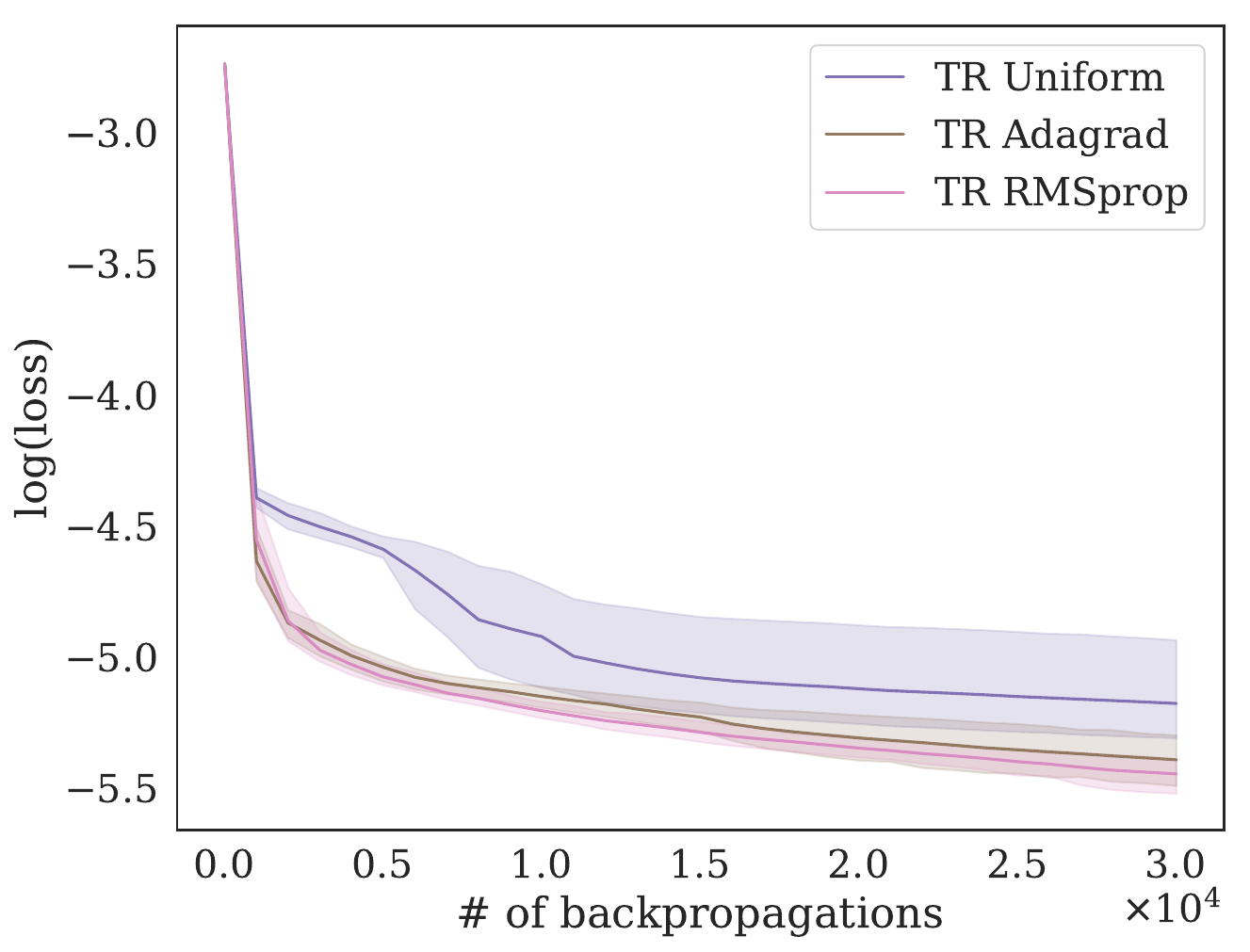} & \includegraphics[width=0.29\linewidth,valign=c]{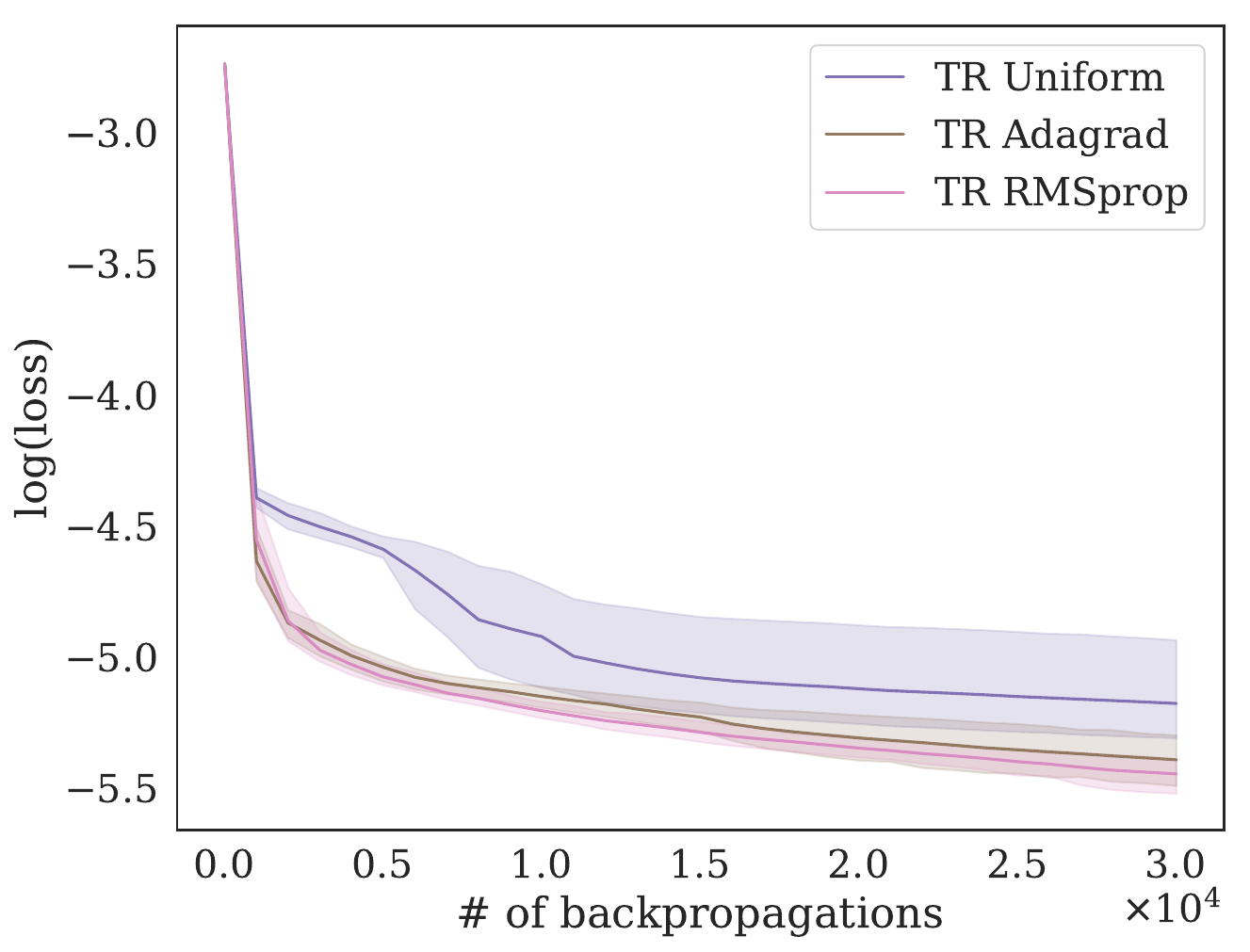} 
             
 	  \end{tabular}
          \caption{ \footnotesize{Log loss over backpropagations. Mean and $95\%$ confidence interval of 10 runs. Green dotted line indicates $99\%$ training accuracy.}}
          \label{fig:2ndorder_results}
\end{figure*}

The proof of this statement is a straight-forward adaption of the proof for spherical constraints, taking into account that the guaranteed model decrease changes when the computed step $s_t$ lies \textit{outside} the Trust Region. Due to the uniform equivalence established in \ref{th:main}, the altered diameter of the trust region along that direction and hence the change factor is always strictly positive and finite.

%% file: 06_experiments.tex

\section{Experiments}\label{sec:EXP}
To validate our claim that ellipsoidal TR methods yield improved performance over spherical ones, we run a set of experiments on two image datasets and three types of network architectures. All methods run on (almost) the same hyperparameters across all experiments (see Table \ref{tab:params} in Appendix B) and employ the preconditionied Steihaug-Toint CG method \citep{steihaug1983conjugate} to solve the subproblems (Eq.~\ref{eq:quadratic_model_app}) with the classical stopping criterion given in Eq.~(\ref{eq:stopping_criterion}).

As depicted in Fig.~\ref{fig:2ndorder_results}, the ellipsoidal TR methods consistently outperform their spherical counterpart in the sense that they reach full training accuracy substantially faster on all problems. Moreover, their limit points are in all cases lower than those of the uniform method. Interestingly, this makes an actual difference in the image reconstruction quality of autoencoders (see Figure~\ref{fig:reconstruct}), where the spherically constrained TR method struggles to escape a saddle. We thus draw the clear conclusion that the ellipsoidal constraints we propose are to be preferred over spherical ones when training neural nets with second-order methods. More experimental and architectural details are provided in App. C.

To put the previous results into context, we also benchmark several state-of-the-art gradient methods. For a fair comparison, we report results in terms of number of backpropagations, epochs and time. All figures can be found in App. C. Our findings are mixed: For small nets such as the MLPs the TR method with RMSProp ellipsoids is superior in all metrics, even when benchmarked in terms of time. However, while Fig. \ref{fig:1storder_results_bp} indicates that ellipsoidal TR methods are slightly superior in terms of backpropagations even for ResNets and Autoencoders, a close look at the Figures \ref{fig:1storder_results_ep} and \ref{fig:1storder_results_time} (App. C) reveals that they at best manage to keep pace with first-order methods in terms of epochs and are inferior in time. Furthermore, only the autoencoders give rise to saddles, which adaptive gradient methods escape faster than vanilla SGD, similarly to the case for second-order methods in Fig. \ref{fig:2ndorder_results}.

%% file: 07_conclusion.tex
\section{Conclusion}
We investigated the use of ellipsoidal trust region constraints for neural networks. We have shown that the RMSProp matrix satisfies the necessary conditions for convergence and our experimental results demonstrate that ellipsoidal TR methods outperform their spherical counterparts significantly. We thus consider the development of further ellipsoids that can potentially adapt even better to the loss landscape such as e.g. (block-) diagonal hessian approximations (e.g.~\cite{bekas2007estimator}) or approximations of higher order derivatives as an interesting direction of future research.

Yet, the gradient method benchmark indicates that the value of Hessian information for neural network training is limited for mainly three reasons: 1) second-order methods rarely yield better limit points, which suggests that saddles and spurious local minima are not a major obstacle; 2) gradient methods can run on smaller batch sizes which is beneficial in terms of epoch and when memory is limited; 3) The per-iteration time complexity is noticeably lower for first-order methods (Figure \ref{fig:1storder_results_time}). These observations suggest that advances in hardware and distributed second-order algorithms (e.g., ~\cite{osawa2018second,dunner2018distributed}) will be needed before Newton-type methods can replace gradient methods in deep learning.

As a side note, we reported elevated levels of diagonal dominance in neural network architectures, which may partially explain the success of \textit{diagonal} preconditioning in first-order method. Further empirical and theoretical investigations of this phenomenon with a particular focus on layer-wise dependencies constitute an interesting direction of future research as for example algorithms such as K-FAC \cite{grosse2016kronecker} seem to achieve good results with \textit{block-diagonal} preconditioning.

\section*{Broader impact}

We consider our work fundamental research with no specific application other than training neural networks in general. 
Hence a broader impact discussion is not applicable.

%% file: 08_appendix.tex
\onecolumn

\appendix

\part*{Appendix A: Proofs}

\section{Notation}
Throughout this work, scalars are denoted by regular lower case letters, vectors by bold lower case letters and matrices as well as tensors by bold upper case letters. By $\|\cdot\|$ we denote an arbitrary norm. For a symmetric positive definite matrix $\Am$ we introduce the compact notation $\|\w\|_\Am = \left( \w^\intercal \Am \w \right)^{1/2}$, where $\w \in \mathbb{R}^d$.

\section{Equivalence of Preconditioned Gradient Descent and first-order Trust Region Methods}
\begin{figure}[h!]
\centering          
          \begin{tabular}{c@{}c@{}}
            \adjincludegraphics[width=0.4\linewidth, trim={22pt 22pt 30pt 30pt},clip]{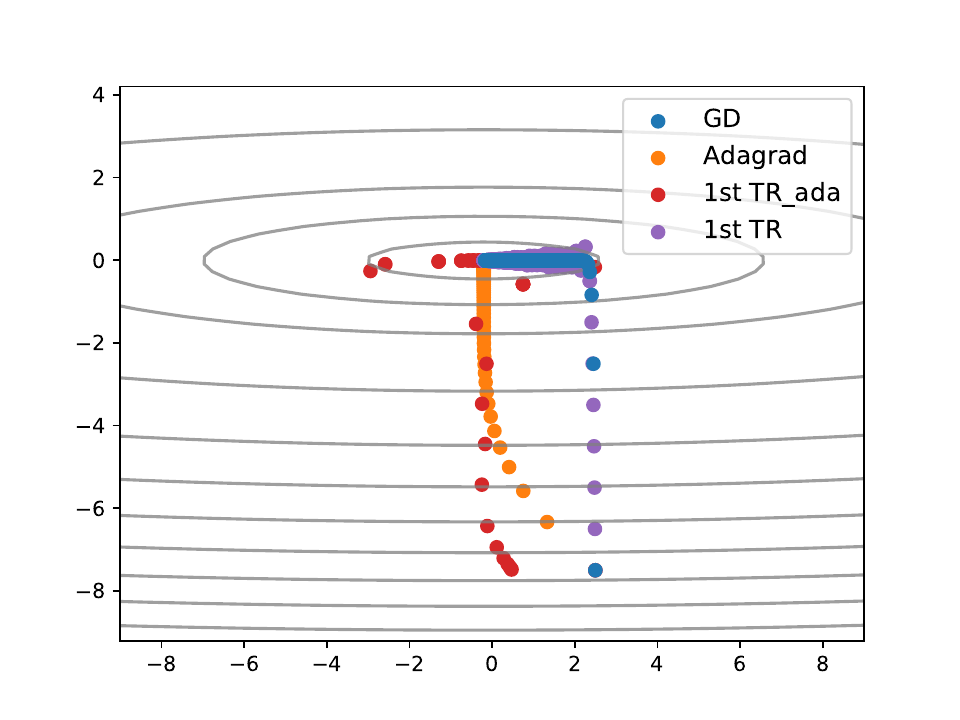} &
             \adjincludegraphics[width=0.4\linewidth, trim={22pt 22pt 30pt 30pt},clip]{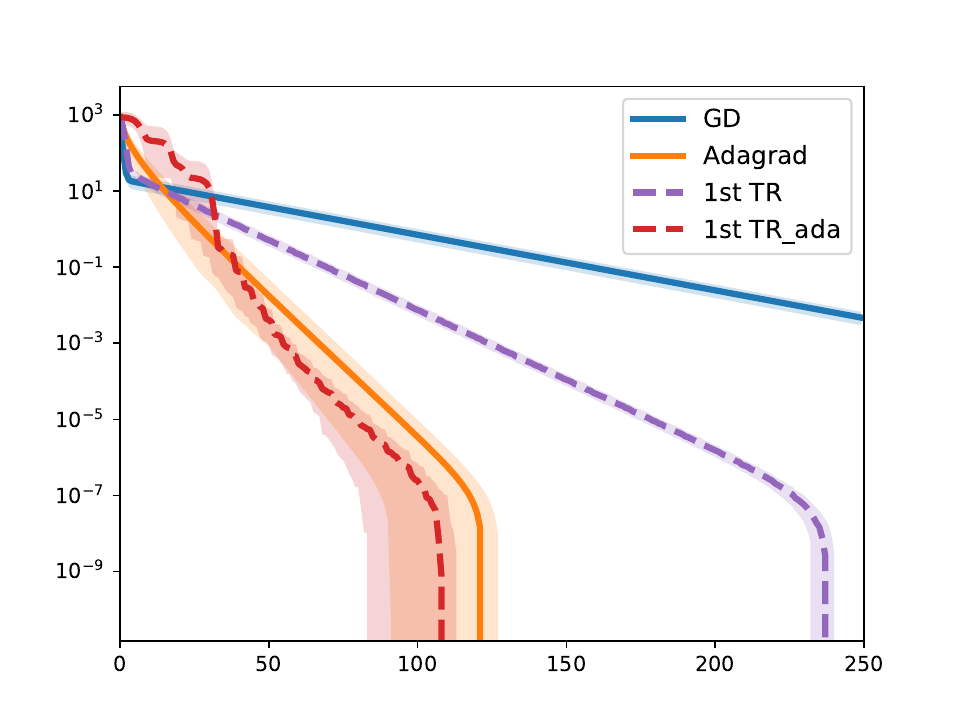}
	  \end{tabular}
          \caption{ \footnotesize{Iterates (left) and log suboptimality (right) of GD, Adagrad and two full-featured first-order TR algorithms of which one (1st TR) is spherically constraint and the other (1st TR$_{\text{ada}}$) uses $\Am_{ada}$ as ellispoid.}}
          \label{fig:1storderTRMs}
\end{figure}

\begin{mdframed}
\begin{theorem}[Theorem 1 restated] 
A preconditioned gradient step
\begin{equation}\label{eq:precond_gd_step_1st_TR_appendix}
\w_{t+1}-\w_t=\s_{t}:=-\eta_t \Am_t^{-1} \g_t 
\end{equation}
with stepsize $\eta_t>0$, symmetric positive definite preconditioner $\Am_t\in \mathbb{R}^{d\times d}$ and $\g_t\not=0$ minimizes a first-order local model around $\w_t\in\mathbb{R} $ in an ellipsoid given by $\Am_t$ in the sense that

\begin{equation}\label{eq:1st_TR_appendix}
\begin{aligned}
\s_{t}:= &\arg\min_{\s\in \mathbb{R}^d} \left[ m_t^1(\s)= \mathcal{L}(\w_t) + \s^\intercal \g_t\right], \quad \text{s.t.}\;\|\s\|_{\Am_t}  \leq \eta_t\|\g_t\|_{\Am_t^{-1}}.
\end{aligned}
\end{equation}
\end{theorem}
\end{mdframed}
\begin{proof}

We start the proof by noting that the optimization problem in Eq. (\ref{eq:1st_TR_appendix}) is convex. For $\eta_t> 0$ the constraint satisfies the Slater condition since $0$ is a strictly feasible point. As a result, any KKT point is a feasible minimizer and vice versa.

Let $L(\s,\lambda)$ denote the Lagrange dual of Eq.~(\ref{eq:1st_TR})
\begin{equation}
L(\s,\lambda):=  \mathcal{L}(\w_t) + \s^\intercal \g_t+\lambda \left(\|\s\|_\Am-\eta_t\|\g_t\|_{\Am_t^{-1}}\right). \end{equation}

Any point $\s$ is a KKT point if and only if the following system of equations is satisfied

    \begin{align}
        \nabla_{\s} L(\s,\lambda)=\g_t+\frac{\lambda}{\|\s\|_{\Am_t}}\Am_t\s=0\label{eq:kkt1}\\
        \lambda \left( \|\s\|_{\Am_t}-\eta_t\|\g_t\|_{\Am_t^{-1}}\right)=0\label{eq:kkt2}.\\
        \|\s\|_{\Am_t}-\eta_t\|\g_t\|_{\Am_t^{-1}}\leq 0 \label{eq:kkt3} \\
        \lambda \geq 0 \label{eq:kkt4}.
    \end{align}

For $\s_{t}$ as given in Eq. (\ref{eq:precond_gd_step})  we have that
\begin{equation}
\|\s_t\|_{\Am_t}=\sqrt{\eta_t^2 \g_t{(\Am_t^{-1})}^\intercal \Am_t\Am_t^{-1}\g_t}=\eta_t\sqrt{\g_t\Am_t^{-1}\g_t}=\eta_t\|\g_t\|_{\Am_t^{-1}}.
\end{equation}

and thus \ref{eq:kkt2} and \ref{eq:kkt3} hold with equality such that any $\lambda\geq0$ is feasible. Furthermore, 
 \begin{equation}
 \begin{aligned}
    \nabla_{\s} L(\s_t,\lambda)&=\nabla f(\w_t)+\frac{\lambda}{\|\s_t\|_{\Am_t}}\Am_t\s_t&\overset{(\ref{eq:precond_gd_step})}{=}\g_t-\eta_t\frac{\lambda}{\eta_t\|\g_t\|_{\Am_t^{-1}}}\Am_t\Am_t^{-1} \g_t
    &=\g_t-\frac{\lambda}{\|\g_t\|_{\Am_t^{-1}}} \g_t
 \end{aligned}
 \end{equation}
 is zero for $\lambda=\|\g_t\|_{\Am^{-1}} \geq 0$. As a result, $\s_t$ is a KKT point of the convex problem \ref{eq:1st_TR} which proves the assertion.

\end{proof}
To illustrate this theoretical result we run gradient descent and Adagrad as well as the two corresponding first-order TR methods\footnote{Essentially Algorithm \ref{alg:src} with $m_t$ based on a \textit{first} order Taylor expansion, i.e. $m_t^1(\s)$ as in Eq.~(\ref{eq:1st_TR_appendix}).} on an ill-conditioned quadratic problem. While the method 1st TR optimizes a linear model within a ball in each iteration, 1st TR$_{\text{ada}}$ optimizes the same model over the ellipsoid given by the Adagrad matrix $\Am_{ada}$. The results in Figure \ref{fig:1storderTRMs} show that the methods behave very similarly to their constant stepsize analogues.

\section{Convergence of ellipsoidal TR methods}


\subsection{Proof sketch}

At a high level, the proof can be divided into two steps: 1) establish that each update decreases the model value and 2) relate the model decrease to the function decrease, therefore proving that the function decreases.

Based on Assumption \ref{as:model_decrease} the proof first relates the \textit{model decrease} in each iteration to the gradient norm $\|\g_t\|$ and the magnitude of the smallest eigenvalue $|\lambda_{\min}(\B_t)|$ as well as $\Delta_t$. In the case of \textit{interior}  solutions ($\|\s_t\|_t<\Delta$), nothing changes compared to spherical Trust Region methods. When the computed step $s_t$ lies \textit{outside} the Trust Region, however, the guaranteed model decrease changes \textit{by a constant factor}, which accounts for the altered diameter of the trust region along that direction. Due to the uniform equivalence established in \ref{th:main} this factor is always strictly positive and finite.

More specifically, the first step of the proof relies on Assumption \ref{as:model_decrease} in order to relate the \textit{model decrease} at each iteration $t$ to three quantities of interest: i) the gradient norm $\|\g_t\|$, ii) the magnitude of the smallest eigenvalue $|\lambda_{\min}(\B_t)|$, and iii) the trust region radius $\Delta_t$. In the case of \textit{interior} solutions ($\|\s_t\|_t<\Delta$), the model decrease is shown as in the spherical Trust Region methods. When the computed step $s_t$ lies \textit{outside} the Trust Region, however, the guaranteed model decrease changes \textit{by a constant factor}, which accounts for the altered diameter of the trust region along that direction. Due to the uniform equivalence established in Lemma \ref{T:uniform_equivalence} this factor is always strictly positive and finite. 

From here on, the proof proceeds in a standard fashion (see e.g. \cite{yao2018inexact}). That is, a lower bound on $\Delta_t$ is established which (i) upper bounds the number of unsuccessful steps and (ii) lower bounds the guaranteed model decrease introduced above, which in turn allows to bound the overall number of successful steps as a fraction of the initial suboptimality in $\mathcal{L}$. Assumption \ref{a:sampling} together with the smoothness assumptions on $\mathcal{L}$ allow to finally relate the progress of each successful step to the actual \textit{function} decrease. Finally, since the function decreases and because it is lower bounded, there is a finite number of steps which we can upper bound.

\subsection{Proof}

In order to prove convergence results for ellipsoidal Trust Region methods one must ensure that the applied norms are coherent during the complete minimization process in the sense that the ellipsoids do not flatten out (or blow up) completely along any given direction. This intuition is formalized in Assumption \ref{a:uniform_equi} which we restate here for the sake of clarity.
\begin{definition}
[Definition \ref{a:uniform_equi} restated]
There exists a constant $\mu\geq 1$ such that
\begin{equation}
\frac{1}{\mu} \|\w\|_{\Am_t} \leq \|\w\|_2 \leq \mu\|\w\|_{\Am_t}, \qquad \forall t, \forall \w \in \mathbb{R}^d.
\end{equation}
\end{definition}

Towards this end, \cite{conn2000trust} identify the following sufficient condition on the basis of which we will prove that our proposed ellipsoid $\Am_{rms}$ is indeed uniformly equivalent under some mild assumptions.

\begin{lemma}[Theorem 6.7.1 in \cite{conn2000trust}]
Suppose that there exists a constant $\zeta\geq1$ such that
\begin{equation}
\frac{1}{\zeta} \leq \sigma_{\min}\left( \Am_t\right) \leq  \sigma_{\max}\left( \Am_t\right) \leq \zeta\qquad \forall t,
\end{equation}
then Definition \ref{a:uniform_equi} holds.
\end{lemma}

Having uniformly equivalent norms is sufficient to prove convergence of ellipsoidal TR methods (se AN.1 and Theorem 6.6.8 in \citep{conn2000trust}). However, it is so far unknown how the ellipsoidal constraints influence the convergence \textit{rate} itself. We here prove that the specific ellipsoidal TR method presented in Algorithm \ref{alg:src} preserves the rate of its spherically-constrained counterpart proposed in \cite{yao2018inexact} (see  Theorem \ref{th:rate} below). 

First, we show that the proposed $\Am_{rms,t}$ ellipsoid satisfies Definition 1.  

\begin{mdframed}
\begin{lemma}[Lemma \ref{T:uniform_equivalence} restated] 
Suppose $\|\g_t\|^2 \leq L_{H}^2$ for all $\w_t \in \mathbb{R}^d,$ $t=1,2,\ldots$ Then there always exists $\epsilon>0$ such that the proposed preconditioning matrices $\Am_{rms,t}$ (Eq.~\ref{eq:rms_ellips}) are uniformly equivalent, i.e. Def.~\ref{a:uniform_equi} holds. The same holds for the diagonal variant.
\end{lemma}
\end{mdframed}

\begin{proof}
The basic building block of our ellipsoid matrix consists of the current and past stochastic gradients $  \Gm_t:=[\g_1,\g_2,\ldots,\g_t].  $

We consider $\Am_{rms}$ which is built up as follows\footnote{This is a generalization of the diagonal variant proposed by \cite{tieleman2012lecture}, which preconditions the gradient step by an elementwise division with the square-root of the following estimate $g_t=(1-\beta)g_{t-1}+\beta \nabla \mathcal{L}(\w_t)^2$.}

\begin{equation}
    \Am_{rms,t}:=\left((1-\beta) \Gm \underbrace{\diag(\beta^t,\beta^{t-1},\ldots, \beta^0)}_{:=\Dm}\Gm^\intercal\right)+\epsilon\mathbf{I}.
\end{equation}

From the construction of $\Am_{rms,t}$ it directly follows that for any unit length vector $\u\in \mathbb{R}^d\setminus\{0\}, \|\u\|_2=1$ we have

\begin{equation}
\begin{aligned}\label{eq:lower_bound_spectrum}
&\u^\intercal\left( (1-\beta)\Gm\Dm\Gm^\intercal +  \epsilon \mathbf{I}\right) \u\\
=&(1-\beta)\u^\intercal\Gm\Dm^{1/2}(\Dm^{1/2})^\intercal\Gm^\intercal \u +\epsilon \|\u\|_2^2\\
=&(1-\beta)\left((\Dm^{1/2})^\intercal\Gm^\intercal\u\right)^\intercal\left((\Dm^{1/2})^\intercal\Gm^\intercal\u\right) +\epsilon \|\u\|_2^2\\
\geq &\epsilon>0,
\end{aligned}
\end{equation}

which proves the lower bound for $\zeta=1/\epsilon$. Now, let us consider the upper end of the spectrum of  $\Am_{rms,t}$. Towards this end, recall the geometric series expansion

\begin{equation}\label{eq:geometric_series}
    \sum_{i=0}^t \beta^{t-i} =  \sum_{i=0}^t \beta^i = \frac{1-\beta^{t+1}}{1-\beta}
\end{equation}

and the fact that $\Gm\Gm^\top$ is a sum of exponentially weighted rank-one positive semi-definite matrices of the form $\g_i\g_i^\intercal$. Thus $$\lambda_{max}(\g_i\g_i^\intercal)=\Tr(\g_i\g_i^\intercal)=\|\nabla \g_i\|^2\leq L_H^2,$$ where the latter inequality holds per assumption for any sample size $|S|$. Combining these facts we get that

\begin{equation}
\begin{aligned}
&\u^\intercal\left( (1-\beta)\Gm\Dm\Gm^\intercal +  \epsilon \mathbf{I}\right) \u\\
=&(1-\beta)\u^\intercal\Gm\Dm\Gm^\intercal\u +\epsilon\|\u\|_2^2\\
= & (1-\beta) \sum_{i=0}^t \beta^{t-i} \u^\intercal\g_i \g_i^\intercal \u+\epsilon\|\u\|_2^2\\
\leq&(1-\beta) \sum_{i=0}^t \beta^{t-i}L_{H}^2\|\u\|_2^2 +\epsilon\|\u\|_2^2\\
= & (1-\beta^{t+1})L_{H}^2+\epsilon.
\end{aligned}
\end{equation}

As a result we have that

\begin{equation}\label{eq:spectrum_bound}
    \epsilon\leq \lambda_{min}\left(\Am_{rms,t}\right)\leq \lambda_{max}\left(\Am_{rms,t}\right) \leq \left(1-\beta^{t+1}\right)L_{H}^2+\epsilon
\end{equation}
Finally, to achieve uniform equivalence we need the r.h.s. of (\ref{eq:spectrum_bound}) to be bounded by $1/\epsilon$. This gives rise to a quadratic equation in $\epsilon$, namely
\begin{equation}
    \epsilon^2+\left(1-\beta^{t+1}\right)L_{H}^2\epsilon-1\leq0
\end{equation}
which holds for any $t$ and any $\beta \in (0,1)$ as long as
\begin{equation}
0\leq\epsilon\leq \frac{1}{2}(\sqrt{L_{H}^4+4}-L_{H}^2).
\end{equation}
Such an $\epsilon$ always exists but one needs to choose smaller and smaller values as the upper bound on the gradient norm grows. For example, the usual value $\epsilon=10^{-8}$ is valid for all $L_{H}^2<9.9\cdot10^{7}$. All of the above arguments naturally extend to the diagonal preconditioner $\diag(\Am_{rms})$.

\end{proof}

Second, we note that it is no necessary to compute the update step by minimizing Eq.~(\ref{eq:tr_step}) to global optimality. Instead, it suffices to do better than the Cauchy- and Eigenpoint simultaneously \citep{conn2000trust,yao2018inexact}. We here adapt this assumption for the case of iteration dependent norms (compare \citep{conn2000trust} Chapter 6). restate this assumption here

\begin{assumption}[Approximate model minimization][A.\ref{as:model_decrease} restated]
Each update step $\s_t$ yields at least as much model decrease as the Cauchy- and Eigenpoint simultaneously, i.e. 

\begin{equation}
    m_t(\s_t)\leq m_t(\s_t^C) \quad \text{and} \quad  m_t(\s_t)\leq m_t(\s_t^E),
\end{equation}
where
\begin{align} \label{eq:cauchy_and_eigenpoint}
\s_t^C := \argmin_{0\le\alpha\leq\Delta_t} m_t(-\alpha \frac{\g_t}{\norm{\g_t}_t})
\quad \text{and} \quad
\s_t^E :=  \argmin_{|\alpha|\leq\Delta_t} m_t(\alpha \u_t),
\end{align}
where $\u_t$ is an approximation to the corresponding negative curvature direction, i.e., for some $ 0 < \nu < 1 $,
$\u_t^\intercal \H_t\u_t \le \nu \left(\frac{\|\u_t\|}{\|\u_t\|_t}\right)^2 \lambda_{min}(\B_t) \; \text{ and } \; \|\u_t\|_t = 1.
$\end{assumption}

In practice, improving upon the Cauchy point is easily satisfied by any Krylov subspace method such as Conjugate Gradients, which ensures convergence to first order critical points. However, while the Steihaug-Toint CG solver can exploit negative curvature, it does not explicitly search for the most curved eigendirection and hence fails to guarantee $m_t(\s_t)\leq m_t(\s_t^E)$. Thus more elaborate Krylov descent methods such as Lanczos method might have to be employed for second-order criticality (See also Appendix~\ref{sec:subproblem_solver} and \cite{conn2000trust} Chapter 7).

We now restate two results from \cite{conn2000trust} that precisely quantify the model decrease guaranteed by Assumption \ref{as:model_decrease}. 

\begin{lemma}[Model decrease: Cauchy Point (Theorem 6.3.1. in \cite{conn2000trust})]\label{lemma:cauchy}
Suppose that $\s_t^C$ is computed as in Eq.~(\ref{eq:cauchy_and_eigenpoint}). Then
\begin{equation}\label{eqn:cauchy_point_trust}
m_t(0)-m_t(\s_t^C) \geq \frac12 \|\g_t\|\min\{\frac{\|\g_t\|}{1+\|\Bm_t\|}, \Delta_t \frac{\|\g_t\|}{\|\g_t\|_t} \}.
\end{equation}
\end{lemma}

\begin{lemma}[Model decrease: Eigenpoint (Theorem 6.6.1 in \cite{conn2000trust})]\label{lemma:eig}
Suppose that $\lambda_{min}(\B_t)<0$ and $\s_t^E$ is computed as in Eq.~(\ref{eq:cauchy_and_eigenpoint}). Then
\begin{equation}\label{eq:ep_decrease}
    m_t(0)-m_t(\s_t^E)\geq -\frac12 \nu \lambda_{min}(\Bm_t) \left(\frac{\|\u_t\|}{\|\u_t\|_t}\right)^2  \Delta_t^2
\end{equation}
\end{lemma}

We are now ready to prove the final convergence results. Towards this end, we closely follow the line of arguments developed in \cite{yao2018inexact}. First, we restate the following lemma which holds independent of the trust region constraint choice.

\begin{lemma}[\cite{yao2018inexact}]\label{lemma:fun}
Assume that $\mathcal{L(\w)}$ is second-order smooth with Lipschitz constants $L_g$ and $L_H$. Furthermore, let Assumption \ref{a:sampling} hold. Then
\begin{equation}\label{eqn:estimation_m_F_trust}
F(\x_t+\s_t) - F(\x_t)-m_t(\x_t) \leq  \s_t^\intercal \left(\nabla F(\x_t) - \g_t\right) + \frac12 \delta_H\|\s_t\|^2 + \frac12 L_{H} \|\s_t\|^3.
\end{equation}
\end{lemma}

Second, we show that any iterate of Algorithm \ref{alg:src} is eventually successful as long as either the gradient norm or the smallest eigenvalue are above (below) the critical values $\epsilon_g$ and $\epsilon_H$ . 

\begin{lemma}[Eventually successful iteration - $\|\g_t\|\geq \epsilon_g$]\label{l:event_succ_g}
Assume that $\mathcal{L(\w)}$ is second-order smooth with Lipschitz constants $L_g$ and $L_H$. Furthermore, let Assumption \ref{a:sampling} and \ref{as:model_decrease} hold and suppose that $\|\g_t\|\geq \epsilon_g$ as well as
\begin{equation}\label{eq:123}
    \delta_g < \frac{1-\eta}{4\mu^2}\epsilon_g, ~\Delta_t \leq \min\left\{\frac{\mu \epsilon_g}{1 + L_g}, \sqrt{\frac{(1-\eta)\epsilon_g}{12L_H}}\frac1{\mu^4}, \frac{(1-\eta)\epsilon_g}{3\mu^4}\right\},
\end{equation}

\noindent then the step $\s_t$ is successful.
\end{lemma}
\begin{proof}
First, by Assumption \ref{as:model_decrease}, Lemma \ref{lemma:cauchy}, $\|\g_t\|\geq\epsilon_g$ and Lemma \ref{T:uniform_equivalence}, we have
\begin{equation}
\begin{aligned}
-m_t(\s_t) 
&\geq \frac12 \|\g_t\|\min\{\frac{\|\g_t\|}{1+\|\Bm_t\|}, \Delta_t \frac{\|\g_t\|}{\|\g_t\|_t} \} \\
& \geq \frac12 \|\g_t\|\min\{\frac{\epsilon_g}{1+\|\Bm_t\|}, \Delta_t \frac{\|\g_t\|}{\|\g_t\|_t}\}\\
& \geq \frac12 \|\g_t\|\min\{\frac{\epsilon_g}{1+\|\Bm_t\|}, \frac{\Delta_t}{\mu}\}\\
&=\frac12 \epsilon_g\frac{\Delta_t}{\mu},
\end{aligned}
\end{equation}

where the last equality uses the above assumed upper bound on $\Delta_t$ of Eq.~(\ref{eq:123}). Using this result together with Lemma \ref{lemma:fun} and the fact that $\|\s_t\|_2\leq \mu \|\s_t\|_t\leq \mu \Delta_t$ due to Lemma \ref{T:uniform_equivalence}, we find
\begin{align}\label{eq:l10}
1 - \rho_t 
&= \frac{\mathcal{L}(\w_t+\s_t)-\mathcal{L}(\w_t)-m_t(\s_t)}{-m_t(\s_t)} \\
&\leq \frac{\delta_g\Delta_t\mu + \frac12 \delta_H\Delta_t^2\mu^2 + \frac12 L_{H} \Delta_t^3\mu^3}{\frac12 \epsilon_g\frac{\Delta_t}{\mu}}\\
& = 2\frac{\delta_g}{\epsilon_g}\mu^2 + \frac{\delta_h}{\epsilon_g} \Delta_t\mu^3 + \frac{L_{H}}{\epsilon_g} \Delta_t^2\mu^4\\
& \leq \frac{1-\eta}2 + \left(\frac{\delta_H}{\epsilon_g} \Delta_t + \frac{L_{H}}{\epsilon_g} \Delta_t^2\right)\mu^4,
\end{align}
where the last inequality makes use of the upper bound assumed on $\delta_g$. Now, we re-use the result of Lemma 10 in \cite{yao2018inexact}, which states that $\left(\frac{\delta_H}{\epsilon_g} \Delta_t + \frac{L_{H}}{\epsilon_g} \Delta_t^2\right)\leq \frac{1-\eta}2 $ for $\Delta_t\leq \min\left\{\sqrt{\frac{(1-\eta)\epsilon_g}{12L_H}},\frac{(1-\eta)\epsilon_g}{3}\right\}$ to conclude that $\left(\frac{\delta_H}{\epsilon_g} \Delta_t + \frac{L_{H}}{\epsilon_g} \Delta_t^2\right)\mu^4\leq \frac{1-\eta}{2}$ for our assumed bound on $\Delta_t$ in Eq.~(\ref{eq:123}). As a result, Eq.~(\ref{eq:l10}) yields
$$1 - \rho_t \le 1 - \eta,$$
which implies that the iteration $t$ is successful.
\end{proof}

\begin{lemma}[Eventually successful iteration - $\lambda_{min}(\Bm_t)\leq -\epsilon_H$]\label{l:event_succ_H}
Assume that $\mathcal{L(\w)}$ is second-order smooth with Lipschitz constants $L_g$ and $L_H$. Furthermore, let Assumption \ref{a:sampling} and \ref{as:model_decrease} hold and suppose that $\|\g_t\|<\epsilon_g$ and $\lambda_{min}(\Bm_t) < -\epsilon_H$. If
\begin{equation}\label{eq:cond_ep}\delta_H < \dfrac{1-\eta}{2}\nu\epsilon_H, \Delta_t \leq \frac{(1-\eta)}{2\mu}\dfrac{\nu\epsilon_H}{L_{H}}\end{equation}

then iteration $t$ is successful.

\end{lemma}
\begin{proof} First, recall Eq.~(\ref{eqn:estimation_m_F_trust}) and note that, since both $\s_t$ and $-\s_t$ are viable search directions, we can assume $\s_t^\intercal\nabla F(\w_t)\leq 0$ w.l.o.g.. 
Then 
\begin{align*}
\mathcal{L}(\w_t+\s_t) - \mathcal{L}(\w_t)-m_t(\w_t) 
&\leq \frac12 \delta_H\|\s_t\|^2 + \frac12 L_{H} \|\s_t\|^3
\end{align*}
Therefore, recalling Eq.~(\ref{eq:ep_decrease}) as well as the fact that $\frac{\|\u_t\|_2}{\|\u_t\|_t}\leq \mu$ and $\|\s_t\|_2\leq \mu \|\s_t\|_t\leq \mu \Delta_t$ due to Lemma \ref{T:uniform_equivalence}
\begin{equation}
\begin{aligned}
1 - \rho_t 
&= \frac{\mathcal{L}(\w_t+\s_t)-\mathcal{L}(\w_t)-m_t(\s_t)}{-m_t(\s_t)} \\
&\leq \frac{\frac12 \delta_H\|\s_t\|^2 + \frac12 L_{H} \|\s_t\|^3}{\frac\nu2|\lambda_{min}(\Bm_t)|\Delta_t^2\mu^2}\\
&\leq  \frac{\frac12 \delta_H\|\s_t\|^2 + \frac12 L_{H} \|\s_t\|^3}{\frac\nu2\epsilon_H\Delta_t^2\mu^2}\\
& \leq  \frac{\frac12 \delta_H\Delta_t^2\mu^2 + \frac12 L_{H} \Delta_t^3\mu^3}{\frac\nu2\epsilon_H\Delta_t^2\mu^2}\\
& = \frac{\delta_H}{\nu\epsilon_H} + \frac{L_{H}\Delta_t\mu}{\nu\epsilon_H} \\
&< 1-\eta,
\end{aligned}
\end{equation}    

where the last second inequality is due to the conditions in Eq.~(\ref{eq:cond_ep}). Therefore, $\rho_t\geq \eta$ and the iteration is successful.
\end{proof}

Together, these two results allow us to establish a lower bound on the trust region radius $\Delta_t$.
\begin{lemma}\label{lemma:lowerbound_deltat}
Assume that $\mathcal{L(\w)}$ is second-order smooth with Lipschitz constants $L_g$ and $L_H$. Furthermore, let Assumption \ref{a:sampling} and \ref{as:model_decrease} hold. Suppose
\begin{equation*}
    \delta_g < \frac{1-\eta}{4}\epsilon_g, ~~~ \delta_H <\min\{ \frac{1-\eta}{2}\nu\epsilon_H,1\}.
\end{equation*}
then for Algorithm \ref{alg:src} we have
\begin{equation}\label{eq:lower_bound_delta_t}
  \Delta_t \geq \frac{1}{\gamma} \min\left\{\frac{\epsilon_g\mu}{1 + L_g}, \sqrt{\frac{(1-\eta)\epsilon_g}{12L_H\mu^8}}, \frac{(1-\eta)\epsilon_g}{3\mu^4},\frac{(1-\eta)}{2\mu} \frac{\nu\epsilon_H}{L_{H}}\right\},\quad \forall t=1,2,\ldots
\end{equation}
\end{lemma}
\begin{proof}
The proof follows directly from  $\Delta_{t}\geq \Delta_{t-1}/\gamma$ as well as the fact that any step is successful as soon as $\Delta_t$ falls below $\min\left\{\frac{\epsilon_g\mu}{1 + L_g}, \sqrt{\frac{(1-\eta)\epsilon_g}{12L_H\mu^8}}, \frac{(1-\eta)\epsilon_g}{3\mu^4},\frac{(1-\eta)}{2\mu} \frac{\nu\epsilon_H}{L_{H}}\right\}$ due to Lemma \ref{l:event_succ_g} and \ref{l:event_succ_H}.
\end{proof}

\begin{lemma}[Number of successful iterations] 
\label{lemma:upperbound_success_TR}	
Under the same setting as Lemma \ref{lemma:lowerbound_deltat}, the number of successful iterations taken by Algorithm \ref{alg:src} is upper bounded by
$$
|T_\text{succ}| \leq \frac{\mathcal{L}(\x_0) - \mathcal{L}(\x^*)}{C\epsilon_H\min\{\epsilon_g^2, \epsilon_H^2 \}},
$$
where  $C:=\eta\min\left\{C_1,C_2\right\}$, $C_1:=\frac12\min \left\{\frac{1}{1+L_g},C_g\right\}$, $C_2:=\frac{\nu\mu^2}{2} \min\left\{C_g^2,C_H^2\right\}$,$C_g:=\min\left\{\frac{\epsilon_g\mu}{1 + L_g}, \sqrt{\frac{(1-\eta)\epsilon_g}{12L_H\mu^8}}, \frac{(1-\eta)\epsilon_g}{3\mu^4}\right\}
$, $C_H:=\frac{(1-\eta)}{2\mu} \frac{\nu\epsilon_H}{L_{H}}$
\end{lemma}
\begin{proof}
Suppose Algorithm \ref{alg:src} does not terminate at iteration $t$. Then either $\|\g_t\|\geq\epsilon_g$ or $\lambda_{min}(\Bm)\leq -\epsilon_h$.
If $\|\g_t\|\geq\epsilon_g$, according to (\ref{eqn:cauchy_point_trust}) and Lemma \ref{T:uniform_equivalence}, we have
\begin{align*}
-m_t(\s_t) 
&\geq \frac12 \|\g_t\|\min\{\frac{\|\g_t\|}{1+\|\H_t\|}, \Delta_t\frac{1}{\mu}\}\\
&\geq \frac12 \epsilon_g\min\{\frac{\epsilon_g}{1+L_g}, C_g\epsilon_g, C_H\epsilon_H\}\\
&\geq  C_1 \epsilon_g\min\{\epsilon_g, \epsilon_H\}.
\end{align*}

Similarly, in the second case $\lambda_{min}(\Bm_t)\leq -\epsilon_h$, from Lemma \ref{T:uniform_equivalence} and \ref{lemma:eig} we have
$$
-m_t(\s_t) \geq \frac12\nu|\lambda_{min}(\Bm_t)|\Delta_t^2\mu^2\geq C_2\epsilon_H  \min\{\epsilon_g^2, \epsilon_H^2\}.
$$
Let $T_\text{succ}$ denote the number of successful iterations. Since $\mathcal{L}(\w)$ is monotonically decreasing, we have
\begin{align*}
\mathcal{L}(\w_0) - \mathcal{L}(\w^*) &\geq \sum_{t=0} \mathcal{L}(\w_t) - \mathcal{L}(\w_{t+1})\\
&\geq \sum_{t\in T_\text{succ}}\mathcal{L}(\w_t) - \mathcal{L}(\w_{t+1})\\
&\geq \sum_{t\in T_\text{succ}}- m_t(\s_t)\eta\\
& \geq  \sum_{t\in T_\text{succ}} C\epsilon_H \min\{\epsilon_g^2, \epsilon_H^2\}\\
& \geq |T_\text{succ}| C\epsilon_H \min\{\epsilon_g^2, \epsilon_H^2\},
\end{align*}
which proves the assertion.
\end{proof}

We are now ready to prove the final result. Particularly, given the lower bound on $\Delta_t$ established in Lemma \ref{lemma:lowerbound_deltat} we find an upper bound on the number of un-successful iterations, which combined with the result of Lemma \ref{lemma:upperbound_success_TR} on the number of successful iterations yields the total iteration complexity of Algorithm \ref{alg:src}.

\begin{framed}
\begin{theorem}[Theorem \ref{th:rate} restated]
Assume that $\mathcal{L(\w)}$ is second-order smooth with Lipschitz constants $L_g$ and $L_H$. Furthermore, let Assumption \ref{a:sampling} and \ref{as:model_decrease} hold. Then Algorithm 1 finds an $\mathcal{O}(\epsilon_g,\epsilon_H)$ first- and second-order stationary point in at most $\mathcal{O}\left(\max\left\{\epsilon_g^{-2}\epsilon_H^{-1},\epsilon_H^{-3}\right\}\right)$ iterations.
\end{theorem}
\end{framed}
\begin{proof}
The result follows by combining the lemmas \ref{lemma:lowerbound_deltat} and \ref{lemma:upperbound_success_TR} as in Theorem 1 of \cite{xu2017second}. Specifically, suppose that Algorithm \ref{alg:src} terminates at iteration T. Then the total number of iterations $T=T_\text{succ}+T_\text{unsucc}$ and $\Delta_T=\Delta_0\cdot \gamma^{T_\text{succ}-T_\text{unsucc}}$. From Lemma \ref{lemma:lowerbound_deltat} we have $\Delta_T\geq \frac{1}{\gamma} \min\left\{\frac{\epsilon_g\mu}{1 + L_g}, \sqrt{\frac{(1-\eta)\epsilon_g}{12L_H\mu^8}}, \frac{(1-\eta)\epsilon_g}{3\mu^4},\frac{(1-\eta)}{2\mu} \frac{\nu\epsilon_H}{L_{H}}\right\} :=\Delta_{inf}$. Hence, $(T_\text{succ}-T_\text{unsucc}) \log(\gamma) \geq \log(\Delta_{inf}/\Delta_0)$, which implies

\begin{equation}\label{eq:unsucc_steps}
T_\text{unsucc} \leq \frac{log(\Delta_0/\Delta_{inf})}{log(\gamma)}+T_{succ}.    
\end{equation}
Finally, combining Eq.~\ref{eq:unsucc_steps} with the upper bound on successful steps from Lemma \ref{lemma:upperbound_success_TR} yields

\begin{equation*}
    T\leq \frac{log(\Delta_0/\Delta_{inf})}{log(\gamma)}+2\frac{\mathcal{L}(\x_0) - \mathcal{L}(\x^*)}{C\epsilon_H\min\{\epsilon_g^2, \epsilon_H^2 \}} \in \mathcal{O}\left(\max\left\{\epsilon_g^{-2}\epsilon_H^{-1},\epsilon_H^{-3}\right\}\right)
\end{equation*}

\end{proof}

\section{Diagonal Dominance in Neural Networks}\label{sec:diag_appendix}
In the following, we make statements about the diagonal share of random matrices. As $\E[\frac{1}{x}]$ might not exist for a random variable $x$, we cannot compute the expectation of the diagonal share but rather of for computing the diagonal share of the expectation of the random matrix in absolute terms. Note that this notion is still meaningful, as the average of many non-diagonally dominated matrices with positive entries cannot become diagonally dominated.  
\subsection{Proof of Proposition \ref{prop:wigner}}
\begin{proposition}[Proposition \ref{prop:wigner} restated]
For random Gaussian Wigner matrix $\Wm$ formed as
\begin{equation} \label{eq:wiggly_wigner}
    \Wm_{i,j}=\Wm_{j,i}:= \begin{cases}
\sim\mathcal{N}(0,\sigma_1^2),\; i<j\\
\sim\mathcal{N}(0,\sigma_2^2),\; i=j,
\end{cases}
\end{equation}
where $\sim$ stands for i.i.d. draws~\citep{wigner1993characteristic}, the diagonal mass of the expected absolute matrix amounts to
\begin{equation}
    \delta_{\E\left[|\Wm|\right]}=\frac{1}{1+(d-1)\frac{\sigma_2}{\sigma_1}}.
\end{equation}
\end{proposition}
\begin{proof}
\begin{equation}
\begin{aligned}\label{eq:diag_mass_wigner}
\delta_{\E,\Wm}&=\frac{\sum_{k=1}^d \E\left[|\Wm_{k,k}|\right]}{\sum_{k=1}^d\sum_{l=1}^d \E\left[|\Wm_{k,l}|\right]}=\frac{d\E\left[|\Wm_{1,1}|\right]}{d\E\left[|\Wm_{1,1}|\right]+d(d-1)\E\left[|\Wm_{1,2}|\right]}\\&
=\frac{d\sigma_1 \sqrt{2/\pi}}{d\sigma_1 \sqrt{2/\pi}+d(d-1)\sigma_2 \sqrt{2/\pi}}=\frac{1}{1+\frac{d(d-1)\sigma_2 \sqrt{2/\pi}}{d\sigma_1 \sqrt{2/\pi}}}\\
&=\frac{1}{1+(d-1)\frac{\sigma_2}{\sigma_1}}
\end{aligned}
\end{equation}
which simplifies to $\frac{1}{d}$ if the diagonal and off-diagonal elements come from the same Gaussian distribution ($\sigma_1=\sigma_2$).
\end{proof}

For the sake of simplicity we only consider Gaussian Wigner matrices but the above argument naturally extends to any distribution with positive expected absolute values, i.e. we only exclude the Dirac delta function as probability density.

\begin{figure}[H]
\centering
\begin{tabular}{cc}
    CONV & MLP \\
    \includegraphics[width=0.4\linewidth,valign=c]{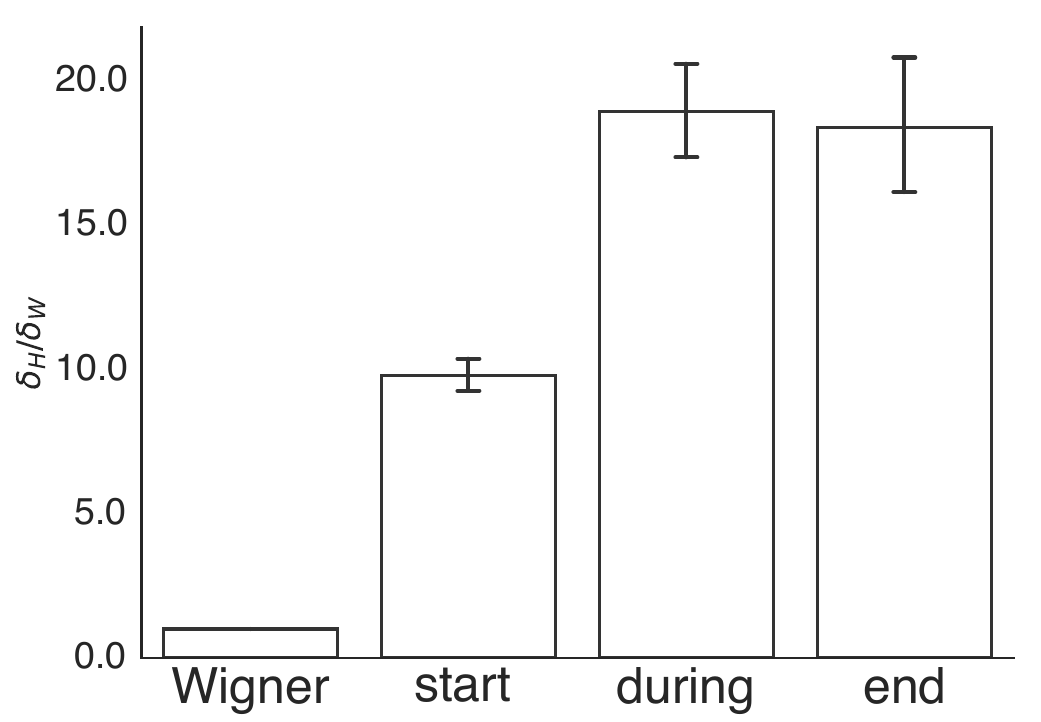}&
    \includegraphics[width=0.4\linewidth,valign=c]{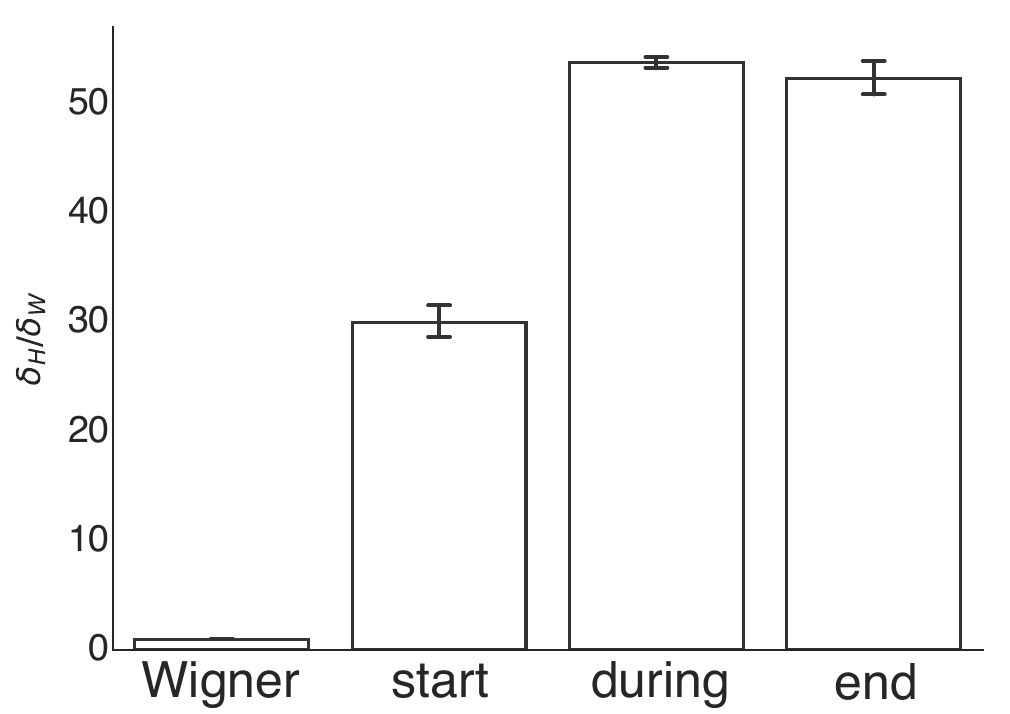}\\
\end{tabular}
\caption{Share of diagonal mass of the Hessian $\delta_\Hm$ relative to $\delta_\Wm$ of the corresponding Wigner matrix at random initialization, after 50\% iterations and at the end of training with RMSprop on MNIST. Average and 95\% confidence interval over 10 runs. See Figure \ref{fig:diag_dominance} for CIFAR-10 results.}\label{table:diag_dominance_mnist}
\end{figure}

\subsection{OLS Baseline}
\begin{figure}[H]
\centering
\begin{tabular}{cc}
     \includegraphics[width=0.475\linewidth,valign=c]{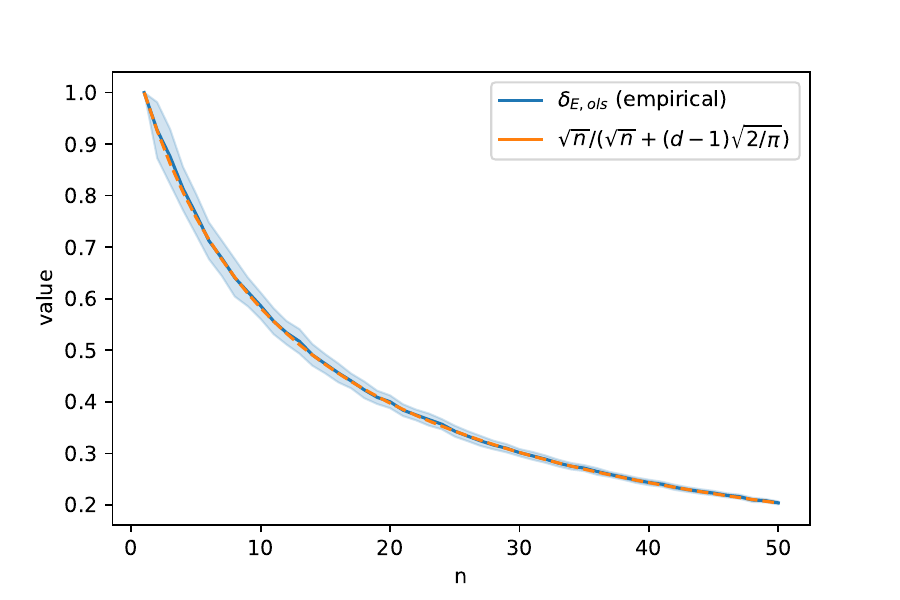}
\end{tabular}\label{fig:OLS}
\caption{Validity of Proposition \ref{prop:ols} in the small $n$ regime for Gaussian data (mean and 94\% confidence interval of 50 independent samples for each value of $n$.}\end{figure}

When considering regression tasks, a direct competitor to neural network models is the classical Ordinary Least Squares (OLS) regression, which minimizes a quadratic loss over a \textit{linear} model. In this case the Hessian simply amounts to the input-covariance matrix $\Hm_{\text{ols}}:=\Xm^\intercal \Xm$, where $\Xm\in \mathbb{R}^{d\times n}$. We here show that the diagonal share of the expected matrix itself also decays in $d$, when $n$ grows to infinity. However, empirical simulations suggest the validity of this result even for much smaller values of $n$ (see Figure \ref{fig:OLS}) and it is likely that finite $n$ results can be derived when adding assumptions such as Gaussian data.

\begin{proposition}[Proposition \ref{prop:ols} restated]
Let $\Xm\in \mathbb{R}^{n\times d}$ and assume each $\x_{i,k}$ is generated i.i.d. with zero-mean finite second moment $\sigma^2>0$. Then the share of diagonal mass of the expected matrix  $\E\left[|\Hm_{\text{ols}}|\right]$ amounts to
\begin{equation}
   \delta_{\E\left[|\Hm_{\text{ols}}|\right]} \overset{n \rightarrow \infty}{\rightarrow} \frac{\sqrt{n}}{\sqrt{n}+(d-1)\sqrt{\frac{2}{\pi}}}
\end{equation}
\end{proposition}
\begin{proof}
\begin{equation}
\begin{aligned}
\delta_{\E\left[|\Hm_{\text{ols}}|\right]} &=\frac{\sum_{k=1}^d \E\left[|(\Hm_{\text{ols}})_{k,k}|\right]}{\sum_{k=1}^d\sum_{l=1}^d \E\left[|(\Hm_{\text{ols}})_{k,l}|\right]}=\frac{\sum_{k=1}^d \E\left[|\sum_{i=1}^n \x_{i,k}^2|\right]}{\sum_{k=1}^d\sum_{l=1}^d \E\left[|  \sum_{i=1}^n \x_{i,k}\x_{i,l} |\right]}\\&=\frac{d \sum_{i=1}^n \E\left[\x_{i,1}^2\right]}{d\sum_{i=1}^n \E\left[\x_{i,1}^2\right]+d(d-1)\E\left[ \left|  \sum_{i=1}^n \x_{i,1}\x_{i,2} \right|\right]}
\end{aligned}
\end{equation}

Where we used the fact that all $\x_{i,k}$ are i.i.d. variables. Per assumption, we have $\E\left[\x_{i,1}^2\right]=\sigma^2, \forall i$. Furthermore, the products $\x_{i,1}\x_{i,2}$ are i.i.d with expectation 0 and variance $\sigma^4$. By the central limit theorem 
$$ Z_N=\frac{1}{\sqrt{n}} \sum_{i=1}^n \x_{i,1}\x_{i,2} \rightarrow Z$$
in law, with $Z \sim {\cal N}(0,\sigma^4)$, since $\E\left[Z_n^2\right]=\frac{1}{n}\E\left[\left(\sum_{i=1}^n \x_{i,1}\x_{i,2}\right)^2\right]=\frac{1}{n}\sum_{i=1}^n \E[\x_{i,1}^2]\E[\x_{i,2}^2]=\sigma^4$ due to the independence assumption.
Then $E\left(\;|Z_N|\; 1_{|Z_N|\geq R} \right)\ \leq \ E(|Z_N|^2/R)\leq \sigma^4/R$. This implies that
$$  E\left( \left|Z_N\right| \right) \rightarrow E(|Z|)=\sqrt{\frac{2}{\pi}} \sigma^2$$

As a result, we have that in the limit of large $n$

\begin{equation}
 \delta_{\E\left[|\Hm_{\text{ols}}|\right]}  \overset{n \rightarrow \infty}{\rightarrow} \frac{dn\sigma^2}{dn\sigma^2+d(d-1)\sqrt{n}\sqrt{\frac{2}{\pi}}\sigma^2 }=\frac{1}{1+\frac{(d-1)\sqrt{\frac{2}{\pi}}}{\sqrt{n}}}=\frac{\sqrt{n}}{\sqrt{n}+(d-1)\sqrt{\frac{2}{\pi}}}
\end{equation}

\end{proof}
\appendix

\part*{Appendix B: Background on second-order optimization}
\section{Newton's Method}
The canonical second-order method is Newton's methods. This algorithm uses the inverse Hessian as a scaling matrix and thus has updates of the form
\begin{equation}
 \w_{t+1} = \w_t- \nabla^2  \mathcal{L}(\w_t)^{-1}\nabla  \mathcal{L}(\w_t),
\end{equation}

which is equivalent to optimizing the local quadratic model 

\begin{equation}\label{eq:newton_model}
m_N (\w_t):= \mathcal{L}(\w_t) + \nabla \mathcal{L}(\w_t)^\intercal \s + \frac{1}{2} \s^\intercal \nabla^2  \mathcal{L}(\w_t) \s
\end{equation}
to \textit{first-order stationarity}. Using curvature information to rescale the steepest descent direction gives Newton's method the useful property of being linearly scale invariant
. This gives rise to a \textit{problem independent} local convergence rate that is super-linear and even quadratic in the case of Lipschitz continuous Hessians (see \cite{nocedal2006nonlinear} Theorem 3.5), whereas gradient descent at best achieves linear local convergence \citep{nesterov2013introductory}.

However, there are certain drawbacks associated with applying classical Newton's method. First of all, the Hessian matrix may be singular and thus not invertible. Secondly, even if it is invertible the local quadratic model (Eq.~\ref{eq:newton_model})
that is minimized in each NM iteration may simply be an inadequate approximation of the true objective. As a result, the Newton step is not necessarily a descent step. It may hence approximate arbitrary critical points (including local maxima) or even diverge. Finally, the cost of forming and inverting the Hessian sum up to $O(nd^2 + d^3)$ and are thus prohibitively high for applications in large dimensional problems. 

\section{Trust Region Methods} 
\subsection{Outer iterations}
Trust region methods are among the most principled approaches to overcome the above mentioned issues. These methods also construct a quadratic model $m_t$ but constrain the subproblem in such a way that the stepsize is restricted to stay within a certain radius $\Delta_t$ within which the model is trusted to be sufficiently adequate
\begin{equation}\label{eq:quadratic_model_app}
\min_{s\in \mathbb{R}^d}\: m_t(\s) = \mathcal{L}(\w_t)+\nabla \mathcal{L}(\w_t)^\intercal \s + \frac{1}{2} \s^\intercal \nabla^2\mathcal{L}(\w_t) \s, \:\:\: s.t.\: \|\s\|\leq \Delta_t.
\end{equation}

Hence, contrary to line-search methods this approach finds the step $\s_t$ and its length $\norm{\s_t}$ \textit{simultaneously} by optimizing (\ref{eq:quadratic_model_app}). Subsequently the actual decrease $\mathcal{L}(\w_t)-\mathcal{L}(\w_t+\s_t)$ is compared to the predicted  decrease $m_t(0)-m_t(\s_t)$ and the step is only accepted if the ratio $\rho:=\mathcal{L}(\w_t)-\mathcal{L}(\w_t+\s_t)/(m_t(0)-m_t(\s_t))$ exceeds some predefined success threshold $\eta_1>0$. Furthermore, the trust region radius is decreased whenever $\rho$ falls below $\eta_1$ and it is increased whenever $\rho$ exceeds the "very successful" threshold $\eta_20$. Thereby, the algorithm adaptively measures the accuracy of the second-order Taylor model -- which may change drastically over the parameter space depending on the behaviour of the higher-order derivatives\footnote{Note that the second-order Taylor models assume constant curvature.} -- and adapts the effective length along which the model is trusted accordingly. See \cite{conn2000trust} for more details.

As a consequence, the plain Newton step $\s_{N,t}=-\left(\nabla^2 \mathcal{L}_t\right)^{-1}\nabla \mathcal{L}_t$ is only taken if it lies within the trust region radius and yields a certain amount of decrease in the objective value. Since many functions look somehow quadratic close to a minimizer the radius can be shown to grow asymptotically under mild assumptions such that eventually full Newton steps are taken in every iteration which retains the local quadratic convergence rate \citep{conn2000trust}.

\begin{figure}[h!]
\centering          
          \begin{tabular}{c@{}c@{}}
            \adjincludegraphics[width=0.4\linewidth, trim={22pt 22pt 30pt 30pt},clip]{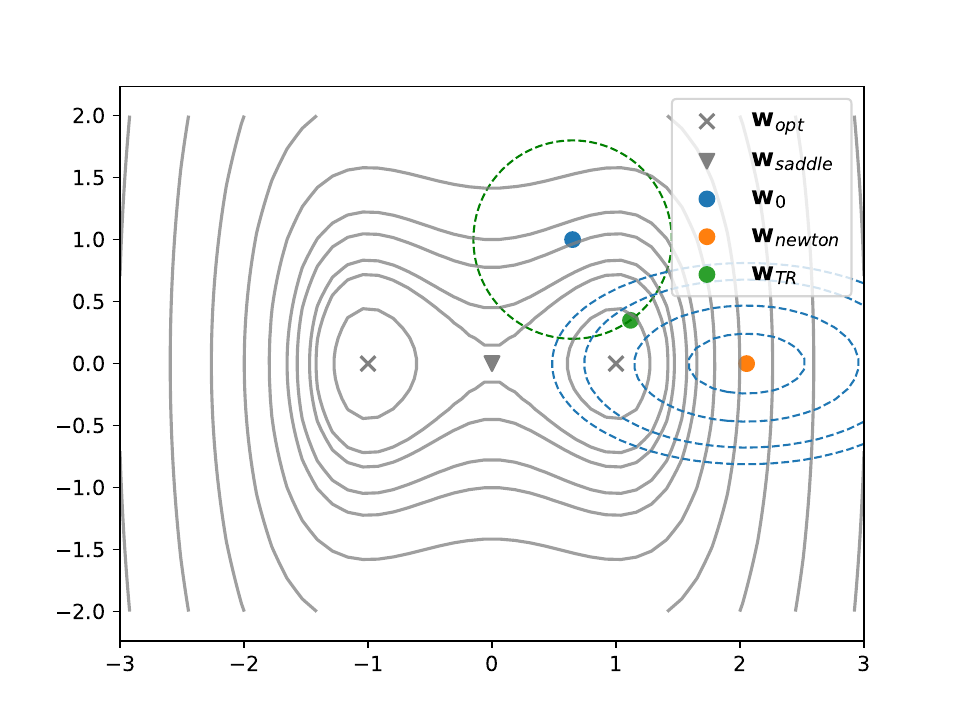} &
             \adjincludegraphics[width=0.4\linewidth, trim={22pt 22pt 30pt 30pt},clip]{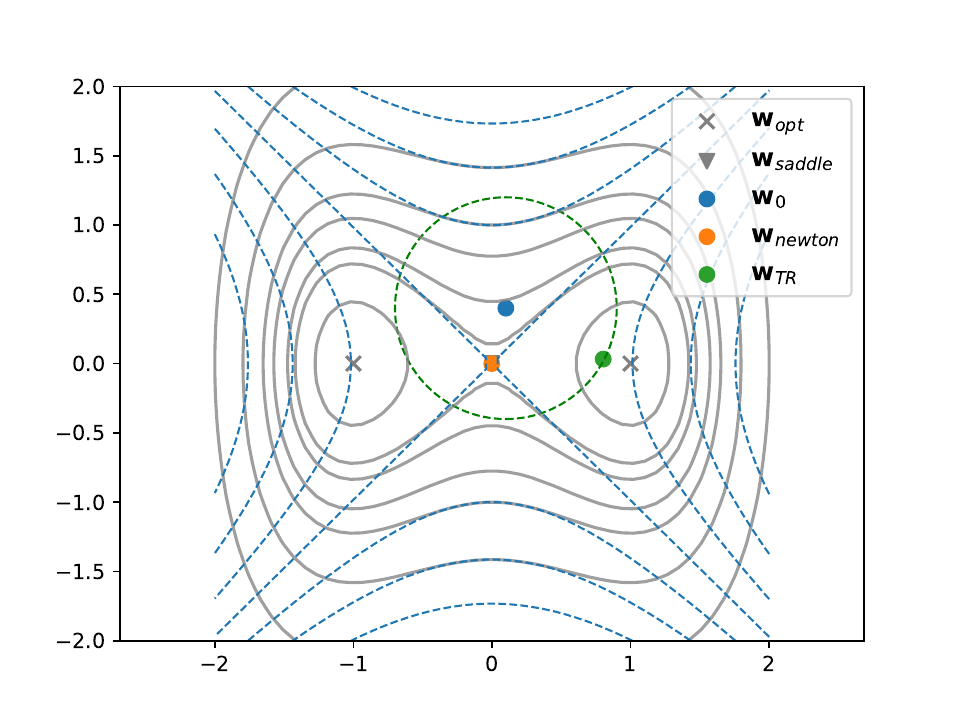}
	  \end{tabular}
          \caption{ \footnotesize{Level sets of the non-convex, coercive objective function $f(\w)=0.5\w_0^2 +0.25 \w_1^4 - 0.5\w_1^2$. Newton's Method makes a local quadratic model (blue dashed lines) and steps to its \textit{critical point}. It may be thus be ascending (left) or attracted by a saddle point (right). TR methods relieve this issue by stepping to the \textit{minimizer} of that model \textit{within} a certain region (green dashed line).}}
          \label{fig:tr_intuition}
\end{figure}
\subsection{Subproblem solver}\label{sec:subproblem_solver}
Interestingly, there is no need to optimize Eq.~(\ref{eq:quadratic_model_app}) to global optimality to retain the remarkable global convergence properties of TR algorithms. Instead, it suffices to do better than the Cauchy- and Eigenpoint\footnote{which are the model minimizers along the gradient and the eigendirection associated with its smallest eigenvalue, respectively.} simultaneously. One popular approach is to minimize $m_t(\s)$ in nested Krylov subspaces. These subspaces naturally include the gradient direction as well as increasingly accurate estimates of the leading eigendirection \begin{equation}\label{eq:krylov}
 \text{span}\{\g_t,\Bm_t\g_t,\Bm_t^2\g_t,\ldots,\Bm_t^j\g_t\}
\end{equation}
until (for example) the stopping criterion
\begin{equation}\label{eq:stopping_criterion}
    \|\nabla m_t(\s_j)\| \leq \|\nabla \mathcal{L}(\w_t)\|\min \{\kappa_K,\|\nabla \mathcal{L}(\w_t)\|^{\theta}\} ,\quad \kappa_K<1,\theta\geq 0
\end{equation}
 
is met, which requires increased accuracy as the underlying trust region algorithm approaches criticality. Conjugate gradients and Lanczos method are two iterative routines that implicitly build up a conjugate and orthogonal basis for such a Krylov space respectively and they converge \textit{linearly} on quadratic objectives with a square-root dependency on the condition number of the Hessian~\citep{conn2000trust}. We here employ the preconditionied Steihaug-Toint CG method \citep{steihaug1983conjugate} in order to cope with possible boundary solutions of (\ref{eq:quadratic_model_app}) but similar techniques exist for the Lanczos solver as well for which we also provide code. As preconditioning matrix for CG we use the same matrix as for the ellipsoidal constraint.

\section{Damped (Gauss-)Newton methods}\label{sec:damped_newton}
An alternative approach to actively constraining the region within which the model is trusted is to instead penalize the step norm in each iteration in a Lagrangian manner. This is done by so-called damped Newton methods that add a $\lambda>0$ multiple of the identity matrix to the second-order term in the model, which leads to the update step

\begin{equation}\label{eq:damped_newton}
\begin{aligned}
\min_{s\in \mathbb{R}^d}\: m_t(\s) &= \mathcal{L}(\w_t)+\nabla \mathcal{L}(\w_t)^\intercal \s + \frac{1}{2} \s^\intercal (\nabla^2\mathcal{L}(\w_t)+\lambda \mathbf{I}) \s \\&= \mathcal{L}(\w_t)+\nabla \mathcal{L}(\w_t)^\intercal \s + \frac{1}{2} \s^\intercal \nabla^2 \mathcal{L}(\w_t)\s + \lambda \|\s\|^2.
\end{aligned}
\end{equation}
This can also be solved hessian-free by conjugate gradients (or other Krylov subspace methods). The penalty parameter $\lambda$ is acting inversely to the trust region radius $\Delta$ and it is often updated accordingly. Such algorithms are commonly known as Levenberg-Marquardt algorithms and they were originally tailored towards solving non-linear least squares problems 
\citep{nocedal2006nonlinear} but they have been proposed for neural network training already early on \citep{hagan1994training}.

Many algorithms in the existing literature replace the use of $\nabla^2\mathcal{L}(\w_t)$ in (\ref{eq:damped_newton}) with the Generalized Gauss Newton matrix~\citep{martens2010deep,chapelle2011improved} or an approximation of the latter \citep{martens2015optimizing}. This matrix constitutes the first part of the well-known Gauss-Newton decomposition 
\begin{equation}\label{eq:GNDECOMP}
\nabla^2 \mathcal{L}(\cdot)= \underbrace{\frac{1}{n} \sum_{i=1}^n \ell''(f_i(\cdot)) \nabla f_i(\cdot)\nabla f_i(\cdot)^\intercal}_{:=\Am_{GGN}} + \frac{1}{n}\sum_{i=1}^n \ell' (f_i(\cdot)) \nabla^2f_i(\cdot),
\end{equation}
where $l'$ and $l''$ are the first and second derivative of $l:\mathbb{R}^{out}\rightarrow\mathbb{R}^+$ assuming that $out=1$ (binary classification and regression task) for simplicity here.

It is interesting to note that the GGN matrix $\Am_{GGN}$ of neural networks is equivalent to the Fisher matrix used in natural gradient descent~\citep{amari1998natural} in many cases like linear activation function and squared error as well as sigmoid and cross-entropy or softmax and negative log-likelihood for which the extended Gauss-Newton is defined \citep{pascanu2013revisiting}. As can be seen in (\ref{eq:GNDECOMP}) the matrix $\Am_{GGN}$ is positive semidefinite (and low rank if $n<d$). As a result, there exist no second-order convergence guarantees for such methods on general non-convex problems. On the other end of the spectrum, the GGN also drops possibly positive terms from the Hessian (see \ref{eq:GNDECOMP}). Hence it is not guaranteed to be an upper bound on the latter in the PSD sense. Essentially, GGN approximations assume that the network is piece-wise linear and thus the GGN and Hessian matrices only coincide in the case of linear and ReLU activations or non-curved loss functions. For any other activation the GGN matrix may approximate the Hessian only asymptotically and if the $\ell'(f_i(\cdot))$ terms in \ref{eq:GNDECOMP} go to zero for all $i\in\{1,\ldots,n\}$. In non-linear least squares such problems are called zero-residual problems and GN methods can be shown to have quadratic local convergence there. In any other case the convergence rate does not exceed the linear local convergence bound of gradient descent. In practice however there are cases where deep neural nets do show negative curvature in the neighborhood of a minimizer~\citep{bottou2018optimization}.
Finally, \cite{dauphin2014identifying} propose the use of the absolute Hessian instead of the GGN matrix in a framework similar to \ref{eq:damped_newton}. This method has been termed \textit{saddle-free Newton} even though its manifold of attraction to a given saddle is non-empty\footnote{It is the same as that for GD, which renders the method unable to escape e.g. when initialized right on a saddle point. To be fair, the manifold of attraction for GD constitutes a measure zero set~\citep{lee2016gradient}.}. 



\begin{figure}[h!]
\centering          
          \begin{tabular}{c@{}c@{}}
            \adjincludegraphics[width=0.4\linewidth, trim={22pt 22pt 30pt 30pt},clip]{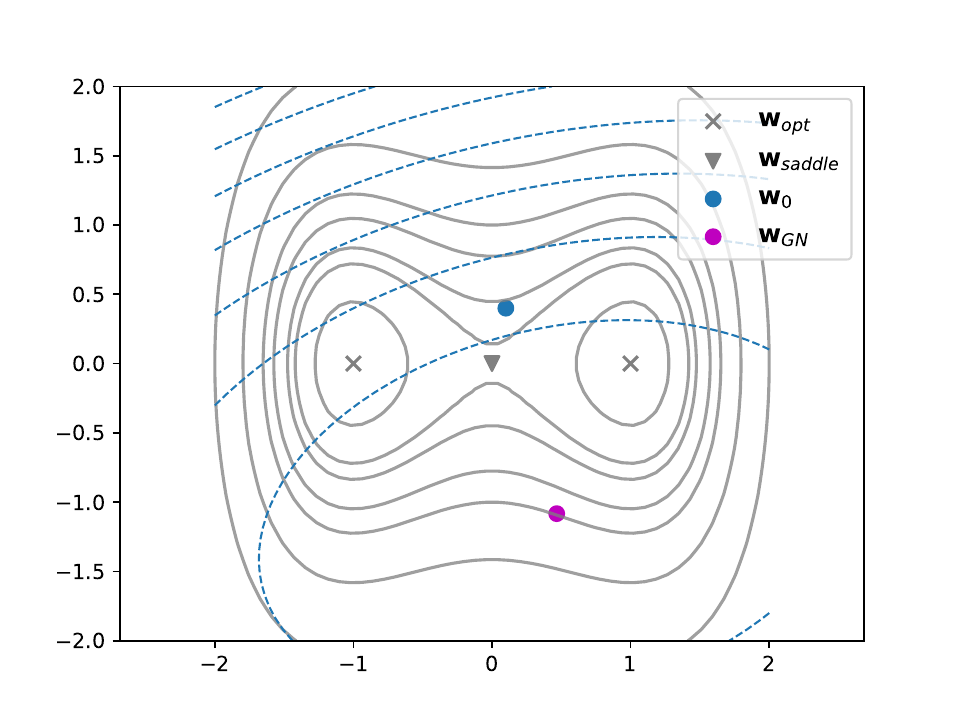} &
             \adjincludegraphics[width=0.4\linewidth, trim={22pt 22pt 30pt 30pt},clip]{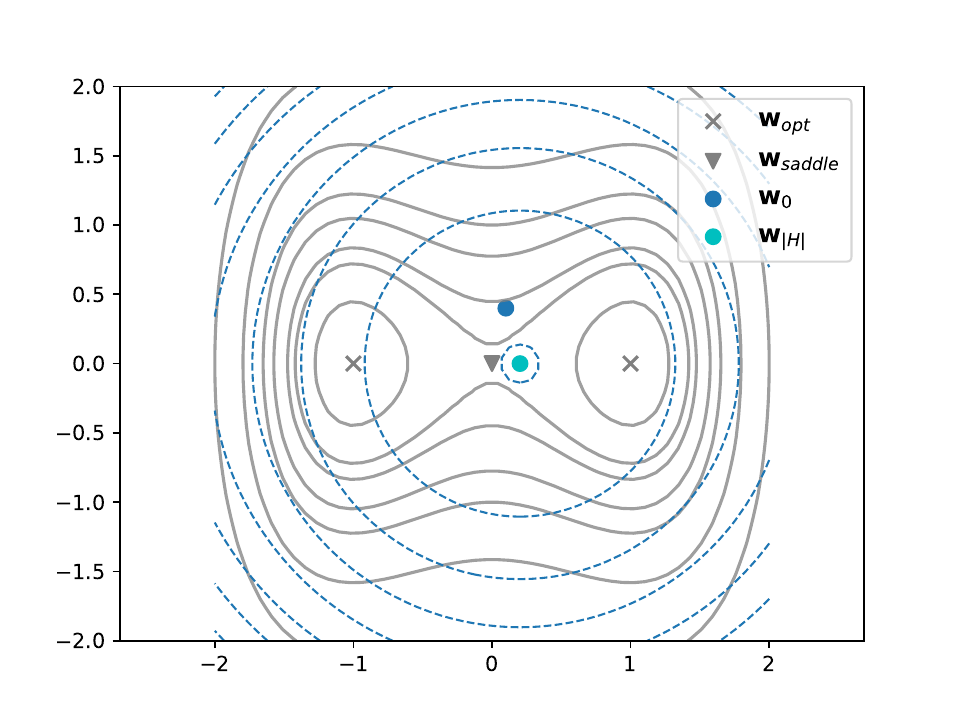}
	  \end{tabular}
          \caption{ \footnotesize{Both, the GGN method and saddle-free Newton method make a positive definite quadratic model around the current iterate and thereby overcome the abstractedness of pure Newton towards the saddle (compare Figure \ref{fig:tr_intuition}). However, (i) none of these methods can escape the saddle once they are in the gradient manifold of attraction and (ii) as reported in \cite{mizutani2008second} the GN matrix can be significantly less well conditioned than the absolute Hessian (here $\kappa_{GN}= 49'487'554$ and $\kappa_{|H|}=1.03$ so we had to add a damping factor of $\lambda=0.1$ to make the GN step fit the plot. }}
          \label{fig:gn_intuition}
\end{figure}

\subsection{Comparison to trust region} \label{sec:TRvsGN}
\textit{Contrary} to TR methods, the Levenberg-Marquardt methods never take plain Newton steps since the regularization is always on ($\lambda>0$). Furthermore, if a positive-definite Hessian approximation like the Generalized Gauss Newton matrix is used, this algorithm is not capable of exploiting negative curvature and there are cases in neural network training where the Hessian is much better conditioned than the Gauss-Newton matrix \citep{mizutani2008second} (also see Figure \ref{fig:gn_intuition}). While some scholars believe that positive-definiteness is a desirable feature \citep{martens2010deep,chapelle2011improved}, we want to point out that following negative curvature directions is necessarily needed to escape saddle points and it can also be meaningful to follow directions of negative eigenvalue $\lambda$ outside a saddle since they guarantee $\mathcal{O}(|\lambda|^3)$ progress, whereas a gradient descent step yields at least $\|\nabla f(\w)\|^2$ progress (both under certain stepsize conditions) and one cannot conclude a-priori which one is better in general \citep{curtis2017exploiting,alain2018negative}. 
Despite these theoretical considerations, many methods based on GGN matrices have been applied to neural network training (see \cite{martens2014new} and references therein) and particularly the hessian-free implementations of \citep{martens2010deep,chapelle2011improved} can be implemented very cheaply \citep{schraudolph2002fast}. 
\vspace{-0.35cm}
\section{Using Hessian information in Neural Networks}\label{sec:2ndOrderDiscussion}
While many theoretical arguments suggest the superiority of regularized Newton methods over gradient based algorithms, several practical considerations cast doubt on this theoretical superiority when it comes to neural network training. Answers to the following questions are particularly unclear: Are saddles even an issue in deep learning? Is superlinear local convergence a desirable feature in machine learning applications (test error)? Are second-order methods more "vulnerable" to sub-sampling noise? Do worst-case iteration complexities even matter in real-world settings? As a result, the value of Hessian information in neural network training is somewhat unclear a-priori and so far a conclusive empirical study is still missing. 

Our empirical findings indicate that the net value of Hessian information for neural network training is indeed somewhat limited for mainly three reasons: 1) second-order methods rarely yield better limit points, which suggests that saddles and spurious local minima are not a major obstacle; 2) gradient methods can indeed run on smaller batch sizes which is beneficial in terms of epoch and when memory is limited; 3) The per-iteration time complexity is noticeably lower for first-order methods. In summary, these observations suggest that advances in hardware and distributed second-order algorithms (e.g., ~\cite{osawa2018second,dunner2018distributed}) will be needed before Newton-type methods can replace (stochastic) gradient methods in deep learning.

\appendix
\part*{Appendix C: Experiment details}
\label{sec:appendix_experiment_details}

\section{Ellipsoidal Trust Region vs. First-order Optimizers}\label{sec:exp_overview}\hypertarget{app_sec:exp_results}{ }

\begin{figure*}[h!]
 \centering 
          \begin{tabular}{c@{}c@{}c@{}c@{}}
        &   \footnotesize{MLP} & \footnotesize{ResNet} & \footnotesize{Autoencoder} \\
        \rotatebox[origin=c]{90}{\footnotesize{Fashion-MNIST}} &     \includegraphics[width=0.29\linewidth,valign=c]{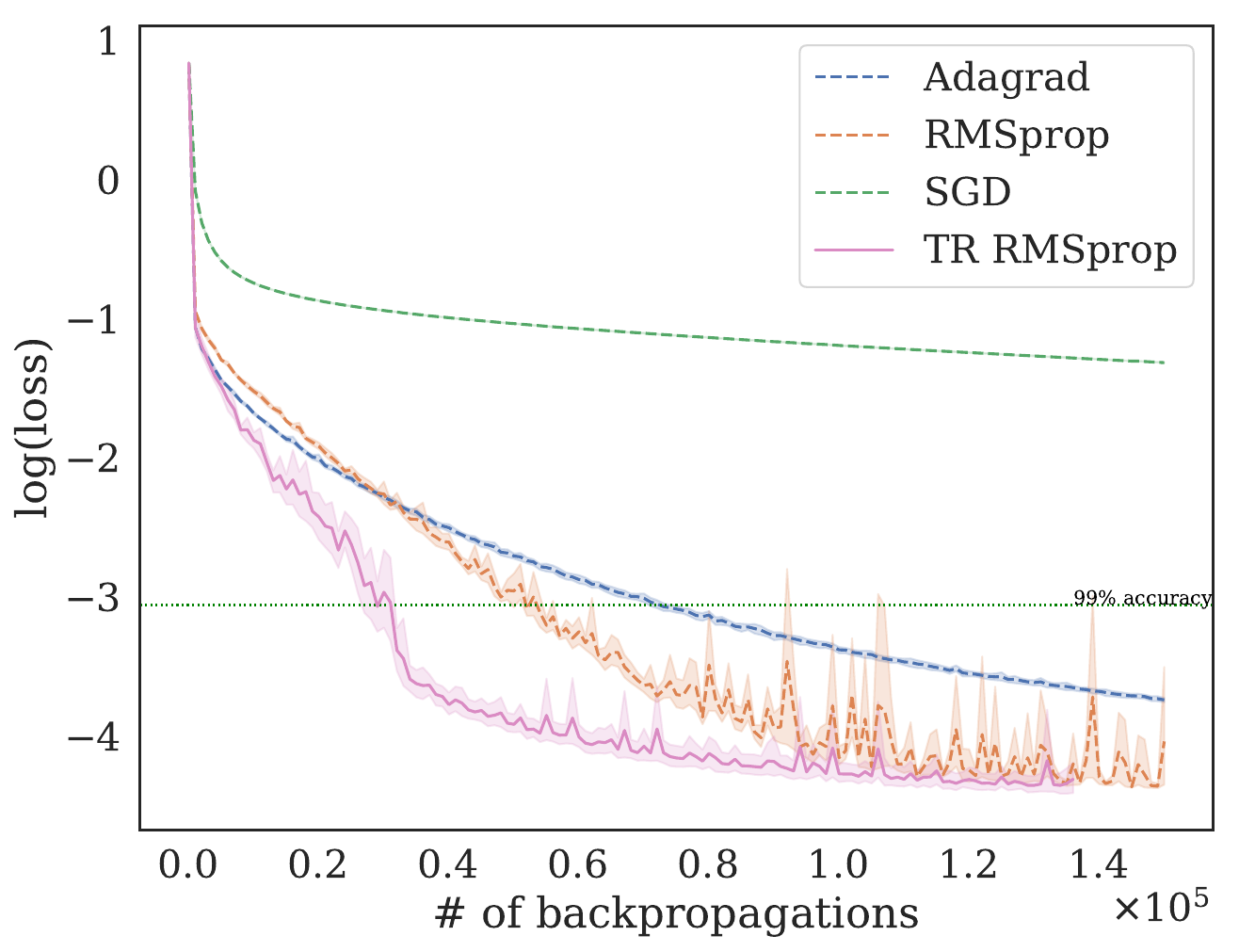}& 
             \includegraphics[width=0.29\linewidth,valign=c]{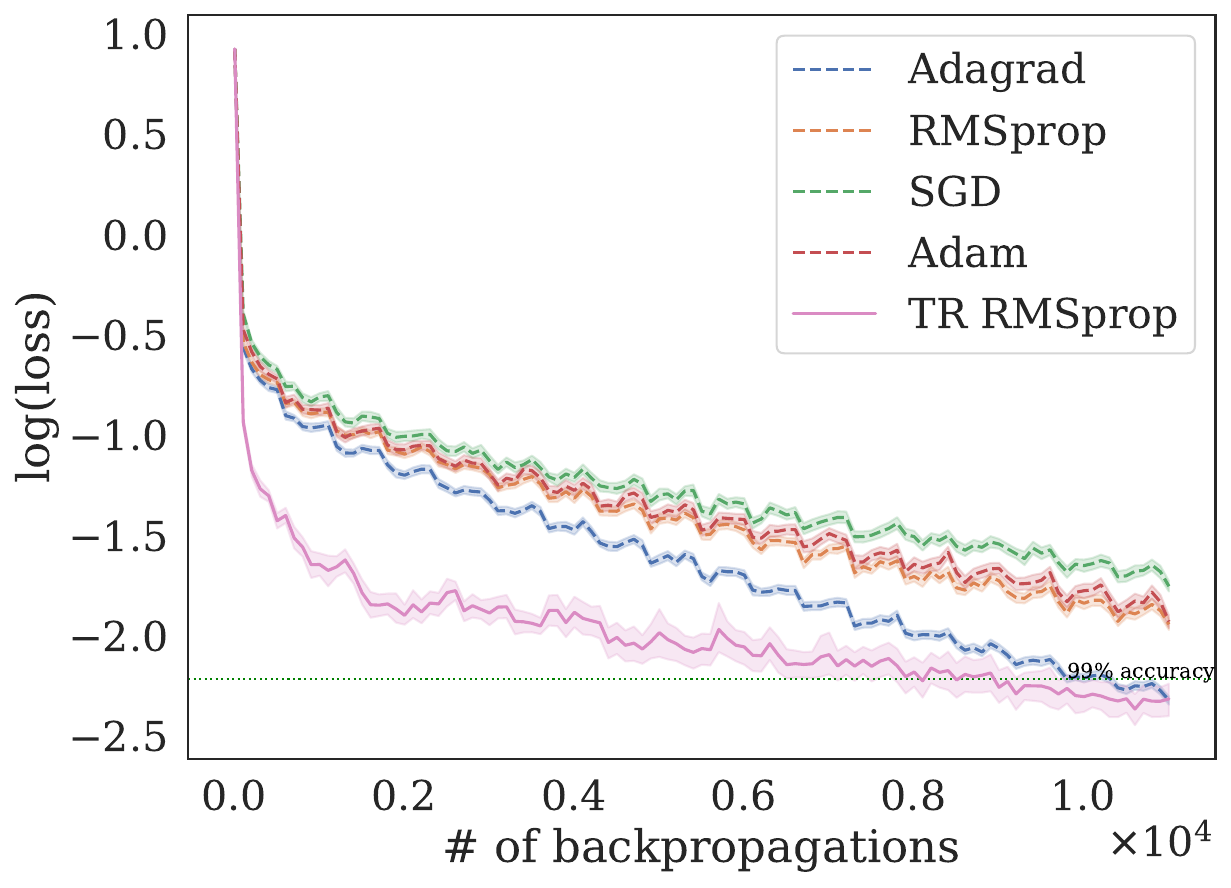} & \includegraphics[width=0.29\linewidth,valign=c]{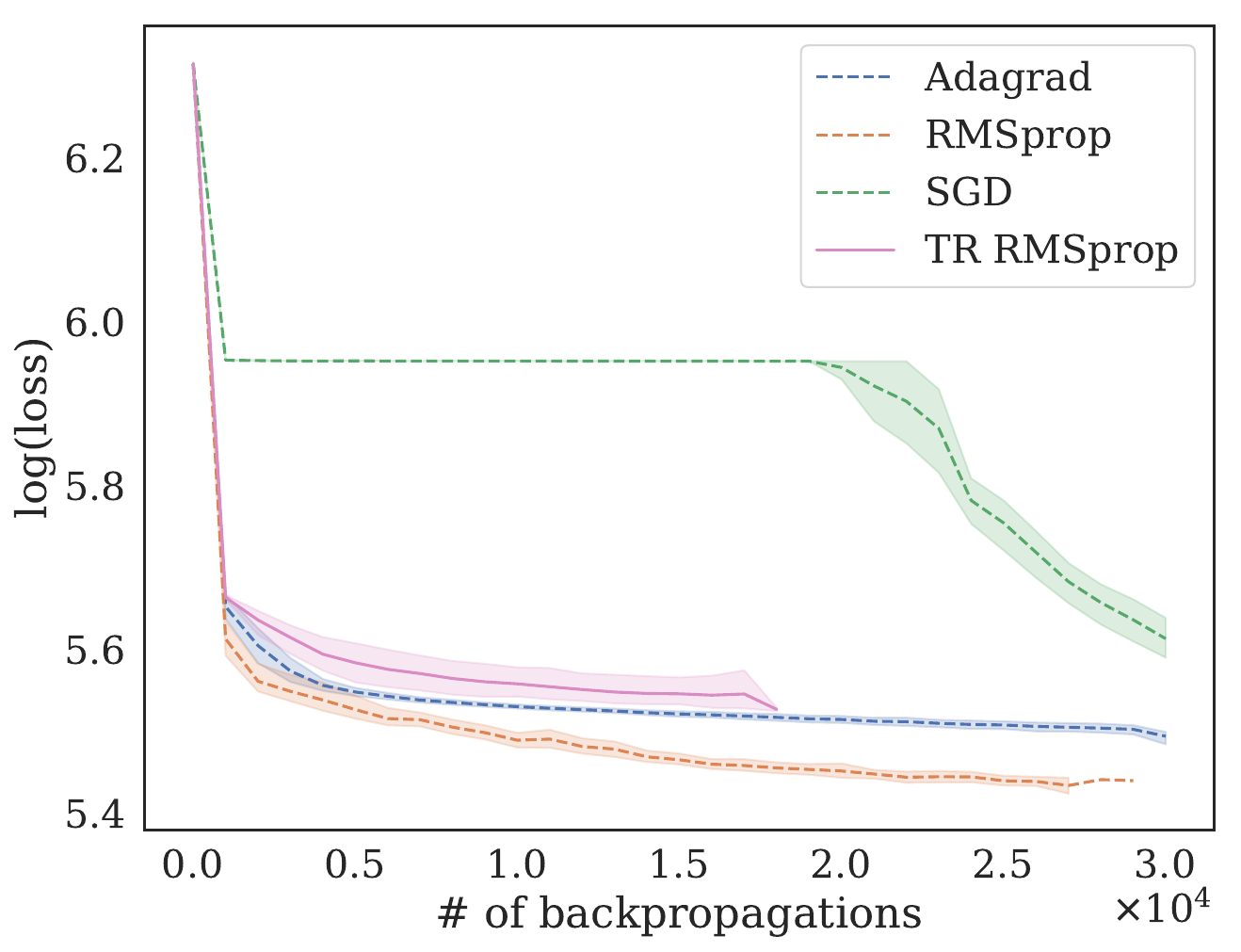}\\
             
             \rotatebox[origin=c]{90}{\footnotesize{CIFAR-10}}&
             \includegraphics[width=0.29\linewidth,valign=c]{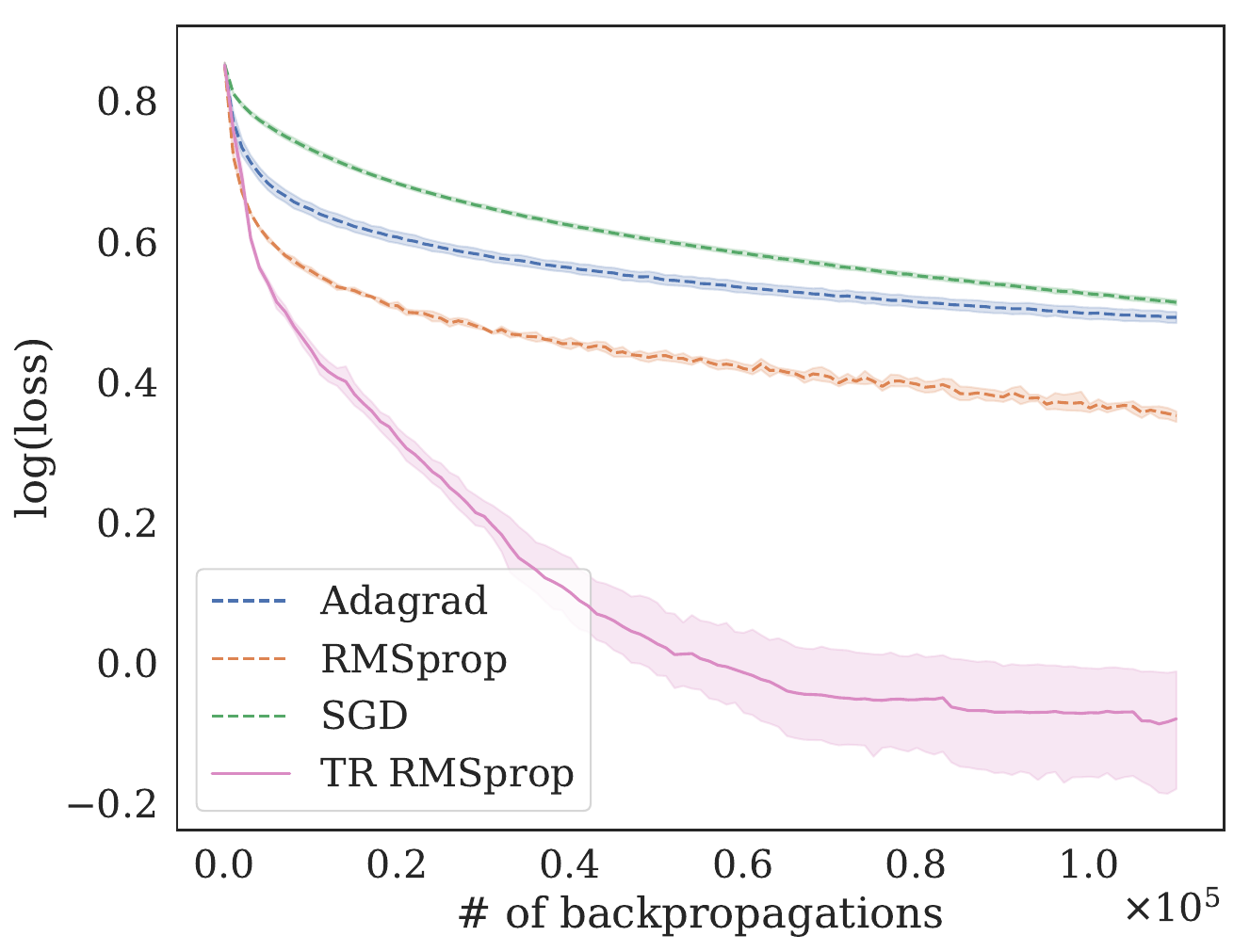}&
             \includegraphics[width=0.29\linewidth,valign=c]{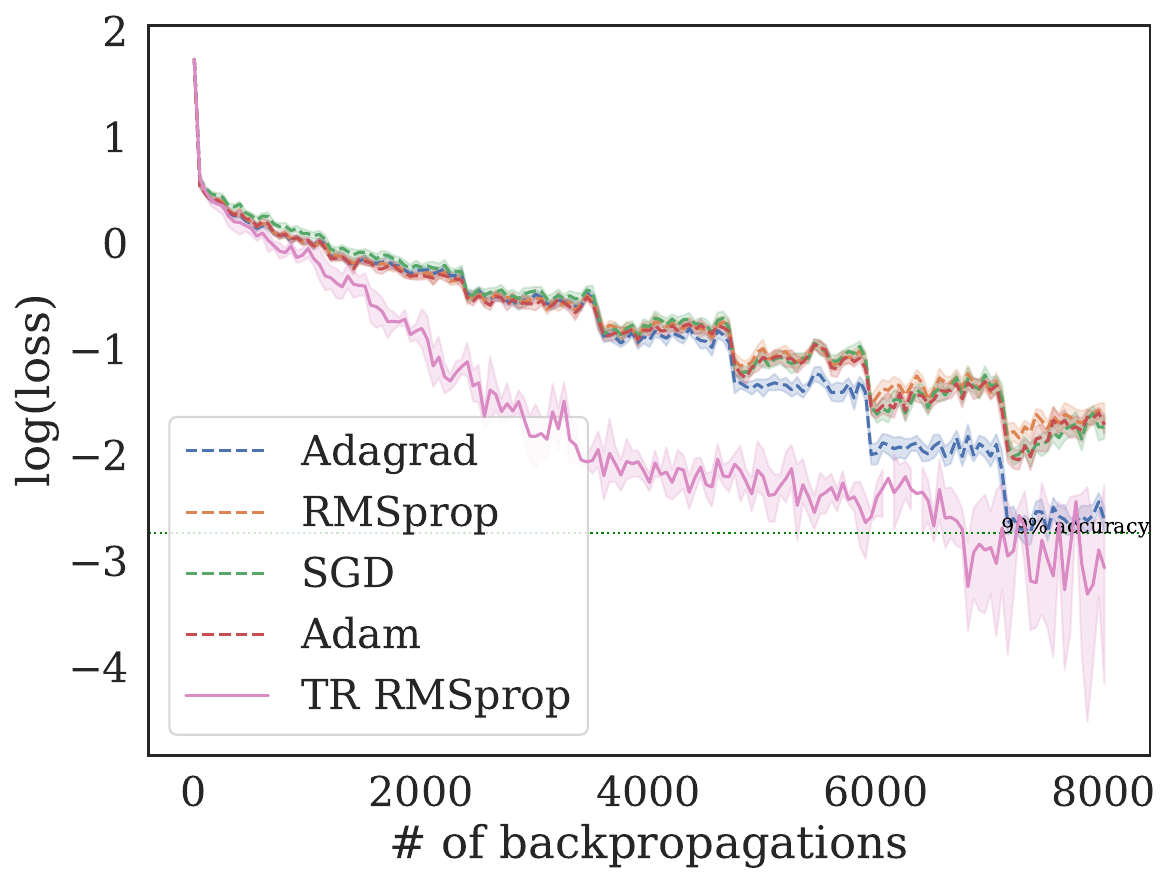} & \includegraphics[width=0.29\linewidth,valign=c]{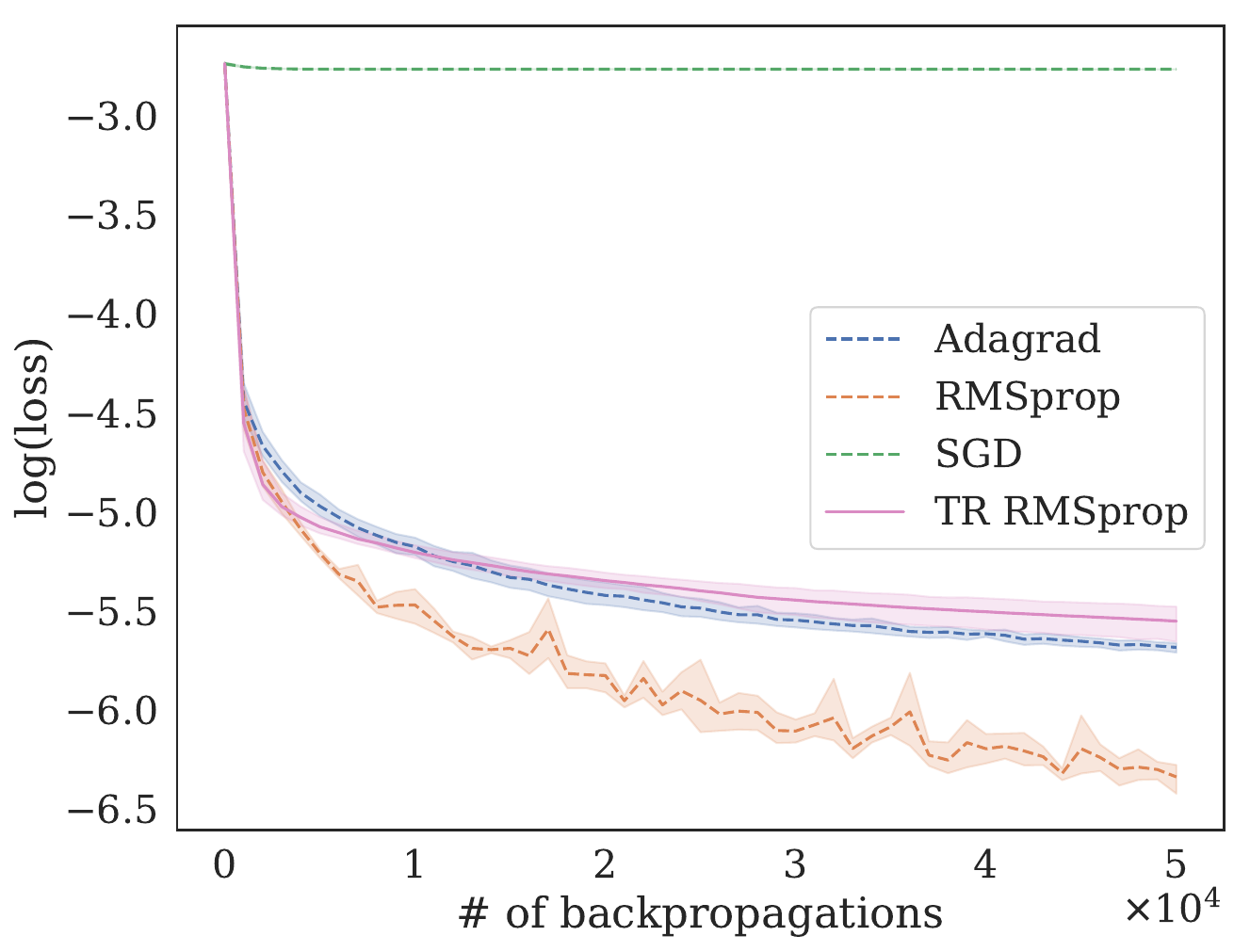} 
             
 	  \end{tabular}
          \caption{ \footnotesize{Log loss over backpropagations. Same setting as Figure~\ref{fig:2ndorder_results}. See Figure~\ref{fig:1storder_results_ep} for epoch results.}}
          \label{fig:1storder_results_bp}
\end{figure*}

\begin{figure}[H]
 \centering 
          \begin{tabular}{l@{}c@{}c@{}c@{}}
             & ResNet18 & Fully-Connected & Autoencoder \\
            \rotatebox[origin=c]{90}{CIFAR-10} &
             \includegraphics[width=0.275\linewidth,valign=c]{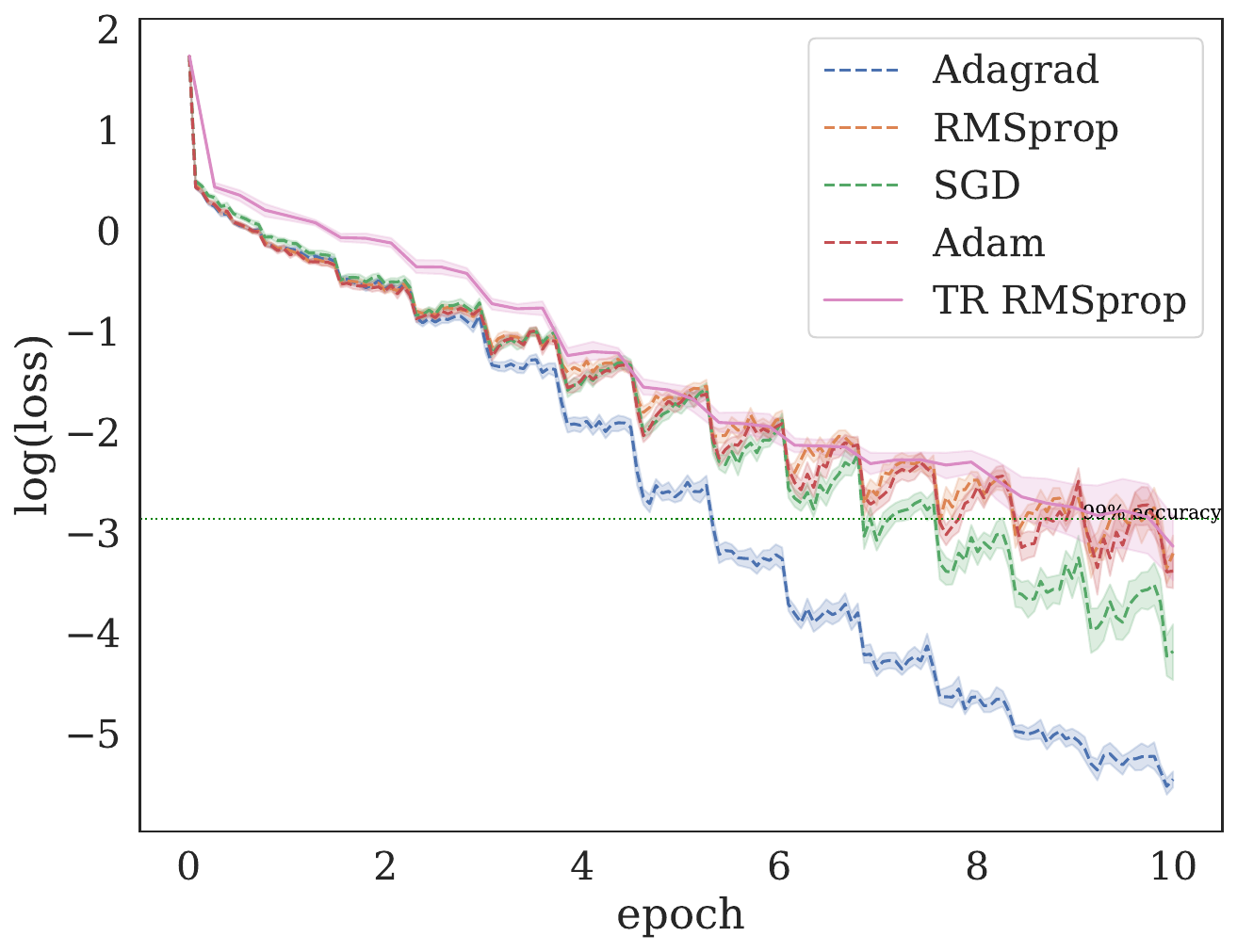}&
             \includegraphics[width=0.275\linewidth,valign=c]{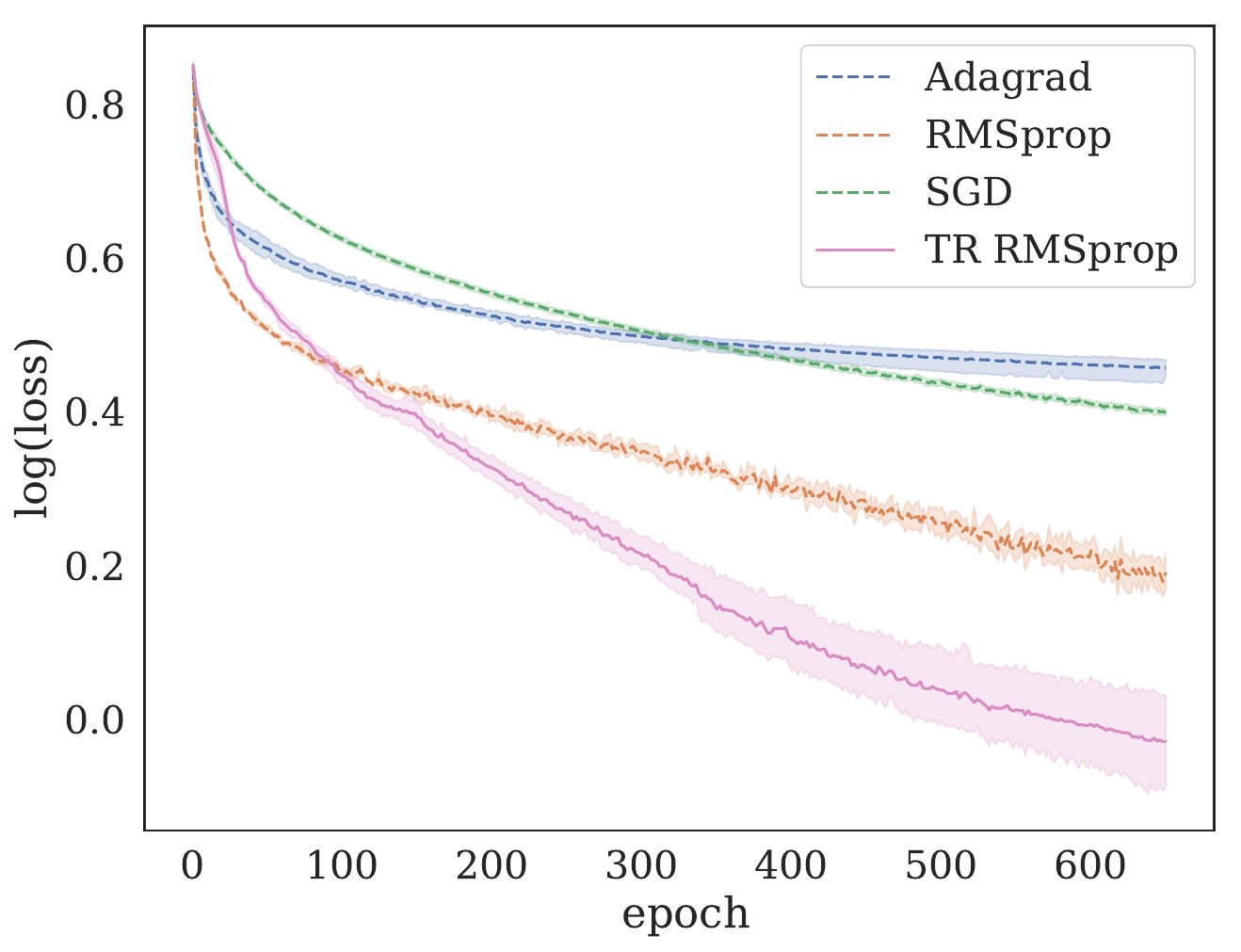}&
             \includegraphics[width=0.275\linewidth,valign=c]{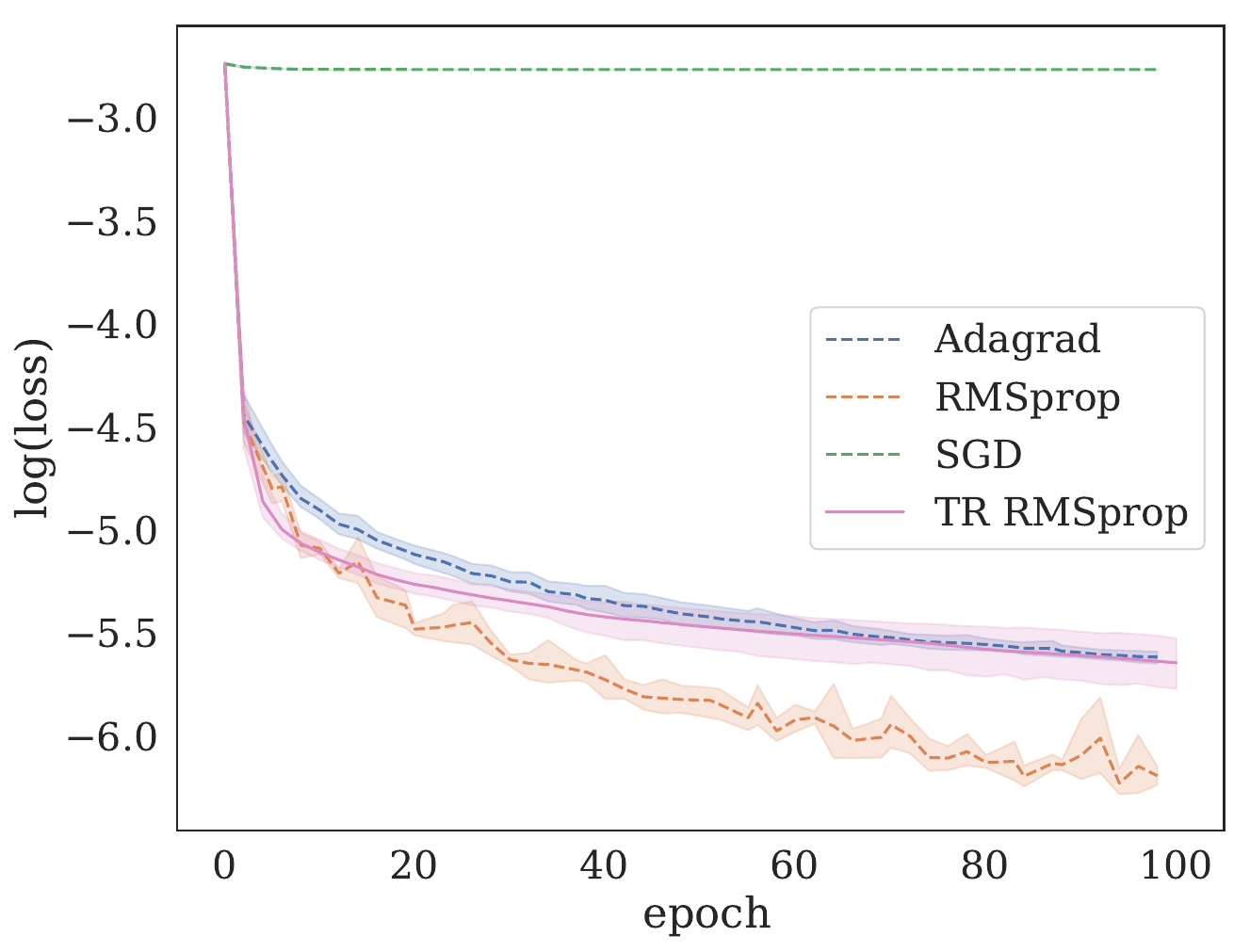}\\
              \rotatebox[origin=c]{90}{Fashion-MNIST} & 
             \includegraphics[width=0.275\linewidth,valign=c]{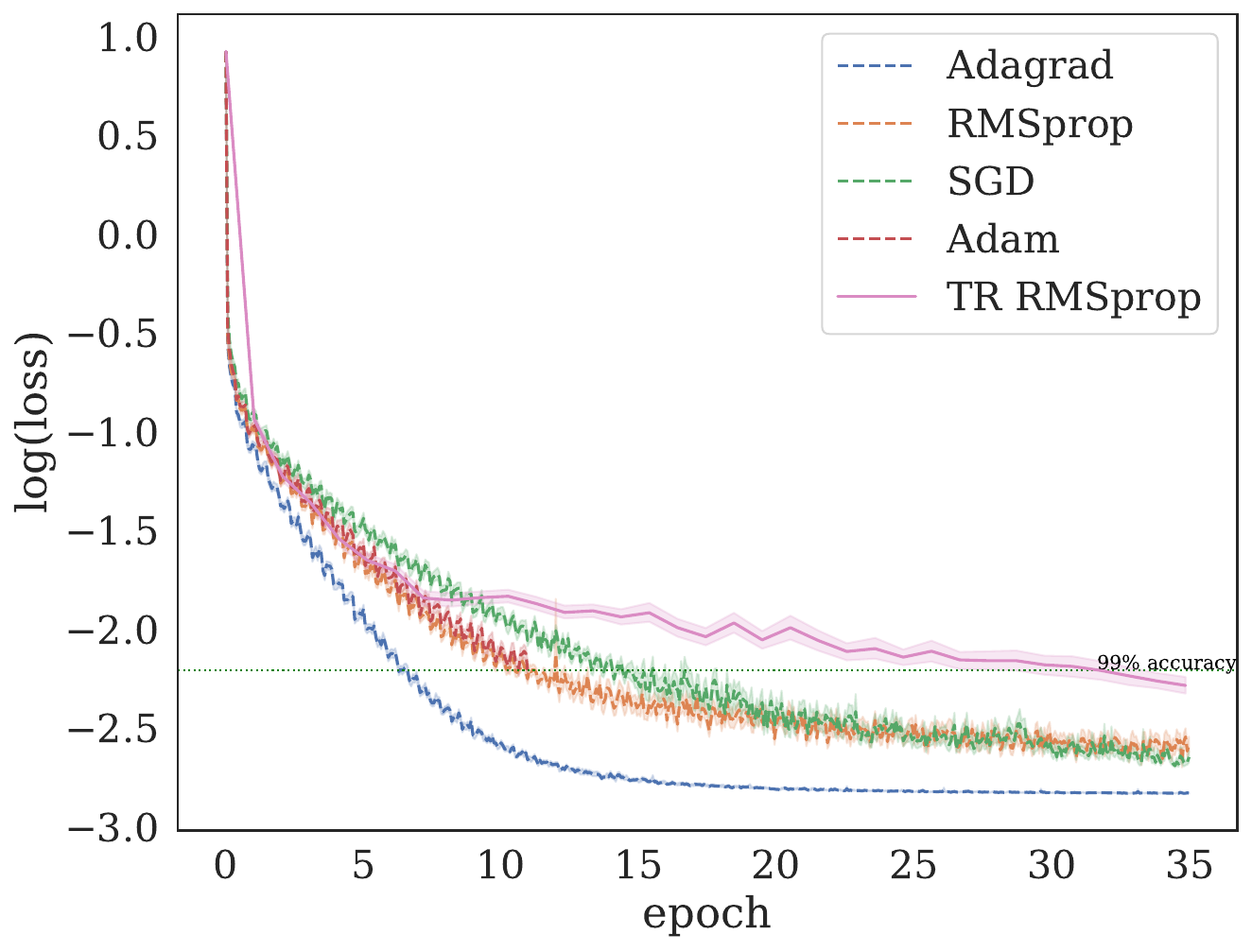}& 
             \includegraphics[width=0.275\linewidth,valign=c]{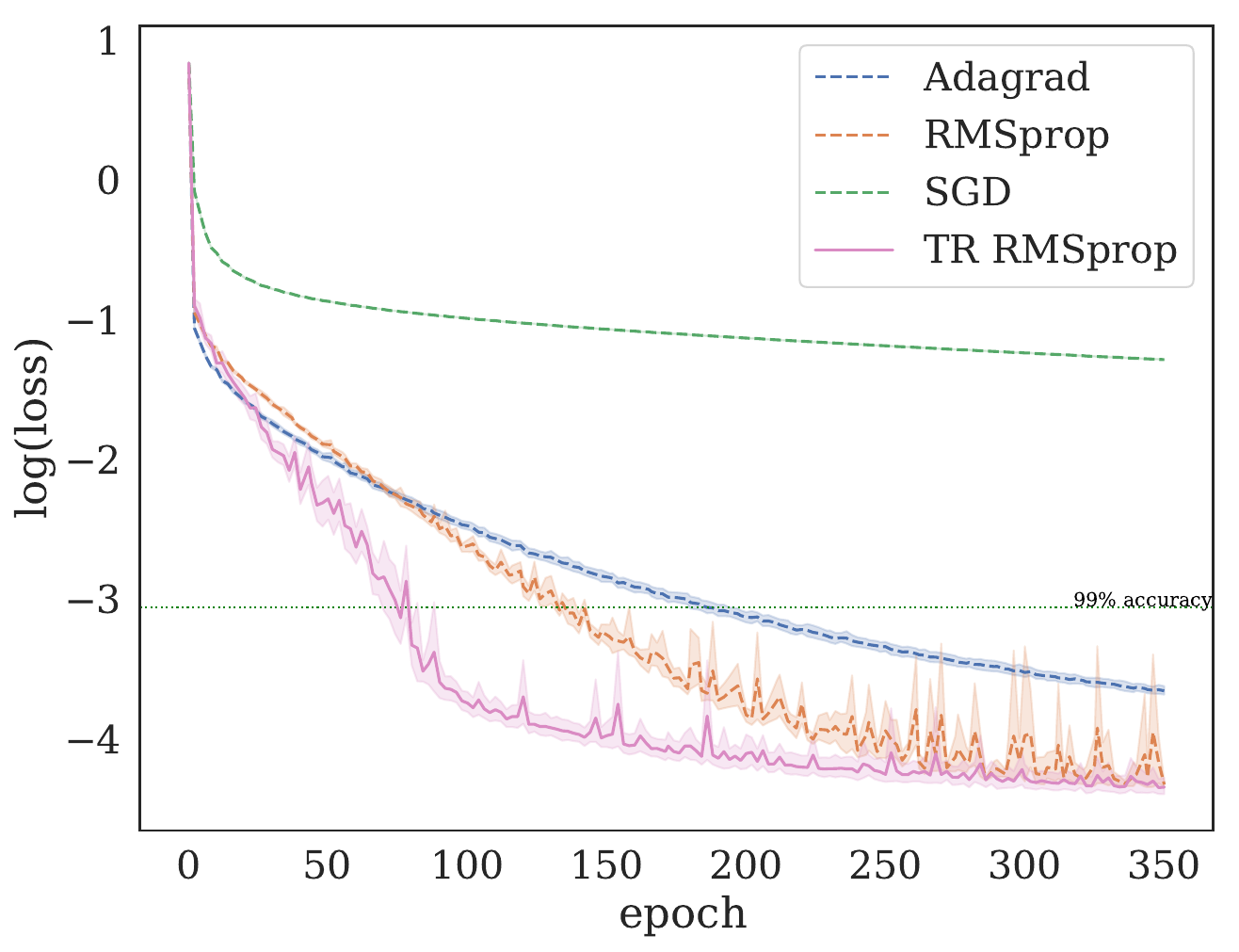}& 
             \includegraphics[width=0.275\linewidth,valign=c]{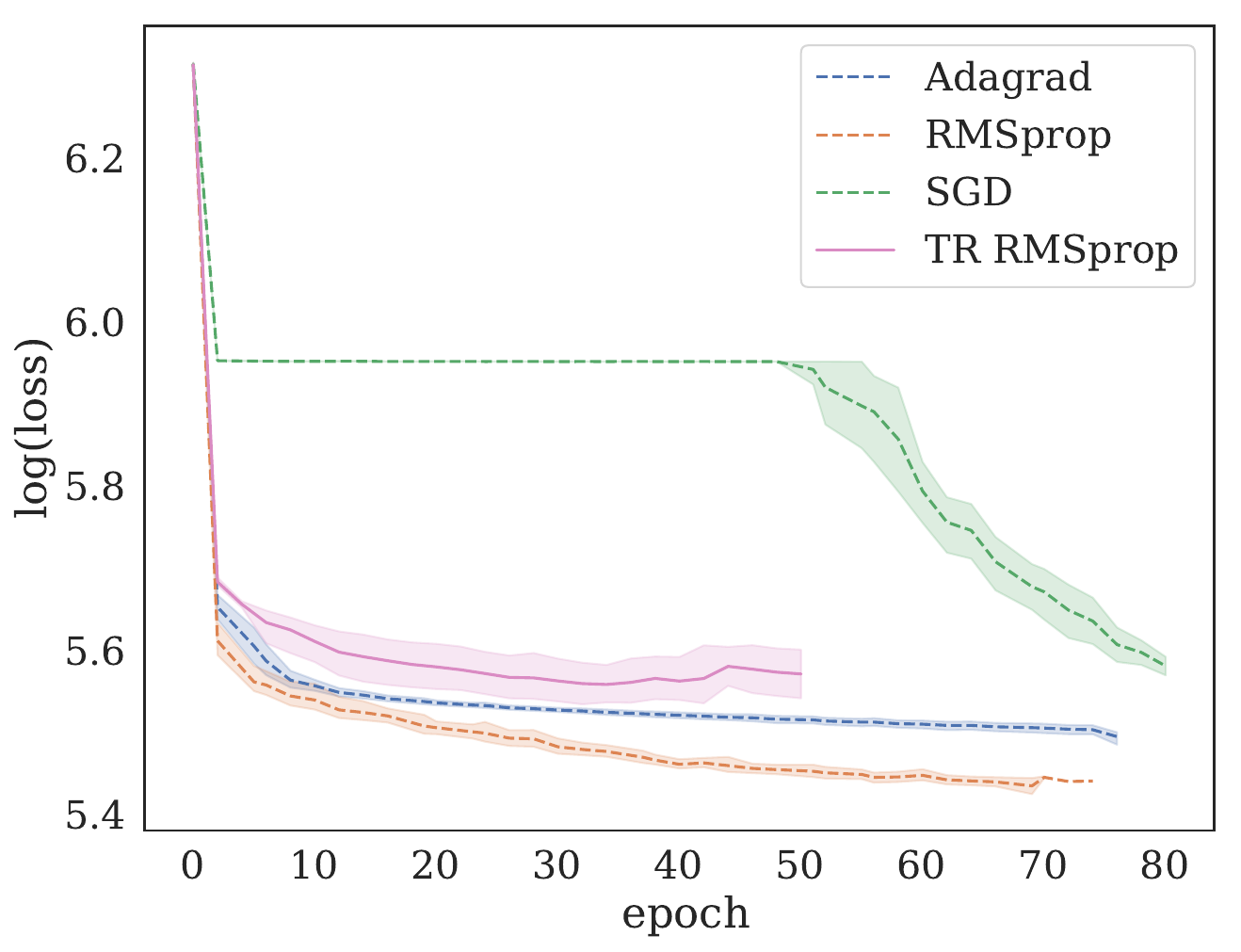}
 	  \end{tabular}
          \caption{Experiment comparing TR and gradient methods in terms of epochs. Average log loss as well as $95\%$ confidence interval shown.}\label{fig:1storder_results_ep}
          \end{figure}

\begin{figure}[H]
 \centering 
          \begin{tabular}{l@{}c@{}c@{}c@{}}
             & ResNet18 & Fully-Connected & Autoencoder \\
            \rotatebox[origin=c]{90}{CIFAR-10} &
             \includegraphics[width=0.275\linewidth,valign=c]{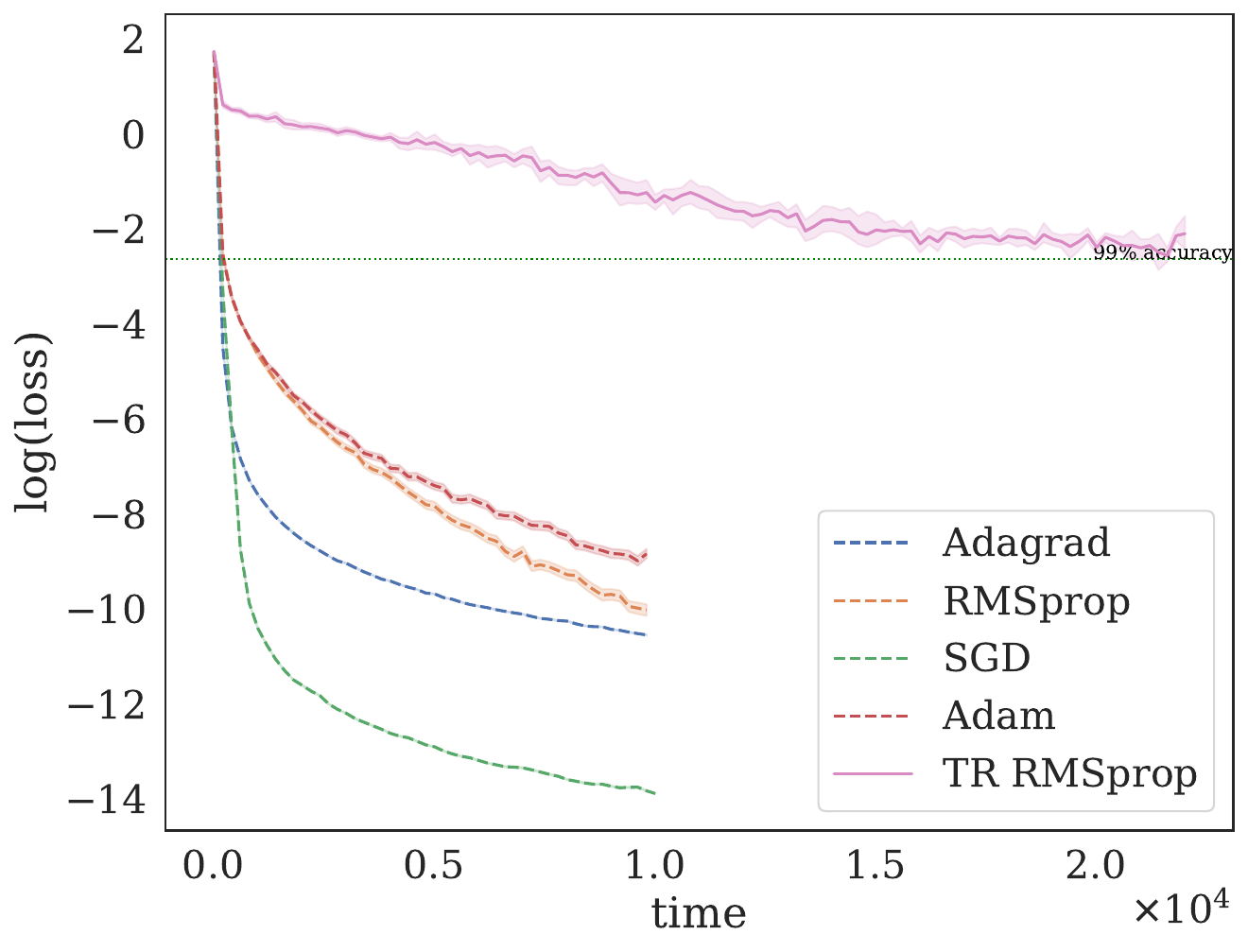}&
             \includegraphics[width=0.275\linewidth,valign=c]{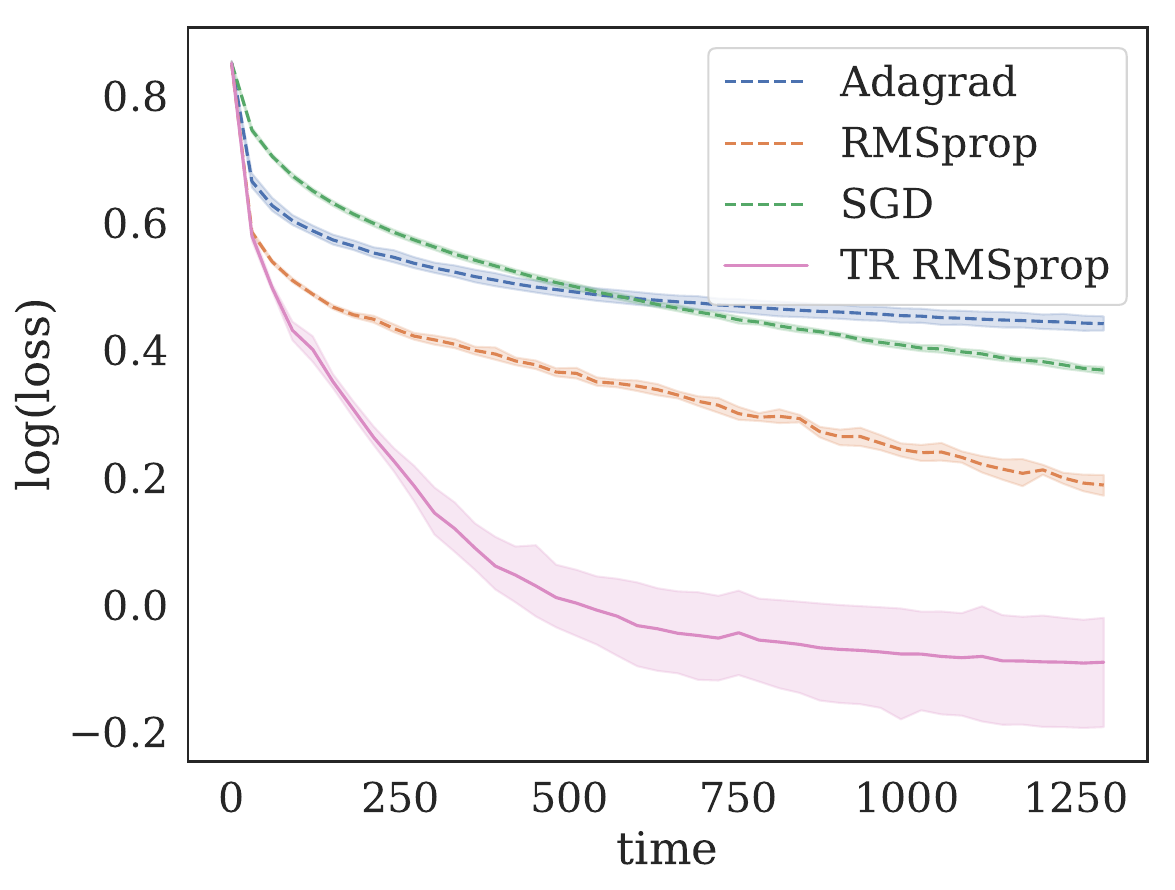}&
             \includegraphics[width=0.275\linewidth,valign=c]{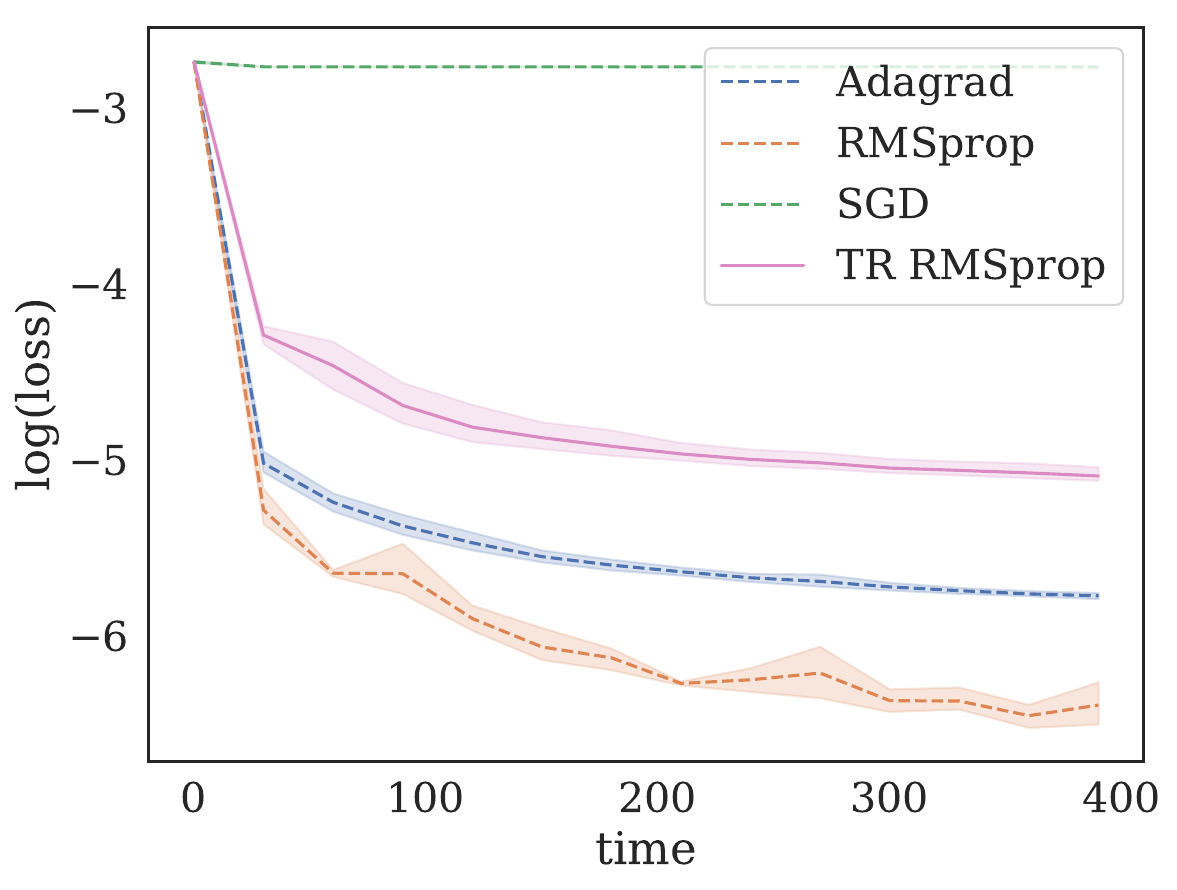}\\
              \rotatebox[origin=c]{90}{Fashion-MNIST} & 
             \includegraphics[width=0.275\linewidth,valign=c]{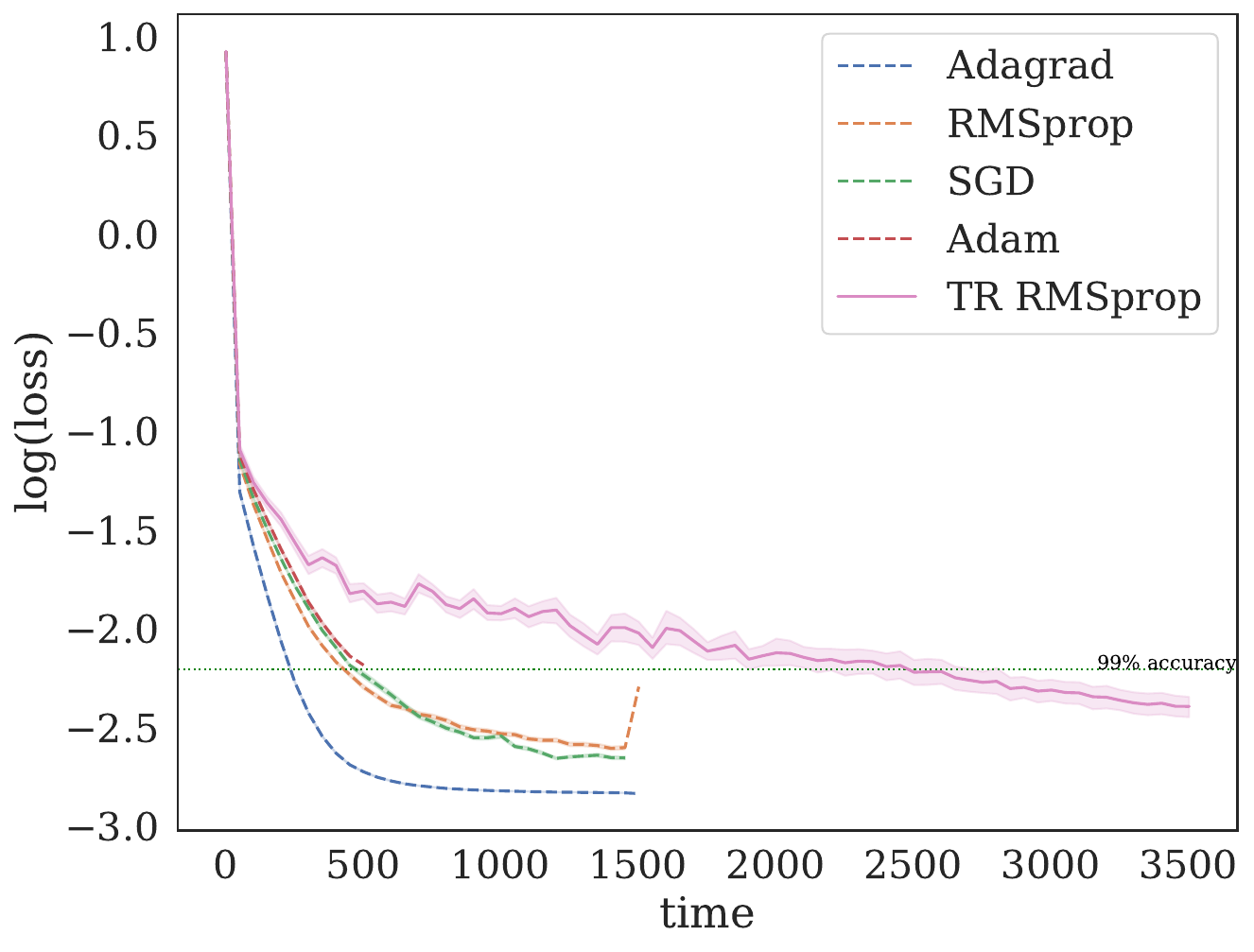}& 
             \includegraphics[width=0.275\linewidth,valign=c]{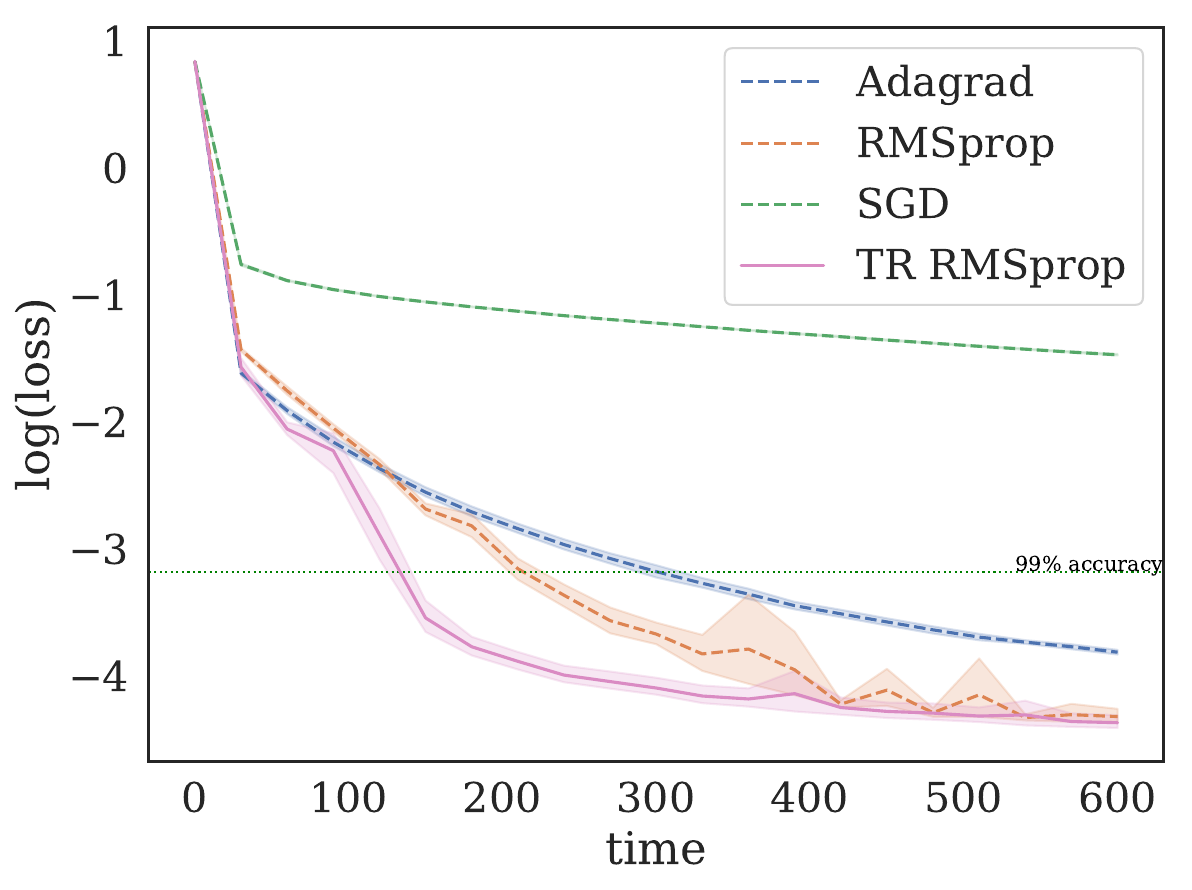}& 
             \includegraphics[width=0.275\linewidth,valign=c]{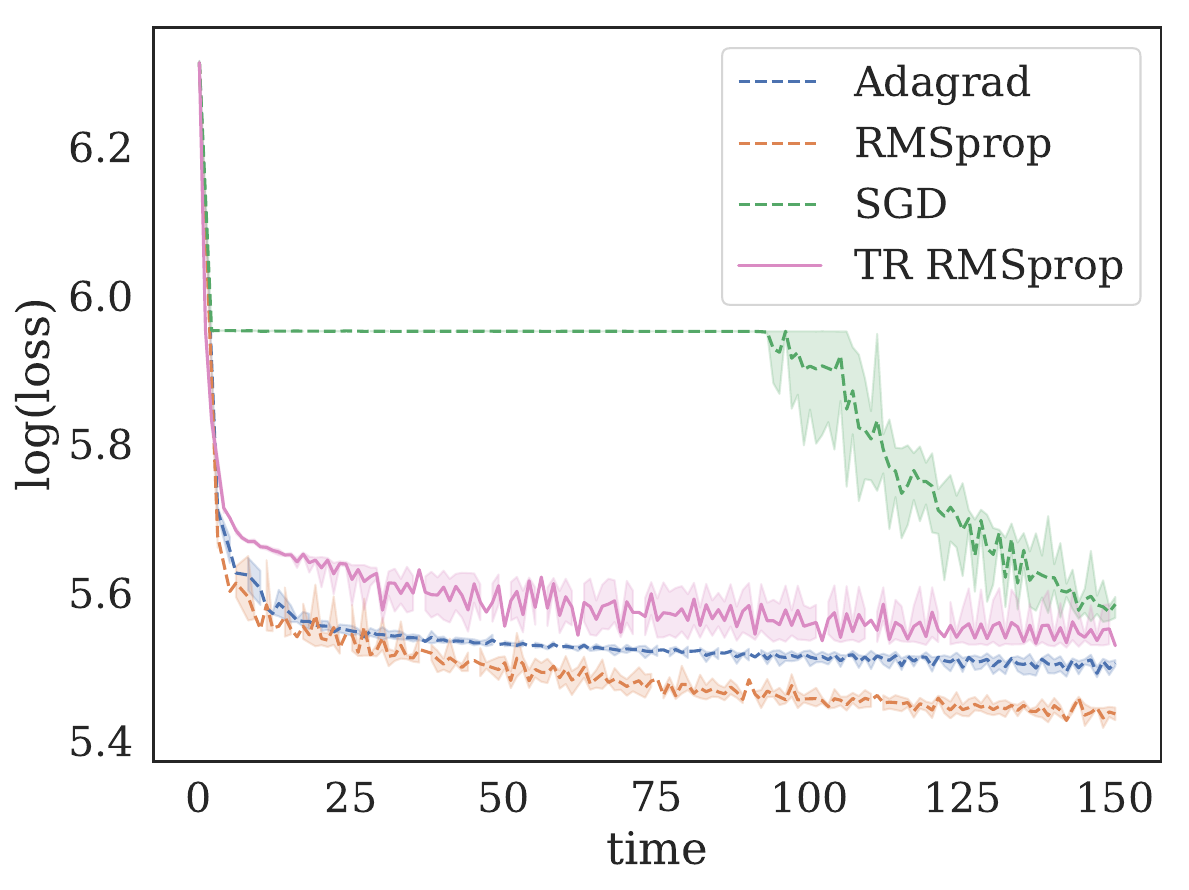}
 	  \end{tabular}
          \caption{Experiment comparing TR and gradient methods in terms of wall-clock time. Average log loss as well as $95\%$ confidence interval shown. The advantage of extremely low-iteration costs of first-order methods is particularly notable in the ResNet18 architecture due to the large network size. }\label{fig:1storder_results_time}
          \end{figure}

To put the previous results into context, we also benchmark several state-of-the-art gradient methods. We fix their sample size to 32 (as advocated e.g. in \cite{masters2018revisiting}) but grid search the stepsize, since it is the ratio of these two quantities that effectively determines the level of stochasticity~\citep{jastrzkebski2017three}.  The TR methods use a batch size of 128 for the ResNet architecture and 512 otherwise\footnote{We observed weaker performance when running with smaller batches, presumably because second-order methods are likely to "overfit" noise in small batches in any given iteration as they extract more information of each batch per step by computing curvature.}. For a fair comparison, we thus report results in terms of number of backpropagations, epochs and time . The findings are mixed: For small nets such as the MLPs the TR method with RMSProp ellipsoids is superior in all metrics, even when benchmarked in terms of time. However, while Fig. \ref{fig:1storder_results_bp} indicates that ellipsoidal TR methods are slightly superior in terms of backpropagations even for bigger nets (ResNets and Autoencoders), a close look at the Figures \ref{fig:1storder_results_ep} and \ref{fig:1storder_results_time} (App. C) reveals that they at best manage to keep pace with first-order methods in terms of epochs and are inferior in time. Furthermore, only the autoencoders give rise to a saddle point, which adaptive gradient methods escape faster than vanilla SGD, just like it was the case for second-order methods (see Fig. \ref{fig:2ndorder_results}).

\hypertarget{app_sec:further}{ }
\section{Default parameters, architectures and datasets} \label{sec:params}
\paragraph{Parameters} Table \ref{tab:params} reports the default parameters we consider. Only for the larger ResNet18 on CIFAR-10, we adapted the batch size to $128$ due to memory constraints.
\begin{table}[H]
\centering
\begin{tabular}{l|lllllllc}
        & $|\mathcal{S}_0|$  & $\Delta_0$ & $\Delta_{\text{max}}$& $\eta_1$ & $\eta_2$ & $\gamma_1$ & $\gamma_2$  & $\kappa_K$ (krylov tol.)  \\\hline

TR$_{\text{uni}}$ & 512                      &        $10^{-4}$ &10    & $10^{-4}$        &     0.95     &      1.1      &     1.5        &0.1\\
TR$_{\text{ada}}$ & 512                      &       $10^{-4}$     &10 &  $10^{-4}$        &     0.95     &       1.1     &      1.5       &0.1 \\
TR$_{\text{rms}}$ & 512   &          $10^{-4}$ &10        &    $10^{-4}$    &    0.95      &      1.1      &     1.75       &     0.1
\end{tabular}
\caption{Default parameters}
\label{tab:params}
\end{table}

\paragraph{Datasets} We use two real-world datasets for image classification, namely CIFAR-10 and Fashion-MNIST\footnote{Both datasets were accessed from https://www.tensorflow.org/api\_docs/python/tf/keras/datasets}. While Fashion-MNIST consists of greyscale $28\times28$ images, CIFAR-10 are colored images of size $32\times32$. Both datasets have a fixed training-test split consisting of 60,000 and 10,000 images, respectively.

\paragraph{Network architectures} The MLP architectures are simple. For MNIST and Fashion-MNIST we use a $784-128-10$ network with tanh activations and a cross entropy loss. The networks has $101'770$ parameters. For the CIFAR-10 MLP we use a $3072-128-128-10$ architecture also with tanh activations and cross entropy loss. This network has $410'880$ parameters.

The Fashion-MNIST autoencoder has the same architecture as the one used in \cite{hinton2006reducing,xu2017second,martens2010deep,martens2015optimizing}. 
The encoder structure is $784-1000-500-250-30$ and the decoder is mirrored. Sigmoid activations are used in all but the central layer. The reconstructed images are fed pixelwise into a binary cross entropy loss. The network has a total of $2'833'000$ parameters. 
The CIFAR-10 autoencoder is taken from the implementation of https://github.com/jellycsc/PyTorch-CIFAR-10-autoencoder.

For the ResNet18, we used the implementation from \href{https://pytorch.org/docs/stable/torchvision/models.html}{torchvision} for CIFAR-10 as well as a \href{https://zablo.net/blog/post/using-resnet-for-mnist-in-pytorch-tutorial/}{modification} of it for Fashion-MNIST that adapts the first convolution to account for the single input channel.

In all of our experiments each method was run on one Tesla P100 GPU using the PyTorch \citep{paszke2017automatic} library.

           
\section{Reconstructed Images from Autoencoders}
\begin{figure}[H]
\centering
\hspace*{-1cm}
          \begin{tabular}{c@{}|c@{}|c@{}|c@{}|c@{}|c@{}|c@{}}
          
          Original& SGD & Adagrad & Rmsprop & TR Uniform & TR Adagrad & TR RMSprop \\ \hline
             \includegraphics[width=0.0476\linewidth,valign=c]{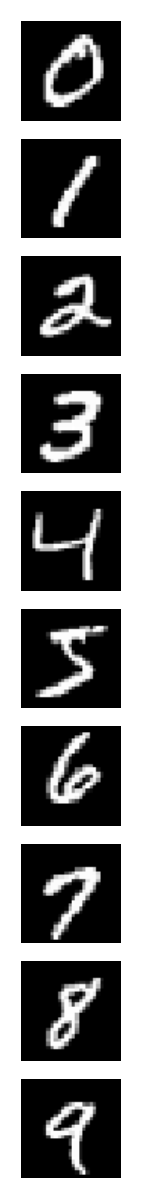}
             \includegraphics[width=0.0476\linewidth,valign=c]{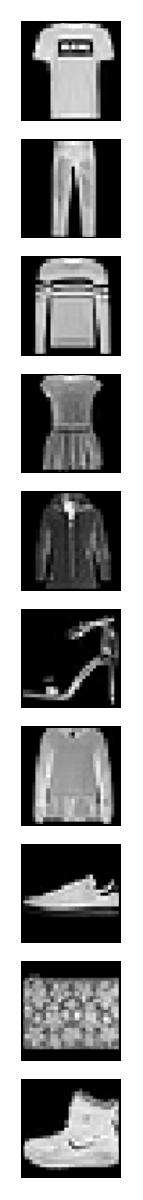}
             \includegraphics[width=0.0476\linewidth,valign=c]{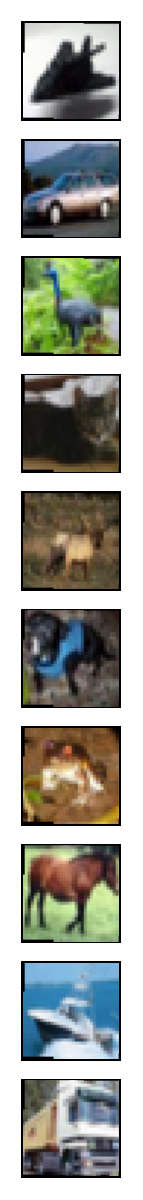}&
             \includegraphics[width=0.0476\linewidth,valign=c]{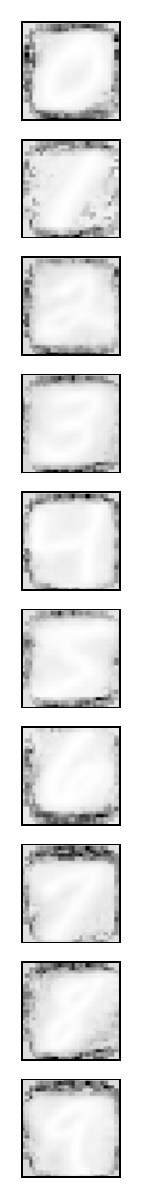}
             \includegraphics[width=0.0476\linewidth,valign=c]{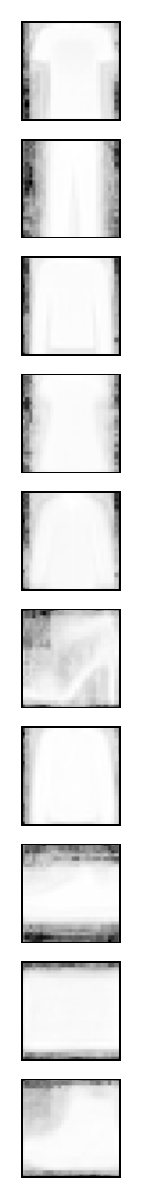}
             \includegraphics[width=0.0476\linewidth,valign=c]{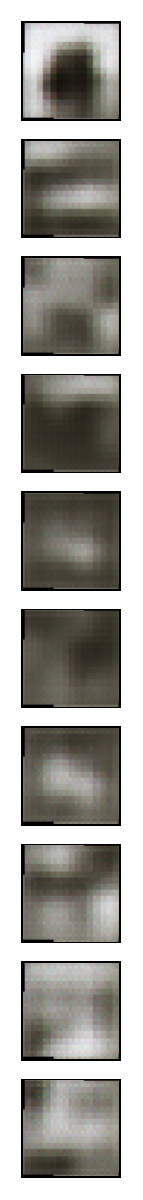}&
            \includegraphics[width=0.0476\linewidth,valign=c]{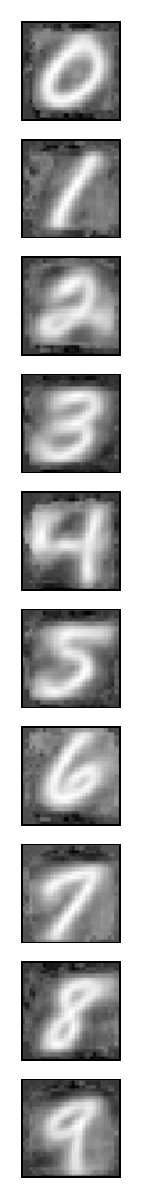}
            \includegraphics[width=0.0476\linewidth,valign=c]{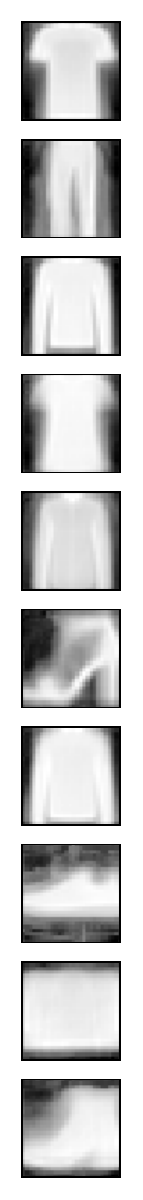}
            \includegraphics[width=0.0476\linewidth,valign=c]{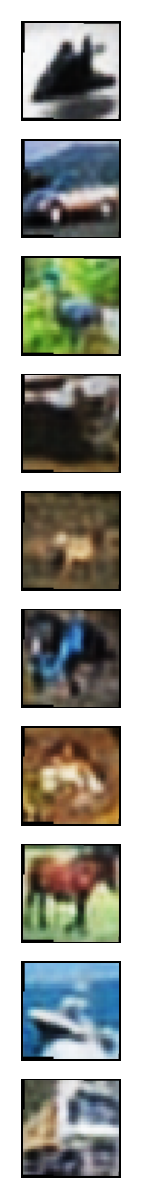}&
            \includegraphics[width=0.0476\linewidth,valign=c]{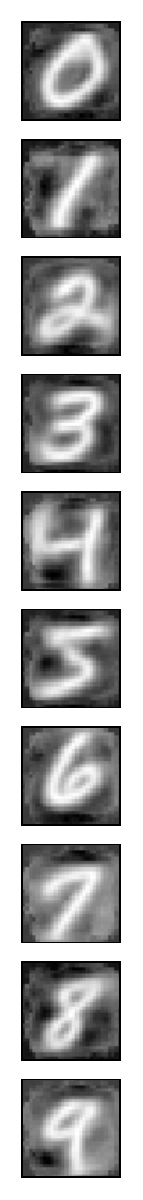}
            \includegraphics[width=0.0476\linewidth,valign=c]{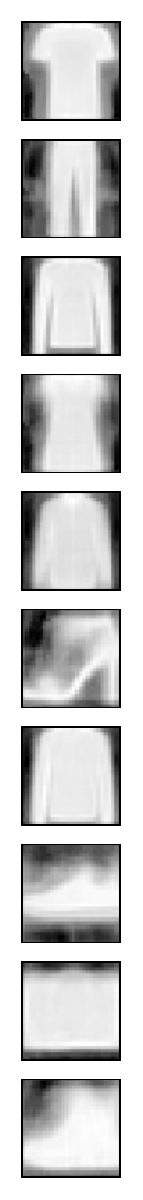}
            \includegraphics[width=0.0476\linewidth,valign=c]{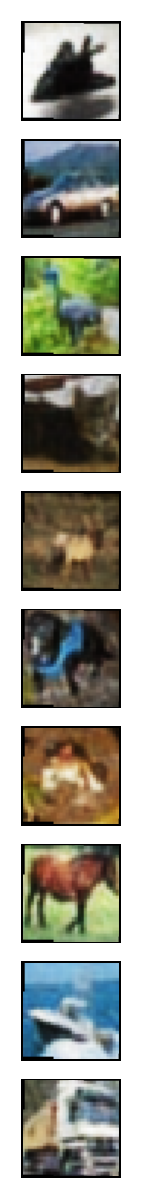}&
            \includegraphics[width=0.0476\linewidth,valign=c]{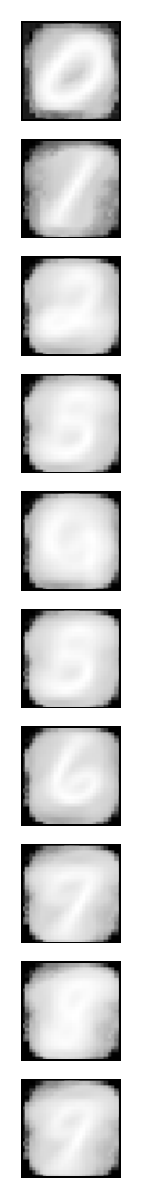}
            \includegraphics[width=0.0476\linewidth,valign=c]{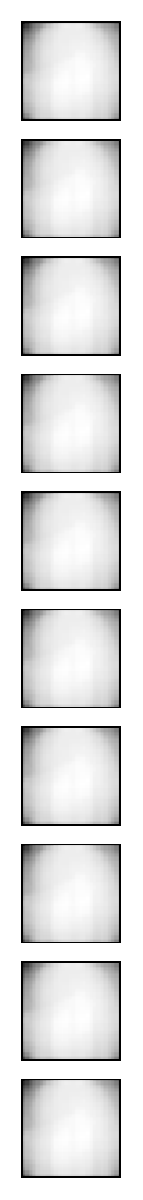}
            \includegraphics[width=0.0476\linewidth,valign=c]{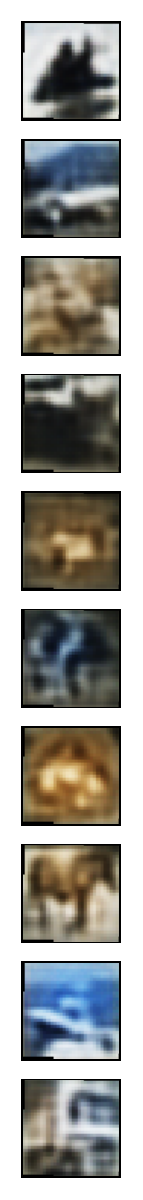}&
            \includegraphics[width=0.0476\linewidth,valign=c]{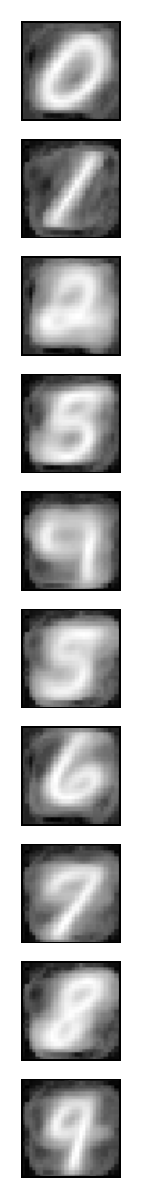}
            \includegraphics[width=0.0476\linewidth,valign=c]{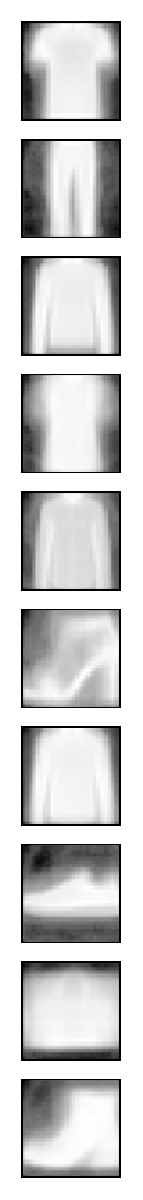}
            \includegraphics[width=0.0476\linewidth,valign=c]{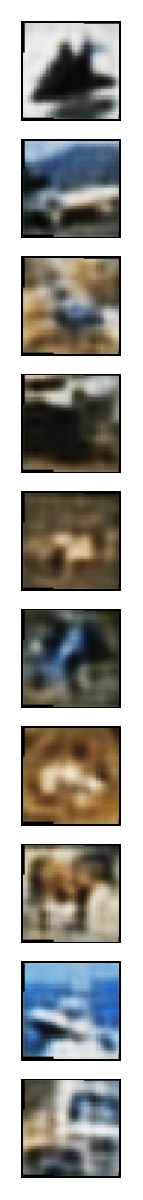}&
            \includegraphics[width=0.0476\linewidth,valign=c]{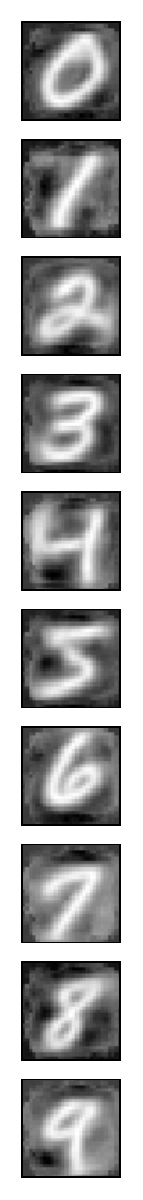}
            \includegraphics[width=0.0476\linewidth,valign=c]{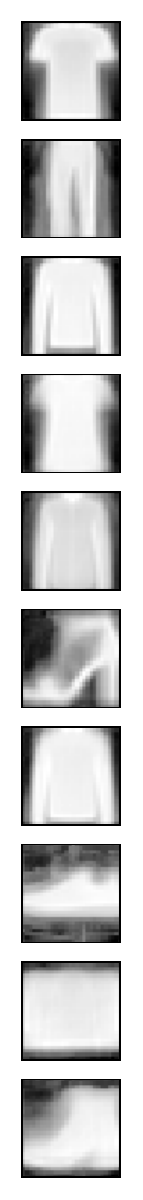}
            \includegraphics[width=0.0476\linewidth,valign=c]{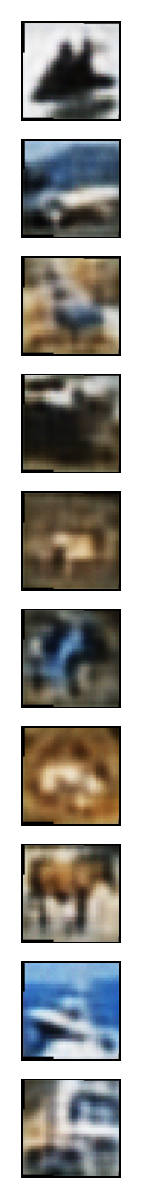}
	  \end{tabular}
          \caption{\footnotesize{Original and reconstructed MNIST digits (left), Fashion-MNIST items (middle), and CIFAR-10 classes (right) for different optimization methods after convergence.}}
     \label{fig:reconstruct}
\end{figure}

%% file: ada_norms.bbl
\begin{thebibliography}{10}

\bibitem{agarwal2018case}
Naman Agarwal, Brian Bullins, Xinyi Chen, Elad Hazan, Karan Singh, Cyril Zhang,
  and Yi~Zhang.
\newblock The case for full-matrix adaptive regularization.
\newblock {\em arXiv preprint arXiv:1806.02958}, 2018.

\bibitem{alain2018negative}
Guillaume Alain, Nicolas Le~Roux, and Pierre-Antoine Manzagol.
\newblock Negative eigenvalues of the hessian in deep neural networks.
\newblock 2018.

\bibitem{allen2017katyusha}
Zeyuan Allen-Zhu.
\newblock Katyusha: The first direct acceleration of stochastic gradient
  methods.
\newblock {\em The Journal of Machine Learning Research}, 18(1):8194--8244,
  2017.

\bibitem{allen2018convergence}
Zeyuan Allen-Zhu, Yuanzhi Li, and Zhao Song.
\newblock A convergence theory for deep learning via over-parameterization.
\newblock {\em arXiv preprint arXiv:1811.03962}, 2018.

\bibitem{amari1998natural}
Shun-Ichi Amari.
\newblock Natural gradient works efficiently in learning.
\newblock {\em Neural computation}, 10(2):251--276, 1998.

\bibitem{becker1988improving}
Sue Becker, Yann Le~Cun, et~al.
\newblock Improving the convergence of back-propagation learning with second
  order methods.
\newblock In {\em Proceedings of the 1988 connectionist models summer school},
  pages 29--37. San Matteo, CA: Morgan Kaufmann, 1988.

\bibitem{bekas2007estimator}
Costas Bekas, Effrosyni Kokiopoulou, and Yousef Saad.
\newblock An estimator for the diagonal of a matrix.
\newblock {\em Applied numerical mathematics}, 57(11-12):1214--1229, 2007.

\bibitem{blanchet2016convergence}
Jose Blanchet, Coralia Cartis, Matt Menickelly, and Katya Scheinberg.
\newblock Convergence rate analysis of a stochastic trust region method for
  nonconvex optimization.
\newblock {\em arXiv preprint arXiv:1609.07428}, 2016.

\bibitem{bottou2010large}
Leon Bottou.
\newblock Large-scale machine learning with stochastic gradient descent.
\newblock In {\em Proceedings of COMPSTAT'2010}, pages 177--186. Springer,
  2010.

\bibitem{bottou2018optimization}
L{\'e}on Bottou, Frank~E Curtis, and Jorge Nocedal.
\newblock Optimization methods for large-scale machine learning.
\newblock {\em SIAM Review}, 60(2):223--311, 2018.

\bibitem{brutzkus2017globally}
Alon Brutzkus and Amir Globerson.
\newblock Globally optimal gradient descent for a convnet with gaussian inputs.
\newblock {\em arXiv preprint arXiv:1702.07966}, 2017.

\bibitem{carmon2017lower}
Yair Carmon, John~C Duchi, Oliver Hinder, and Aaron Sidford.
\newblock Lower bounds for finding stationary points i.
\newblock {\em arXiv preprint arXiv:1710.11606}, 2017.

\bibitem{cartis2012complexity}
Coralia Cartis, Nicholas~IM Gould, and Ph~L Toint.
\newblock Complexity bounds for second-order optimality in unconstrained
  optimization.
\newblock {\em Journal of Complexity}, 28(1):93--108, 2012.

\bibitem{cartis2011adaptive}
Coralia Cartis, Nicholas~IM Gould, and Philippe~L Toint.
\newblock Adaptive cubic regularisation methods for unconstrained optimization.
  part i: motivation, convergence and numerical results.
\newblock {\em Mathematical Programming}, 127(2):245--295, 2011.

\bibitem{cartis2012much}
Coralia Cartis, Nicholas~IM Gould, and Philippe~L Toint.
\newblock {\em How Much Patience to You Have?: A Worst-case Perspective on
  Smooth Noncovex Optimization}.
\newblock Science and Technology Facilities Council Swindon, 2012.

\bibitem{cartis2017global}
Coralia Cartis and Katya Scheinberg.
\newblock Global convergence rate analysis of unconstrained optimization
  methods based on probabilistic models.
\newblock {\em Mathematical Programming}, pages 1--39, 2017.

\bibitem{chapelle2011improved}
Olivier Chapelle and Dumitru Erhan.
\newblock Improved preconditioner for hessian free optimization.
\newblock In {\em NIPS Workshop on Deep Learning and Unsupervised Feature
  Learning}, volume 201, 2011.

\bibitem{chen2018stochastic}
Ruobing Chen, Matt Menickelly, and Katya Scheinberg.
\newblock Stochastic optimization using a trust-region method and random
  models.
\newblock {\em Mathematical Programming}, 169(2):447--487, 2018.

\bibitem{conn2000trust}
Andrew~R Conn, Nicholas~IM Gould, and Philippe~L Toint.
\newblock {\em Trust region methods}.
\newblock SIAM, 2000.

\bibitem{curtis2017exploiting}
Frank~E Curtis and Daniel~P Robinson.
\newblock Exploiting negative curvature in deterministic and stochastic
  optimization.
\newblock {\em arXiv preprint arXiv:1703.00412}, 2017.

\bibitem{curtis2017trust}
Frank~E Curtis, Daniel~P Robinson, and Mohammadreza Samadi.
\newblock A trust region algorithm with a worst-case iteration complexity of
  $\mathcal{O}(\epsilon^{3-2}) $ for nonconvex optimization.
\newblock {\em Mathematical Programming}, 162(1-2):1--32, 2017.

\bibitem{daneshmand2018escaping}
Hadi Daneshmand, Jonas Kohler, Aurelien Lucchi, and Thomas Hofmann.
\newblock Escaping saddles with stochastic gradients.
\newblock {\em arXiv preprint arXiv:1803.05999}, 2018.

\bibitem{dauphin2014identifying}
Yann~N Dauphin, Razvan Pascanu, Caglar Gulcehre, Kyunghyun Cho, Surya Ganguli,
  and Yoshua Bengio.
\newblock Identifying and attacking the saddle point problem in
  high-dimensional non-convex optimization.
\newblock In {\em Advances in neural information processing systems}, pages
  2933--2941, 2014.

\bibitem{defazio2014saga}
Aaron Defazio, Francis Bach, and Simon Lacoste-Julien.
\newblock Saga: A fast incremental gradient method with support for
  non-strongly convex composite objectives.
\newblock In {\em Advances in neural information processing systems}, pages
  1646--1654, 2014.

\bibitem{du2017gradient}
Simon~S Du, Chi Jin, Jason~D Lee, Michael~I Jordan, Aarti Singh, and Barnabas
  Poczos.
\newblock Gradient descent can take exponential time to escape saddle points.
\newblock In {\em Advances in Neural Information Processing Systems}, pages
  1067--1077, 2017.

\bibitem{du2018power}
Simon~S Du and Jason~D Lee.
\newblock On the power of over-parametrization in neural networks with
  quadratic activation.
\newblock {\em arXiv preprint arXiv:1803.01206}, 2018.

\bibitem{duchi2011adaptive}
John Duchi, Elad Hazan, and Yoram Singer.
\newblock Adaptive subgradient methods for online learning and stochastic
  optimization.
\newblock {\em Journal of Machine Learning Research}, 12(Jul):2121--2159, 2011.

\bibitem{dunner2018distributed}
Celestine D{\"u}nner, Aurelien Lucchi, Matilde Gargiani, An~Bian, Thomas
  Hofmann, and Martin Jaggi.
\newblock A distributed second-order algorithm you can trust.
\newblock {\em arXiv preprint arXiv:1806.07569}, 2018.

\bibitem{ge2015escaping}
Rong Ge, Furong Huang, Chi Jin, and Yang Yuan.
\newblock Escaping from saddle points-online stochastic gradient for tensor
  decomposition.
\newblock In {\em COLT}, pages 797--842, 2015.

\bibitem{goh2017why}
Gabriel Goh.
\newblock Why momentum really works.
\newblock {\em Distill}, 2017.

\bibitem{gratton2017complexity}
Serge Gratton, Cl{\'e}ment~W Royer, Lu{\'\i}s~N Vicente, and Zaikun Zhang.
\newblock Complexity and global rates of trust-region methods based on
  probabilistic models.
\newblock {\em IMA Journal of Numerical Analysis}, 2017.

\bibitem{grosse2016kronecker}
Roger Grosse and James Martens.
\newblock A kronecker-factored approximate fisher matrix for convolution
  layers.
\newblock In {\em International Conference on Machine Learning}, pages
  573--582, 2016.

\bibitem{hagan1994training}
Martin~T Hagan and Mohammad~B Menhaj.
\newblock Training feedforward networks with the marquardt algorithm.
\newblock {\em IEEE transactions on Neural Networks}, 5(6):989--993, 1994.

\bibitem{hinton2006reducing}
Geoffrey~E Hinton and Ruslan~R Salakhutdinov.
\newblock Reducing the dimensionality of data with neural networks.
\newblock {\em science}, 313(5786):504--507, 2006.

\bibitem{jastrzkebski2017three}
Stanislaw Jastrzebski, Zachary Kenton, Devansh Arpit, Nicolas Ballas, Asja
  Fischer, Yoshua Bengio, and Amos Storkey.
\newblock Three factors influencing minima in sgd.
\newblock {\em arXiv preprint arXiv:1711.04623}, 2017.

\bibitem{kingma2014adam}
Diederik~P Kingma and Jimmy Ba.
\newblock Adam: A method for stochastic optimization.
\newblock {\em arXiv preprint arXiv:1412.6980}, 2014.

\bibitem{kohler2017sub}
Jonas~Moritz Kohler and Aurelien Lucchi.
\newblock Sub-sampled cubic regularization for non-convex optimization.
\newblock In {\em International Conference on Machine Learning}, 2017.

\bibitem{lecun2012efficient}
Yann~A LeCun, L{\'e}on Bottou, Genevieve~B Orr, and Klaus-Robert M{\"u}ller.
\newblock Efficient backprop.
\newblock In {\em Neural networks: Tricks of the trade}, pages 9--48. Springer,
  2012.

\bibitem{lee2016gradient}
Jason~D Lee, Max Simchowitz, Michael~I Jordan, and Benjamin Recht.
\newblock Gradient descent only converges to minimizers.
\newblock In {\em Conference on Learning Theory}, pages 1246--1257, 2016.

\bibitem{li2018convergence}
Xiaoyu Li and Francesco Orabona.
\newblock On the convergence of stochastic gradient descent with adaptive
  stepsizes.
\newblock {\em arXiv preprint arXiv:1805.08114}, 2018.

\bibitem{li2017convergence}
Yuanzhi Li and Yang Yuan.
\newblock Convergence analysis of two-layer neural networks with relu
  activation.
\newblock In {\em Advances in Neural Information Processing Systems}, pages
  597--607, 2017.

\bibitem{liu2018stochastic}
Liu Liu, Xuanqing Liu, Cho-Jui Hsieh, and Dacheng Tao.
\newblock Stochastic second-order methods for non-convex optimization with
  inexact hessian and gradient.
\newblock {\em arXiv preprint arXiv:1809.09853}, 2018.

\bibitem{ma2019inefficiency}
Linjian Ma, Gabe Montague, Jiayu Ye, Zhewei Yao, Amir Gholami, Kurt Keutzer,
  and Michael~W Mahoney.
\newblock Inefficiency of k-fac for large batch size training.
\newblock {\em arXiv preprint arXiv:1903.06237}, 2019.

\bibitem{martens2010deep}
James Martens.
\newblock Deep learning via hessian-free optimization.
\newblock In {\em ICML}, volume~27, pages 735--742, 2010.

\bibitem{martens2014new}
James Martens.
\newblock New insights and perspectives on the natural gradient method.
\newblock {\em arXiv preprint arXiv:1412.1193}, 2014.

\bibitem{martens2015optimizing}
James Martens and Roger Grosse.
\newblock Optimizing neural networks with kronecker-factored approximate
  curvature.
\newblock In {\em International conference on machine learning}, pages
  2408--2417, 2015.

\bibitem{masters2018revisiting}
Dominic Masters and Carlo Luschi.
\newblock Revisiting small batch training for deep neural networks.
\newblock {\em arXiv preprint arXiv:1804.07612}, 2018.

\bibitem{mizutani2008second}
Eiji Mizutani and Stuart~E Dreyfus.
\newblock Second-order stagewise backpropagation for hessian-matrix analyses
  and investigation of negative curvature.
\newblock {\em Neural Networks}, 21(2-3):193--203, 2008.

\bibitem{nesterov2013introductory}
Yurii Nesterov.
\newblock {\em Introductory lectures on convex optimization: A basic course},
  volume~87.
\newblock Springer Science \& Business Media, 2013.

\bibitem{nesterov2006cubic}
Yurii Nesterov and Boris~T Polyak.
\newblock Cubic regularization of newton method and its global performance.
\newblock {\em Mathematical Programming}, 108(1):177--205, 2006.

\bibitem{nocedal2006nonlinear}
Jorge Nocedal and Stephen~J Wright.
\newblock {\em Numerical optimization, 2nd Edition}.
\newblock Springer, 2006.

\bibitem{osawa2018second}
Kazuki Osawa, Yohei Tsuji, Yuichiro Ueno, Akira Naruse, Rio Yokota, and Satoshi
  Matsuoka.
\newblock Second-order optimization method for large mini-batch: Training
  resnet-50 on imagenet in 35 epochs.
\newblock {\em arXiv preprint arXiv:1811.12019}, 2018.

\bibitem{pascanu2013revisiting}
Razvan Pascanu and Yoshua Bengio.
\newblock Revisiting natural gradient for deep networks.
\newblock {\em arXiv preprint arXiv:1301.3584}, 2013.

\bibitem{paszke2017automatic}
Adam Paszke, Sam Gross, Soumith Chintala, Gregory Chanan, Edward Yang, Zachary
  DeVito, Zeming Lin, Luca Desmaison, Alban aComplexity bounds for second-order
  optimality in unconstrained optimizationnd~Antiga, and Adam Lerer.
\newblock Automatic differentiation in pytorch.
\newblock 2017.

\bibitem{pearlmutter1994fast}
Barak~A Pearlmutter.
\newblock Fast exact multiplication by the hessian.
\newblock {\em Neural computation}, 6(1):147--160, 1994.

\bibitem{robbins1985stochastic}
Herbert Robbins and Sutton Monro.
\newblock A stochastic approximation method.
\newblock In {\em The Annals of Mathematical Statistics - Volume 22, Number 3}.
  Institute of Mathematical Statistics, 1951.

\bibitem{schmidt2017minimizing}
Mark Schmidt, Nicolas Le~Roux, and Francis Bach.
\newblock Minimizing finite sums with the stochastic average gradient.
\newblock {\em Mathematical Programming}, 162(1-2):83--112, 2017.

\bibitem{schraudolph2002fast}
Nicol~N Schraudolph.
\newblock Fast curvature matrix-vector products for second-order gradient
  descent.
\newblock {\em Neural computation}, 14(7):1723--1738, 2002.

\bibitem{shalev2017failures}
Shai Shalev-Shwartz, Ohad Shamir, and Shaked Shammah.
\newblock Failures of gradient-based deep learning.
\newblock {\em arXiv preprint arXiv:1703.07950}, 2017.

\bibitem{steihaug1983conjugate}
Trond Steihaug.
\newblock The conjugate gradient method and trust regions in large scale
  optimization.
\newblock {\em SIAM Journal on Numerical Analysis}, 20(3):626--637, 1983.

\bibitem{tieleman2012lecture}
Tijmen Tieleman and Geoffrey Hinton.
\newblock Lecture 6.5-rmsprop: Divide the gradient by a running average of its
  recent magnitude.
\newblock {\em COURSERA: Neural networks for machine learning}, 4(2):26--31,
  2012.

\bibitem{tripuraneni2017stochastic}
Nilesh Tripuraneni, Mitchell Stern, Chi Jin, Jeffrey Regier, and Michael~I
  Jordan.
\newblock Stochastic cubic regularization for fast nonconvex optimization.
\newblock {\em arXiv preprint arXiv:1711.02838}, 2017.

\bibitem{van1998solving}
Patrick Van Der~Smagt and Gerd Hirzinger.
\newblock Solving the ill-conditioning in neural network learning.
\newblock In {\em Neural networks: tricks of the trade}, pages 193--206.
  Springer, 1998.

\bibitem{ward2018adagrad}
Rachel Ward, Xiaoxia Wu, and Leon Bottou.
\newblock Adagrad stepsizes: Sharp convergence over nonconvex landscapes, from
  any initialization.
\newblock {\em arXiv preprint arXiv:1806.01811}, 2018.

\bibitem{wigner1993characteristic}
Eugene~P Wigner.
\newblock Characteristic vectors of bordered matrices with infinite dimensions
  i.
\newblock In {\em The Collected Works of Eugene Paul Wigner}, pages 524--540.
  Springer, 1993.

\bibitem{xu2017second}
Peng Xu, Farbod Roosta-Khorasan, and Michael~W Mahoney.
\newblock Second-order optimization for non-convex machine learning: An
  empirical study.
\newblock {\em arXiv preprint arXiv:1708.07827}, 2017.

\bibitem{xu2017newton}
Peng Xu, Farbod Roosta-Khorasani, and Michael~W Mahoney.
\newblock Newton-type methods for non-convex optimization under inexact hessian
  information.
\newblock {\em arXiv preprint arXiv:1708.07164}, 2017.

\bibitem{yao2018inexact}
Zhewei Yao, Peng Xu, Farbod Roosta-Khorasani, and Michael~W Mahoney.
\newblock Inexact non-convex newton-type methods.
\newblock {\em arXiv preprint arXiv:1802.06925}, 2018.

\bibitem{zeiler2012adadelta}
Matthew~D Zeiler.
\newblock Adadelta: an adaptive learning rate method.
\newblock {\em arXiv preprint arXiv:1212.5701}, 2012.

\end{thebibliography}
